\documentclass[12pt]{article}
\usepackage{authblk}
\usepackage{amsmath}
\usepackage{amssymb}
\usepackage{amsfonts}
\usepackage{physics} 
\usepackage[colorlinks = true, citecolor = magenta, urlcolor = blue]{hyperref} 
\usepackage[ruled,vlined,linesnumbered]{algorithm2e}
\usepackage{graphicx}
\usepackage{natbib}
\usepackage{url} 
\usepackage[section]{placeins}
\usepackage{multirow}
\usepackage{threeparttable}

\newcommand*\samethanks[1][\value{footnote}]{\footnotemark[#1]}

\def\spacingset#1{\renewcommand{\baselinestretch}%
{#1}\small\normalsize} \spacingset{1}

\newcommand{\argmax}[1]{\underset{#1}{\operatorname{arg}\,\operatorname{max}}\;}
\newcommand{\argmin}[1]{\underset{#1}{\operatorname{arg}\,\operatorname{min}}\;}

\newcommand{\bfX}{\mathbf{X}}
\newcommand{\bfY}{\mathbf{Y}}
\newcommand{\bfx}{\mathbf{x}}

\newcommand{\bfa}{\mathbf{a}}
\newcommand{\bfb}{\mathbf{b}}

\newcommand{\bfz}{\mathbf{z}}
\newcommand{\bfw}{\mathbf{w}}

\newcommand{\bfA}{\mathbf{A}}
\newcommand{\bfB}{\mathbf{B}}
\newcommand{\bfF}{\mathbf{F}}
\newcommand{\bfG}{\mathbf{G}}
\newcommand{\bfH}{\mathbf{H}}
\newcommand{\bfZ}{\mathbf{Z}}
\newcommand{\bfW}{\mathbf{W}}

\newcommand{\bfR}{\mathbf{R}}

\newcommand{\bfI}{\mathbf{I}}

\newcommand{\bfbeta}{\boldsymbol{\beta}}

\newcommand{\bfSigma}{\boldsymbol{\Sigma}}
\newcommand{\bfTheta}{\boldsymbol{\Theta}}

\newcommand{\bfmu}{\boldsymbol{\mu}}
\newcommand{\bftau}{\boldsymbol{\tau}}

\newcommand{\bfepsilon}{\boldsymbol{\epsilon}}
\newcommand{\bfnu}{\boldsymbol{\nu}}

\newcommand{\bfgamma}{\boldsymbol{\gamma}}

\newcommand{\calI}{\mathcal{I}}
\newcommand{\calL}{\mathcal{L}}
\newcommand{\calN}{\mathcal{N}}
\newcommand{\calO}{\mathcal{O}}
\newcommand{\calP}{\mathcal{P}}

\newcommand{\calU}{\mathcal{U}}

\newcommand{\bbR}{\mathbb{R}}

\newcommand{\bbE}{\mathbb{E}}

\newcommand{\rmT}{\mathrm{T}}
\newcommand{\rmPr}{{\rm Pr}}
\newcommand{\diag}{{\rm diag}}
\newcommand{\const}{\text{const}}
\newcommand{\ELBO}{{\rm ELBO}}
\newcommand{\KL}{{\rm KL}}
\newcommand{\PVE}{{\rm PVE}}
\newcommand{\Var}{{\rm Var}}
\newcommand{\missing}{{\rm missing \  ratio}}

\begin{document}

\title{\bf MFAI: A Scalable Bayesian Matrix Factorization Approach to Leveraging Auxiliary Information}

\author[1]{Zhiwei Wang}
\author[1]{Fa Zhang}
\author[1]{Cong Zheng}
\author[1]{Xianghong Hu}
\author[2]{\\Mingxuan Cai\thanks{Corresponding authors}}
\author[1]{Can Yang\samethanks}

\affil[1]{Department of Mathematics, The Hong Kong University of Science and Technology}
\affil[2]{Department of Biostatistics, City University of Hong Kong}

\maketitle



\bigskip
\begin{abstract}
In various practical situations, matrix factorization methods suffer from poor data quality, such as high data sparsity and low signal-to-noise ratio (SNR).
Here, we consider a matrix factorization problem by utilizing auxiliary information, which is massively available in real-world applications, to overcome the challenges caused by poor data quality.
Unlike existing methods that mainly rely on simple linear models to combine auxiliary information with the main data matrix, we propose to integrate gradient boosted trees in the probabilistic matrix factorization framework to effectively leverage auxiliary information (MFAI).
Thus, MFAI naturally inherits several salient features of gradient boosted trees, such as the capability of flexibly modeling nonlinear relationships and robustness to irrelevant features and missing values in auxiliary information.
The parameters in MFAI can be automatically determined under the empirical Bayes framework, making it adaptive to the utilization of auxiliary information and immune to overfitting.
Moreover, MFAI is computationally efficient and scalable to large datasets by exploiting variational inference.
We demonstrate the advantages of MFAI through comprehensive numerical results from simulation studies and real data analyses.
Our approach is implemented in the R package \textit{mfair} available at \url{https://github.com/YangLabHKUST/mfair}.
\end{abstract}

\noindent%
\vfill

\newpage
\spacingset{1.5} 
\section{Introduction}
\label{sec:intro}

Matrix factorization \citep{srebro2004maximum, rennie2005fast, salakhutdinov2007probabilistic, salakhutdinov2008bayesian} is widely used when handling large-scale data.
It has become an important topic in the fields of applied mathematics, statistics, and machine learning because of its broad applications.
For example, as motivated by the Netflix Prize \citep{bell2007lessons, koren2009matrix}, matrix factorization has emerged as an effective method for inferring the unobserved entries, commonly referred to as the matrix completion problem. \citep{candes2009exact, takacs2009scalable, cai2010singular, candes2010power, mazumder2010spectral, ilin2010practical}.
Matrix factorization can also help uncover the underlying structures of datasets from diverse research topics, such as background modeling via low-rank approximation in moving object detection \citep{zhou2012moving, zhou2014low}, dimension reduction and adjustment for confounding variations \citep{yang2013accounting, lin2016simultaneous}.

Although existing matrix factorization methods have been used in various applications, major challenges still remain owing to low-quality data in practice.
First, the observed matrix can be very sparse for the matrix completion problem \citep{agarwal2009regression}.
For example, the MovieLens 100K dataset \citep{harper2015movielens} has only 100K observed ratings of 1,682 movies from 943 users, resulting in $94\%$ of entries missing in the rating matrix.
Such a high sparsity issue in matrix completion limits the accuracy of matrix factorization methods.
Second, the observed matrix can be quite noisy, and matrix factorization in low signal-to-noise ratio (SNR) settings tends to overfit easily \citep{mazumder2010spectral}.
Effective extraction of signals in the low SNR setting becomes critical for the success of matrix factorization.

A promising way to overcome the above challenges is to leverage auxiliary information \citep{singh2008relational, xu2013speedup, fithian2018flexible, agarwal2009regression, porteous2010bayesian, park2013hierarchical, gonen2014kernelized}, which is massively available in real-world applications \citep{zakeri2018gene, velten2022identifying, shang2022spatially}.
For example, besides the main matrix of movie ratings, information about users and movies is often available, including user profiles and descriptive words for movies.
To date, there have been a number of studies on matrix factorization with auxiliary information.
These methods can be roughly grouped into two categories: regularized methods and Bayesian methods.
For regularized methods, they often assume some shared structures between auxiliary information and the main matrix, such that auxiliary information can be incorporated to regularize the factorization of the main matrix.
This strategy can be very helpful when the observed data matrix is sparse or in a low SNR regime.
To name a few, collective matrix factorization (CMF) \citep{singh2008relational} jointly factorizes the main matrix along with the auxiliary information matrix, assuming they share some latent factors.
Inductive matrix completion (IMC) \citep{xu2013speedup} assumes that the main matrix lies in the subspace spanned by the auxiliary information matrix.
For Bayesian methods, they often build a probabilistic model where auxiliary information is incorporated through a linear model.
For example, regression-based latent factor models (RLFM) \citep{agarwal2009regression} assume that the latent factor matrices are generated from the auxiliary information via linear regression.
Bayesian matrix factorization with side information (BMFSI) \citep{porteous2010bayesian} augments the factor matrices with additional terms that incorporate auxiliary information using a linear model.

Despite many efforts in the incorporation of auxiliary information, several main issues remain.
First, existing methods rely on linear models to combine auxiliary information with the main matrix, which may limit its role because linear models are not flexible enough.
A more flexible framework is highly desired to take full advantage of auxiliary information.
Second, the computational costs of existing methods are often quite expensive, even though only linear models are used.
For example, Bayesian methods often use sampling methods to approximate posterior distributions, such as Markov Chain Monte Carlo (MCMC) \citep{neal1993probabilistic}.
However, sampling methods are often too computationally expensive to scale up to handle large datasets.
For some regularized methods, efficient implementation is also lacking due to the challenge of parallelization \citep{hubbard2017parallel, zilber2022inductive}.
The computational issue will become more serious when we allow more flexible nonlinear models.
Third, incorporating irrelevant information will not improve but degrade the performance.
Existing methods largely rely on parameter tuning to control the amount of auxiliary information incorporated.
Although cross-validation can help with this, it will become very annoying and time-consuming when there are many tuning parameters.
Statistical methods that can adaptively leverage auxiliary information are highly demanding.

In this article, we develop a scalable Bayesian \textbf{M}atrix \textbf{F}actorization approach to adaptively leveraging \textbf{A}uxiliary \textbf{I}nformation (MFAI).
MFAI allows a flexible nonlinear model to incorporate auxiliary information, enriching its role in the matrix factorization framework.
Specifically, MFAI is a unified probabilistic approach to integrating gradient boosted trees \citep{freund1996experiments, breiman1998arcing, friedman2000additive, mason1999boosting, friedman2001greedy, buhlmann2007boosting, sigrist2021gradient} with matrix factorization.
Through innovations in the model and algorithm designs, MFAI has several unique advantages over existing matrix factorization methods.
First, MFAI naturally inherits several salient features of gradient boosted trees, such as the capability of flexibly modeling nonlinear relationships, robustness to irrelevant features and missing values in predictors, and ranking the relative importance of auxiliary information, which offers more interpretable insights \citep{elith2008working, sigrist2020gaussian, sigrist2022latent, grinsztajn2022tree}.
Second, the parameters in MFAI can be automatically determined under the empirical Bayes framework, making it adaptive to the utilization of auxiliary information.
Third, MFAI is computationally efficient and scalable to large datasets by exploiting variational inference (VI) \citep{bishop2006pattern, blei2017variational}.
Through comprehensive simulation experiments and real data studies, we demonstrate that MFAI can perform better in matrix factorization and completion tasks than the existing methods.
The R package \textit{mfair} is available at \url{https://github.com/YangLabHKUST/mfair}, serving as a user-friendly tool for matrix factorization with auxiliary information.

\section{Methods}
\label{sec:meth}



\subsection{The MFAI Model}
\label{ssec:model}

Given the main data matrix $\bfY \in \bbR^{N \times M}$ of $N$ samples and $M$ features, we consider the following matrix factorization problem:
\begin{equation}\label{low_rank}
    \bfY = \bfZ \bfW^{\rmT} + \bfepsilon,
\end{equation}
where $\bfZ \in \bbR^{N \times K}$ and $\bfW \in \bbR^{M \times K}$ are two matrices with $K \leq \min \left\{ N, M \right\}$, and $\bfepsilon \in \bbR^{N \times M}$ is a matrix of residual error terms.
Here, we adopt the terminology of factor analysis and refer to $\bfZ$ as the ``factors'', $\bfW$ as the ``loadings'', and $K$ as the number of factors.
We can further expand the above formulation as the sum of the $K$ factors
\begin{equation}\label{mf}
    \bfY = \sum_{k = 1}^K \bfZ_{\cdot k} \bfW_{\cdot k}^{\rmT} + \bfepsilon,
\end{equation}
where $\bfZ_{\cdot k}$ and $\bfW_{\cdot k}$ are the $k$-th column of $\bfZ$ and $\bfW$, respectively.
Let’s take a user-movie rating matrix $\bfY$ as an example, where the observed entry $\bfY_{nm}$ represents the rating score of user $n$ for movie $m$.
We may assume that the $K$ factors represent the $K$ traits of movies, where $\bfW_{\cdot k}$ corresponds to the scores of $M$ movies related to the $k$-th trait (e.g., action or emotional), $\bfZ_{\cdot k}$ corresponds to the preference of $N$ users on the $k$-th trait.
The product of preference and score on the $k$-th trait measures the strength of the $k$-th factor.
The final rating depends on the overall effects of $K$ factors.
To perform matrix factorization of $\bfY$, we not only have observed entries in the main matrix but also some auxiliary information.
For example, we often have user information $\bfX \in \bbR^{N \times C}$ in the above movie rating case, where each row also represents a user with $C$ covariates, such as gender, age, and occupation.
We incorporate auxiliary information into matrix factorization by assuming that the users' preferences are associated with their covariates.
Specifically, we relate $\bfZ_{\cdot k}$ and auxiliary covariates $\bfX$ using the following probabilistic model:
\begin{equation} \label{z_prior}
    \bfZ_{\cdot k} \sim \calN_N \left( F_k \left( \bfX \right), \beta_k^{-1} \bfI_N \right), \  k = 1, \dots, K,
\end{equation}
where $F_k \left( \bfX \right) \in \bbR^{N \times 1}$ is the mean vector of the factor $\bfZ_{.k}$, $\beta_k$ is the precision, $\bfI_N \in \bbR^{N \times N}$ is an identity matrix, and $\calN_N(\bfmu, \bfSigma)$ denotes the $N$-variate Gaussian distribution with mean $\bfmu$ and covariance $\bfSigma$.
Note that $F_k \left( \bfX \right)$ is the row-wise evaluation of the unknown function $F_k: \bbR^{C} \rightarrow \bbR$, $F_k \left( \bfX \right) = \left( F_k \left( \bfX_{1 \cdot} \right), \dots, F_k \left( \bfX_{N \cdot} \right) \right)^{\rmT}$, where $\bfX_{n \cdot} = \left( \bfX_{n1}, \dots, \bfX_{nC} \right)^{\rmT} \in \bbR^{C \times 1}$ is the $n$-th row of $\bfX$ containing auxiliary information for the $n$-th sample, $n = 1, \dots, N$.
Regarding $F_k \left( \cdot \right)$, it is often assumed to be a linear function, $F_k \left( \bfX_{n \cdot} \right) = \gamma_0 + \bfX_{n \cdot}^{\rmT} \bfgamma$, where $\gamma_0$ is the intercept and $\bfgamma \in \bbR^{C \times 1}$ represents regression coefficients.
However, the linearity assumption can be too simplified to characterize a flexible relationship between $\bfZ_{\cdot k}$ and $\bfX$.
In the MFAI model, we assume that $F_k \left( \cdot \right)$ in \eqref{z_prior} is a nonlinear function represented by a tree ensemble
\begin{equation}\label{F}
    F_k \left( \cdot \right) = \sum_{t = 1}^{T_k} f_{k}^{t} \left( \cdot \right),
\end{equation}
where $f_{k}^{t} \left( \cdot \right)$ is a regression tree \citep{breiman1984classification}, and $T_k$ is the total number of trees.
We then assign an independent Gaussian prior for the corresponding $k$-th loading $\bfW_{.k}$
\begin{equation}\label{w_prior}
    \bfW_{.k} \sim \calN_M(0, \bfI_M),
\end{equation}
which can push the variability to the factor $\bfZ_{\cdot k}$ side and partially help avoid the non-identifiability issue.
Matrices $\bfZ$ and $\bfW$ here are often referred to as latent variables in the statistical machine learning literature.
At last, we assume independent Gaussian error terms
\begin{equation}\label{error}
    \bfepsilon_{nm} \sim \calN(0, \tau^{-1}), \  n = 1, \dots, N \  \text{and} \  m = 1, \dots, M,
\end{equation}
where $\tau$ is the shared precision parameter for all $\bfepsilon_{nm}$.
The advantages of the proposed model are threefold.
First, tree ensembles are able to capture a more flexible relationship between $\bfZ_{\cdot k}$ and $\bfX$.
Second, the proposed model naturally inherits several salient features of regression trees, such as ranking variable importance and handling missing values.
Third, we can develop an efficient algorithm to estimate the nonlinear model and make it scalable to large datasets.

Let $\bfTheta = \{ \tau, \bfbeta \} = \{ \tau; \beta_1, \dots, \beta_K \}$ be the collection of model parameters and $\bfF \left( \cdot \right) = \{ F_1 \left( \cdot \right), \dots, F_K \left( \cdot \right) \}$ be the collection of $K$ unknown functions.
Combining model \eqref{mf} \eqref{z_prior} \eqref{w_prior} \eqref{error}, we can write down the joint probabilistic model as
\begin{equation}\label{joint}
    \begin{aligned}
        \rmPr \left( \bfY, \bfZ, \bfW \mid \bfTheta, \bfF \left( \cdot \right) \right) & = \rmPr \left( \bfY \mid \bfZ, \bfW; \tau \right) \rmPr \left(\bfZ \mid \bfbeta, \bfF \left( \cdot \right) \right) \rmPr \left( \bfW \right) \\
        & = \rmPr \left( \bfY \mid \bfZ, \bfW; \tau \right) \prod_{k=1}^K \rmPr \left( \bfZ_{.k} \mid \beta_k, F_k \left( \cdot \right) \right) \prod_{k=1}^K \rmPr \left( \bfW_{.k} \right).
    \end{aligned} 
\end{equation}
As an empirical Bayes approach, we can adaptively estimate $\bfTheta$ and $\bfF \left( \cdot \right)$ by optimizing the log marginal likelihood
\begin{equation}\label{max_marginal}
    \begin{aligned}
        \left(\widehat{\bfTheta}, \widehat{\bfF} \left( \cdot \right) \right) & = \argmax{\bfTheta, \bfF \left( \cdot \right)} \log \rmPr \left( \bfY \mid \bfTheta, \bfF \left( \cdot \right) \right) \\
        & = \argmax{\bfTheta, \bfF \left( \cdot \right)} \log \int \rmPr \left( \bfY, \bfZ, \bfW \mid \bfTheta, \bfF \left( \cdot \right) \right) \, d\bfZ \, d\bfW.
    \end{aligned} 
\end{equation}
Then, we can infer the latent factors and loadings by obtaining their posterior probability as
\begin{equation}\label{post}
    \begin{aligned}
    \rmPr \left( \bfZ, \bfW \mid \bfY; \widehat{\bfTheta}, \widehat{\bfF}\left( \cdot \right) \right) = \frac{\rmPr \left( \bfY, \bfZ, \bfW \mid \widehat{\bfTheta}, \widehat{\bfF} \left( \cdot \right) \right)}{\rmPr \left( \bfY \mid \widehat{\bfTheta}, \widehat{\bfF}\left( \cdot \right) \right)}.
    \end{aligned}
\end{equation}

\subsection{Fitting the MFAI Model}
\label{ssec:fit}

We begin our algorithm design with the single-factor case, i.e., $K = 1$, and extend our algorithm to the multi-factor case in Section \ref{sssec:kfac}.
The single-factor MFAI model is as follows
\begin{equation} \label{rank1}
    \begin{aligned}
        & \bfY = \bfz \bfw^{\rmT} + \bfepsilon, \\
        & \bfz \sim \calN_N(F \left( \bfX \right), \beta^{-1}\bfI_N), \\
        & \bfw \sim \calN_M(0, \bfI_M), \\
        & \epsilon_{nm} \sim \calN(0, \tau^{-1}), \  n = 1, \dots, N \  \text{and} \  m = 1, \dots, M,
    \end{aligned}
\end{equation}
where $\bfz \in \bbR^{N \times 1}$, $\bfw \in \bbR^{M \times 1}$, and $F \left( \bfX \right) \in \bbR^{N \times 1}$ denotes $\left( F \left( \bfX_{1 \cdot} \right), \dots, F \left( \bfX_{N \cdot} \right) \right)^{\rmT}$.
Then, the joint probabilistic model becomes
\begin{equation}\label{joint_rank1}
    \rmPr \left( \bfY, \bfz, \bfw \mid \bfTheta, F \left( \cdot \right) \right) = \rmPr \left( \bfY \mid \bfz, \bfw; \tau \right) \rmPr \left( \bfz \mid \beta, F \left( \cdot \right) \right) \rmPr \left( \bfw \right),
\end{equation}
where $\bfTheta = \{ \tau, \beta \}$ is the collection of model parameters and $F \left( \cdot \right)$ is an unknown function.
The goal is to estimate $\bfTheta$ and $F \left( \cdot \right)$ by optimizing the log marginal likelihood
\begin{equation}\label{max_marginal_rank1}
    \begin{aligned}
        \left(\widehat{\bfTheta}, \widehat{F} \left( \cdot \right) \right) & = \argmax{\bfTheta, F \left( \cdot \right)} \log \rmPr \left( \bfY \mid \bfTheta, F \left( \cdot \right) \right) \\
        & = \argmax{\bfTheta, F \left( \cdot \right)} \log \int \rmPr \left( \bfY, \bfz, \bfw \mid \bfTheta, F \left( \cdot \right) \right) \, d\bfz \, d\bfw.
    \end{aligned} 
\end{equation}
The posterior probability of $\bfz$ and $\bfw$ is given as
\begin{equation}\label{post_rank1}
    \rmPr \left( \bfz, \bfw \mid \bfY; \widehat{\bfTheta}, \widehat{F} \left( \cdot \right) \right) = \frac{\rmPr \left( \bfY, \bfz, \bfw \mid \widehat{\bfTheta}, \widehat{F} \left( \cdot \right) \right)}{\rmPr \left( \bfY \mid \widehat{\bfTheta}, \widehat{F} \left( \cdot \right) \right)}.
\end{equation}

\subsubsection{Approximate Bayesian Inference}
\label{sssec:vi}

The Bayesian inference using \eqref{max_marginal_rank1} and \eqref{post_rank1} is intractable since the marginal likelihood $\rmPr \left( \bfY \mid \bfTheta, F \left( \cdot \right) \right)$ cannot be computed by marginalizing all latent variables.
To tackle the Bayesian inference problem, there are two main methods: Markov Chain Monte Carlo (MCMC) \citep{neal1993probabilistic}, which is a sampling-based approach, and variational inference (VI), which is an approximation-based approach \citep{bishop2006pattern, blei2017variational}.
The advantage of the sampling-based methods is that they produce exact results asymptotically.
In practice, however, they are often too computationally expensive for large-scale problems.
Here, we propose a variational expectation-maximization (EM) algorithm to perform approximate Bayesian inference, with details in Appendix Section A.
To apply variational approximation, we first introduce $q \left( \bfz, \bfw \right)$ as an approximated distribution of posterior $\rmPr \left( \bfz, \bfw \mid \bfY; \bfTheta, F \left( \cdot \right) \right)$.
Then, we can obtain the evidence lower bound (ELBO) of the logarithm of the marginal likelihood using Jensen’s inequality
\begin{equation}\label{jensen}
    \begin{aligned}
        \log \rmPr \left( \bfY \mid \bfTheta, F \left( \cdot \right) \right)
        & = \log \int \rmPr \left( \bfY, \bfz, \bfw \mid \bfTheta, F \left( \cdot \right) \right) \, d\bfz \, d\bfw \\
        & \geq \int q \left( \bfz, \bfw \right) \log \frac{\rmPr \left( \bfY, \bfz, \bfw \mid \bfTheta, F \left( \cdot \right) \right)}{q \left( \bfz, \bfw \right)} \, d\bfz \, d\bfw \\
        & = \bbE_{q} \left[ \log \rmPr \left( \bfY, \bfz, \bfw \mid \bfTheta, F \left( \cdot \right) \right) \right] - \bbE_{q} \left[ \log q \left( \bfz, \bfw \right) \right] \\
        & \triangleq \ELBO \left( q; \bfTheta, F \left( \cdot \right) \right),
    \end{aligned}
\end{equation}
where the equality holds if and only if $q \left( \bfz, \bfw \right)$ is the exact posterior $\rmPr \left( \bfz, \bfw \mid \bfY; \bfTheta, F \left( \cdot \right) \right)$.
We can also derive the above inequality by decomposing the logarithm of the marginal likelihood as
\begin{equation}\label{log_marginal_decomp}
    \log \rmPr \left( \bfY \mid \bfTheta, F \left( \cdot \right) \right)
    = \ELBO \left( q; \bfTheta, F \left( \cdot \right) \right) + \KL \left( q \parallel \rmPr \left( \bfz, \bfw \mid \bfY; \bfTheta, F \left( \cdot \right) \right) \right),
\end{equation}
for any choice of  $q \left( \bfz, \bfw \right)$, where
\begin{equation}\label{kl}
    \KL \left( q \parallel \rmPr \left( \bfz, \bfw \mid \bfY; \bfTheta, F \left( \cdot \right) \right) \right)
    = - \int q \left( \bfz, \bfw \right) \log \frac{\rmPr \left( \bfz, \bfw \mid \bfY; \bfTheta, F \left( \cdot \right) \right)}{q \left( \bfz, \bfw \right)} \, d\bfz \, d\bfw,
\end{equation}
is the Kullback-Leibler (KL) divergence \citep{kullback1951information}.
The bound follows from the fact that $\KL \left( \cdot \parallel \cdot \right) \geq 0$ and the equality holds if and only if the two distributions are the same.
Then, instead of maximizing $\log \rmPr \left( \bfY \mid \bfTheta, F \left( \cdot \right) \right)$, we can iteratively maximize the ELBO with respect to the variational approximate posterior $q$, the model parameters $\bfTheta$, and the function $F \left( \cdot \right)$
\begin{equation}\label{max_elbo}
    \left( \widehat{q}; \widehat{\bfTheta}, \widehat{F}\left( \cdot \right) \right) = \argmax{q; \bfTheta, F \left( \cdot \right)} \ELBO \left( q; \bfTheta, F \left( \cdot \right) \right).
\end{equation}
Using the terminology in the EM algorithm, maximization of ELBO with respect to $q$ is known as the E-step, which is equivalent to minimizing the KL divergence in \eqref{log_marginal_decomp}, and maximization of ELBO with respect to $\bfTheta$ and $F \left( \cdot \right)$ is known as the M-step.
Actually, our variational EM algorithm is the EM algorithm with a variational E-step, that is, the computation of an approximate posterior.

To approximate the posterior distribution, we consider the following factorization of variational distribution $q \left( \bfz, \bfw \right)$ based on the mean-field theory \citep{bishop2006pattern, blei2017variational}:
\begin{equation} \label{mean_field}
    q \left( \bfz, \bfw \right) = q \left( \bfz \right) q \left( \bfw \right).
\end{equation}
Without further assumptions, we show that the optimal solutions of $q \left( \bfz \right)$ and $q \left( \bfw \right)$ in the E-step are given as two multivariate Gaussian distributions
\begin{equation}
    q \left( \bfz \right) = \calN_N \left( \bfz \mid \bfmu, \bfA \right),\  q \left( \bfw \right) = \calN_M \left( \bfw \mid \bfnu, \bfB \right), \label{qz_qw}
\end{equation}
where $\bfmu \in \bbR^{N \times 1}$ and $\bfnu \in \bbR^{M \times 1}$ are posterior mean vectors, $\bfA \in \bbR^{N \times N}$ and $\bfB \in \bbR^{M \times M}$ are posterior covariance matrices
\begin{equation} \label{cov_matrix}
    \bfA = a^2 \bfI_N, \  \bfB = b^2 \bfI_M.
\end{equation}
Now suppose that we are at the $t$-th step of the variational EM algorithm, and we have obtained the current estimates $\left\{ \bfmu^{(t-1)}, {a^2}^{(t-1)}; \bfnu^{(t-1)}, {b^2}^{(t-1)} \right\}$, $\bfTheta^{(t-1)} = \left\{ \tau^{(t-1)}, \beta^{(t-1)} \right\}$, and $F^{(t-1)} \left( \cdot \right)$ at the $(t-1)$-th step.
To maximize ELBO in the $t$-th E-step, we can update variational parameters as
\begin{equation} \label{update_q}
    \begin{aligned}
        & {a^2}^{(t)} = \frac{1}{\beta^{(t-1)} + \tau^{(t-1)} \left( \left\| \bfnu^{(t-1)} \right\|_2^2 + M {b^2}^{(t-1)} \right)}, \\
        & \bfmu^{(t)} = {a^2}^{(t)} \left( \beta^{(t-1)} F^{(t-1)} \left( \bfX \right) + \tau^{(t-1)} \bfY \bfnu^{(t-1)} \right), \\
        & {b^2}^{(t)} = \frac{1}{1 + \tau^{(t-1)} \left( \left\| \bfmu^{(t)} \right\|_2^2 + N {a^2}^{(t)} \right)}, \\
        & \bfnu^{(t)} = {b^2}^{(t)} \tau^{(t-1)} \bfY^{\rmT} \bfmu^{(t)},
    \end{aligned}
\end{equation}
where $\left\| \cdot \right\|_{2}$ for a vector denotes the Euclidean norm.
It is obvious that the variational approximate posterior $q^{(t)} \left( \bfz, \bfw \right)$ only depends on the values of parameters $\left\{ \bfmu^{(t)}, {a^2}^{(t)}; \bfnu^{(t)}, {b^2}^{(t)} \right\}$.
We also note that each entry of $\bfz$ has the same posterior variance ${a^2}^{(t)}$, and $\bfw$ as well.
Under the variational approximation framework, $q^{(t)} \left( \bfz \right)$ can be viewed as the approximation to the posterior distribution $\rmPr \left( \bfz \mid \bfY; \bfTheta^{(t-1)}, F^{(t-1)} \left( \cdot \right) \right)$, and $q^{(t)} \left( \bfw \right)$ can be viewed as the approximation to the posterior distribution $\rmPr \left( \bfw \mid \bfY; \bfTheta^{(t-1)}, F^{(t-1)} \left( \cdot \right) \right)$.
Naturally, we can infer $\bfz$ and $\bfw$ using the posterior mean $\bfmu^{(t)}$ and $\bfnu^{(t)}$ after the convergence of the algorithm.

With $q^{(t)}$ updated in the $t$-th E-step, we can update $\bfTheta = \left\{ \tau, \beta \right\}$ and $F \left( \cdot \right)$ in the M-step
\begin{equation} \label{opt_elbo}
    \left( \bfTheta^{(t)}, F^{(t)} \left( \cdot \right) \right) = \argmax{\bfTheta, F \left( \cdot \right)} \ELBO \left( q^{(t)}; \bfTheta, F \left( \cdot \right) \right),
\end{equation}
where
\begin{equation}\label{elbo}
    \begin{aligned}
        & \ELBO \left(q^{(t)} \left( \bfz, \bfw \mid \bfmu^{(t)}, {a^2}^{(t)}; \bfnu^{(t)}, {b^2}^{(t)} \right); \tau, \beta, F \left( \cdot \right) \right) \\
        = & \bbE_{q^{(t)} \left( \bfz, \bfw \right)} \left[ \log \rmPr \left( \bfY, \bfz, \bfw \mid \bfTheta, F \left(\cdot \right) \right) \right] - \bbE_{q^{(t)} \left( \bfz, \bfw \right)} \left[ \log q^{(t)}(\bfz, \bfw) \right] \\   
        =& - \frac{\tau}{2} \left( \left\| \bfY^{(t)} - \bfmu^{(t)} {\bfnu^{(t)}}^{\rmT} \right\|_{\text{F}}^2 + \left\| \left( {\bfmu^{(t)}}^2 + \diag \left( \bfA^{(t)} \right) \right) \left({\bfnu^{(t)}}^2 + \diag \left( \bfB^{(t)} \right) \right)^{\rmT} - {\bfmu^{(t)}}^2 \left( {\bfnu^{(t)}}^2 \right)^{\rmT} \right\|_{1,1} \right)\\
        & + \frac{NM}{2} \log \tau + \frac{N}{2} \log \beta - \frac{\beta}{2} \left( \left\| \bfmu^{(t)} - F \left( \bfX \right) \right\|_2^2 + N {a^2}^{(t)} \right) + \const.
    \end{aligned} 
\end{equation}
Here, for a matrix, $\|\cdot\|_{\text{F}}$ denotes the Frobenius norm, and  $\|\cdot\|_{1,1}$ denotes the entry-wise matrix norm, that is, the sum of the absolute value of all the entries.
We first consider optimizing model parameters with the estimate of the function fixed as $F^{(t-1)} \left( \cdot \right)$.
By solving 
\begin{equation} \label{m_step_para}
    \left. \pdv{\ELBO \left( q; \bfTheta, F \left( \cdot \right) \right)}{\bfTheta} \right|_{q = q^{(t)}, \ F \left( \cdot \right) = F^{(t-1)} \left( \cdot \right)} = 0
\end{equation}
we obtain the update equations for parameter estimation in the $t$-th M-step
\begin{equation} \label{update_para}
    \begin{aligned}
        & \tau^{(t)} = \frac{NM}{\left\| \bfY - \bfmu^{(t)} {\bfnu^{(t)}}^{\rmT} \right\|_{\text{F}}^2 + \left\| \left( {\bfmu^{(t)}}^2 + \diag \left( \bfA^{(t)} \right) \right) \left( {\bfnu^{(t)}}^2 + \diag \left( \bfB^{(t)} \right) \right)^{\rmT} - {\bfmu^{(t)}}^2 \left( {\bfnu^{(t)}}^2 \right)^{\rmT} \right\|_{1,1}}, \\
        & \beta^{(t)} = \frac{N}{\left\| \bfmu^{(t)} - F^{(t-1)} \left( \bfX \right) \right\|_2^2 + N {a^2}^{(t)}},
    \end{aligned}
\end{equation}
where $\bfmu^2 \in \bbR^{N \times 1}$ denotes Hadamard product $\bfmu^2 = \bfmu \odot \bfmu$, and $\diag \left( \bfA \right) \in \bbR^{N \times 1}$ denotes a vector containing all the entries on the main diagonal of $\bfA$. 
To maximize ELBO with respect to $F \left( \cdot \right)$, we propose to update $F \left( \cdot \right)$ from its current estimate $F^{(t-1)} \left( \cdot \right)$ to its new estimate $F^{(t)} \left( \cdot \right)$ by constructing the following additive model:
\begin{equation} \label{boosting}
    F^{(t)} \left( \cdot \right) = F^{(t-1)} \left( \cdot \right) + f^{(t)} \left( \cdot \right),
\end{equation}
where $f^{(t)} \left( \cdot \right)$ is chosen to be a single regression tree \citep{breiman1984classification} in MFAI.
Clearly, this leads to the following optimization problem:
\begin{equation} \label{find_f}
    \begin{gathered}
        f^{(t)} \left( \cdot \right) = \argmin{f \left( \cdot \right)} \left\| \bfmu^{(t)} - F^{(t-1)} \left( \bfX \right) - f \left( \bfX \right) \right\|_2^2,
    \end{gathered}
\end{equation}
where $\bfmu^{(t)}$ is given in the $t$-th E-step and $F^{(t-1)} \left( \cdot \right)$ is the current estimated function. 
After obtaining  $f^{(t)} \left( \cdot \right)$ from \eqref{find_f}, we update $F \left( \cdot \right)$ as
\begin{equation} \label{shrinkage}
    F^{(t)} \left( \cdot \right) = F^{(t-1)} \left( \cdot \right) + s \cdot f^{(t)} \left( \cdot \right),
\end{equation}
where $0 < s < 1$ is the so-called shrinkage parameter or learning rate, whose default value is set relatively small ($s = 0.1$) in our implementation to avoid potential overfitting.
Clearly, information in the auxiliary matrix is gradually incorporated to modulate the prior of $\bfz$, and the corresponding posterior mean will be updated as $\bfmu^{(t+1)}= {a^2}^{(t)} \left( \beta^{(t)} F^{(t)} \left( \bfX \right) + \tau^{(t)} \bfY \bfnu^{(t)} \right)$ in the $(t+1)$-th step.

To summarize, the proposed algorithm is a variational EM algorithm and its convergence is naturally guaranteed.
In the E-step, we optimize the variational distribution $q \left( \bfz, \bfw \right)$ to maximize the ELBO.
In the M-step, we update the model parameters $\bfTheta$ and function $F \left( \cdot \right)$ to optimize the ELBO.
The novelty of the proposed algorithm comes from the way we update function $F \left( \cdot \right)$, where we combine the gradient boosting strategy \eqref{shrinkage} into the iterations of our variational EM algorithm.
In such a way, the nonlinear relationship between auxiliary information $\bfX$ and factor $\bfz$ can be built up in a stage-wise fashion, yielding a very stable way to incorporate auxiliary information in matrix factorization.
In the meanwhile, the algorithm fully exploits the advantages of tree-based methods when fitting a single tree $f^{(t)} \left( \cdot \right)$ in the $t$-th step, such as ranking variable importance and handling missing values with trees.
Last but not least, our algorithm can be efficient when handling large-scale problems because trees are scalable to large auxiliary matrix $\bfX$.
We denote the proposed algorithm to fit the single-factor model as MFAI\_SF and summarize it in Algorithm \ref{algo_rank1}.
\begin{algorithm}[!htb]
    \label{algo_rank1}
    \caption{Fitting the Single-Factor MFAI Model}
    \SetAlgoLined
    \KwData{Main data matrix $\bfY$ and auxiliary matrix $\bfX$}
    \KwResult{Estimate of the latent variables $\widehat{\bfz} = \bfmu$ and $\widehat{\bfw} = \bfnu$}
    \BlankLine
    initialize approximate posterior $q^{(0)} = q \left( \bfmu^{(0)}, {a^2}^{(0)}; \bfnu^{(0)}, {b^2}^{(0)} \right)$ \;
    initialize model parameters $\bfTheta^{(0)} = \left\{ \tau^{(0)}, \beta^{(0)} \right\}$ \;
    initialize function $F^{(0)} \left( \cdot \right) = 0$, then prior means $F^{(0)} \left( \bfX \right) = 0$ \;
    $t \leftarrow 0$ \;
    \Repeat{convergence criterion satisfied}{
        $t \leftarrow t + 1$ \;
        $\bfmu^{(t)}, {a^2}^{(t)}; \bfnu^{(t)}, {b^2}^{(t)} \leftarrow \argmax{\bfmu, a^2; \bfnu, b^2} \ELBO \left( q \left( \bfmu, a^2; \bfnu, b^2 \right); \tau^{(t-1)}, \beta^{(t-1)}; F^{(t-1)} \left( \cdot \right) \right)$ \tcp*{update the variational approximation of posterior}
        $\tau^{(t)}, \beta^{(t)} \leftarrow \argmax{\tau, \beta} \ELBO \left( q \left( \bfmu^{(t)}, {a^2}^{(t)}; \bfnu^{(t)}, {b^2}^{(t)} \right); \tau, \beta; F^{(t-1)} \left( \cdot \right) \right)$ \tcp*{update the model parameters}
        $f^{(t)} \left( \cdot \right) \leftarrow \argmax{f \left( \cdot \right)} \ELBO \left( q \left( \bfmu^{(t)}, {a^2}^{(t)}; \bfnu^{(t)}, {b^2}^{(t)} \right); \tau^{(t)}, \beta^{(t)}; F^{(t-1)} \left( \cdot \right) + f \left( \cdot \right) \right)$ \tcp*{compute the functional gradient}
        $F^{(t)} \left( \cdot \right) \leftarrow F^{(t-1)} \left( \cdot \right) + s \cdot f^{(t)} \left( \cdot \right)$ \tcp*{update the function}
    }
    \KwRet{$\bfmu^{(t)}, {a^2}^{(t)}; \bfnu^{(t)}, {b^2}^{(t)}; \tau^{(t)}, \beta^{(t)}; F^{(t)} \left( \cdot \right)$.}
\end{algorithm}

To assess the scalability of MFAI, we analyze the computational complexity of each step in the proposed algorithm.
We begin with the computations of the approximate posterior mean and variance in the E-step, which are given by \eqref{update_q}.
The updates for both $\bfmu$ and $\bfnu$ rely on matrix-vector multiplications, resulting in a computational complexity of $\calO \left( NM \right)$.
The updates for $a^2$ and $b^2$ primarily involve the calculation of the $\ell^2$-norm and require $\calO \left( M \right)$ and $\calO \left( N \right)$ computations, respectively.
Moving on to the M-step, we need to update the model parameters $\bfTheta$ as in \eqref{update_para} and fit a single regression tree as in \eqref{find_f}.
The update for $\tau$ entails matrix multiplications and matrix norm calculations, resulting in a total computational complexity of $\calO \left( NM \right)$.
The update for $\beta$ also requires the calculation of the $\ell^2$-norm, resulting in the computational complexity of $\calO \left( N \right)$.
To fit a single regression tree, we need to sort the data for each node and each auxiliary feature, which takes $\calO \left( N \log N \right)$ computations.
Following this, we traverse the data points to find the best threshold, which takes $\calO \left( N \right)$ computations.
Considering all $C$ auxiliary features, the total computational complexity would be of $\calO \left( C N \log N \right)$.
These complexity analyses provide insights into the computational requirements of the MFAI algorithm and its scalability.



\subsubsection{Missing Data}
\label{sssec:miss}

One important feature of MFAI is its ability to handle missing data, either in the main matrix $\bfY$ or in the auxiliary matrix $\bfX$.
To handle $\bfY$ with missing entries, we first make the typical assumption that they are missing at random (MAR) \citep{rubin1976inference, little1987statistical}, that is, given the observed data, the missingness does not depend on the unobserved data or latent variables.
Then, we can consider the following probabilistic model only for the observed entries $\bfY^{\text{obs}}$:
\begin{equation} \label{lh_missing}
    \begin{aligned}
        \rmPr \left( \bfY^{\text{obs}} \mid \bfz, \bfw; \tau \right) = \prod_{\left(n, m \right) \in  \Omega^\text{obs}} \rmPr \left( \bfY_{nm} \mid \bfz, \bfw; \tau \right),
    \end{aligned}
\end{equation}
where $\Omega^\text{obs}$ is the collection of the indices of the observed entries of $\bfY$.
Within the approximate Bayesian inference framework, the update equations for the variational approximate posteriors are similar to \eqref{update_q}.
For $n$-th entry of $\bfz$, the posterior variance ${\bfa^2}_n^{(t)}$ and posterior mean $\bfmu_n^{(t)}$ are updated at the $t$-th step as
\begin{equation} \label{update_qz_miss}
    \begin{aligned}
        & {\bfa^2}_n^{(t)} = \frac{1}{\beta^{(t-1)} + \tau^{(t-1)} \sum_{m^{\prime} \in \Omega_{n \cdot}^{\text{obs}}} \left( \left( \bfnu_{m^{\prime}}^{(t-1)} \right)^2 + {\bfb^2}_{m^{\prime}}^{(t-1)} \right)}, \\
        & \bfmu_n^{(t)} = {\bfa^2}_n^{(t)} \left( \beta^{(t-1)} F^{(t-1)} \left( \bfX_{n \cdot} \right) + \tau^{(t-1)} \sum_{m^{\prime} \in \Omega_{n \cdot}^{\text{obs}}} \bfY_{n m^{\prime}} \bfnu_{m^{\prime}}^{(t-1)} \right), \\
    \end{aligned}
\end{equation}
where $\Omega_{n \cdot}^{\text{obs}}$ denotes the collection of the indices of the observed entries in the $n$-th row of $\bfY$.
Here, we can clearly see that auxiliary information has been incorporated for matrix imputation.
The posterior mean $\bfmu_n^{(t)}$ is updated as a weighted average between observed information and current prior information in $F^{(t-1)} \left( \cdot \right)$.
Importantly, the weights $\beta^{(t-1)}$ and $\tau^{(t-1)}$, as well as prior information $F^{(t-1)} \left( \cdot \right)$ are all adaptively estimated from data rather than pre-fixed.
Similarly for the $m$-th entry of $\bfw$, the posterior variance ${\bfb^2}_m^{(t)}$ and posterior mean $\bfnu_m^{(t)}$ are updated as
\begin{equation} \label{update_qw_miss}
    \begin{aligned}
        & {\bfb^2}_m^{(t)} = \frac{1}{1 + \tau^{(t-1)} \sum_{n^{\prime} \in \Omega_{\cdot m}^{\text{obs}}} \left( \left( \bfmu_{n^{\prime}}^{(t-1)} \right)^2 + {\bfa^2}_{n^{\prime}}^{(t-1)} \right)}, \\
        & \bfnu_m^{(t)} = {\bfb^2}_m^{(t)} \tau^{(t-1)} \sum_{n^{\prime} \in \Omega_{\cdot m}^{\text{obs}}} \bfY_{n^{\prime} m} \bfmu_{n^{\prime}}^{(t-1)}, \\
    \end{aligned}
\end{equation}
where $\Omega_{\cdot m}^{\text{obs}}$ denotes the collection of the indices of the observed entries in the $m$-th column of $\bfY$.
Regarding the model parameters, they can be updated as
\begin{equation} \label{update_para_miss}
    \begin{aligned}
        & \tau^{(t)} = \frac{\left| \Omega^{\text{obs}} \right|}{\left\| \calP_{\Omega^{\text{obs}}} \left( \bfY - \bfmu^{(t)} {\bfnu^{(t)}}^{\rmT} \right) \right\|_{\text{F}}^2 + \left\| \calP_{\Omega^{\text{obs}}} \left( \left( {\bfmu^{(t)}}^2 + {\bfa^2}^{(t)} \right) \left( {\bfnu^{(t)}}^2 + {\bfb^2}^{(t)} \right)^{\rmT} - {\bfmu^{(t)}}^2 \left( {\bfnu^{(t)}}^2 \right)^{\rmT} \right) \right\|_{1,1}}, \\
        & \beta^{(t)} = \frac{N}{\left\| \bfmu^{(t)} - F^{(t-1)} \left( \bfX \right) \right\|_2^2 + \left\| {\bfa^2}^{(t)} \right\|_1},
    \end{aligned}
\end{equation}
where $\calP$ is a projection operator and $\calP_{\Omega} \left( \bfY \right)$ outputs a matrix with the same dimension as that of $\bfY$
\begin{equation} \label{proj}
    \left( \calP_{\Omega} \left( \bfY \right) \right)_{nm} =
        \begin{cases}
        \bfY_{nm}, & \text{if $\left(n, m \right) \in \Omega$}, \\
        0, & \text{otherwise}.
    \end{cases}
\end{equation}
The update for the function $F \left( \cdot \right)$ remains the same as \eqref{find_f} and \eqref{shrinkage}.
As for the missing data in auxiliary information, it is clear that only the update steps involving the auxiliary matrix $\bfX$ need to be reconsidered.
Using the \textit{rpart} package \citep{rpart}, any observation with value for the dependent variable (i.e., $\bfmu_n$) and at least one independent variable (i.e., one of $\left\{ \bfX_{n1}, \dots, \bfX_{nC} \right\}$) will participate in the modeling.
For each split, the observation with the missing split variable will be split based on the best surrogate variable; if that's missing, then by the next best, and so on \citep{therneau2022introduction}.

\subsubsection{Ranking the Importance of Auxiliary Covariates}
\label{sssec:ranking}

Auxiliary covariates may not be equally important for identifying the factor $\bfz$.
In a single tree, the importance of a variable is given by the total goodness of all the splits, either as a primary or a surrogate variable.
Specifically, an overall measure of variable importance is the sum of the goodness for each split in which it was the primary variable, then plus the adjusted goodness for all splits in which it was a surrogate \citep{therneau2022introduction}.
The higher the importance value, the more the variable contributes to improving the model.
By inheriting the merit of the regression trees, the model given by MFAI can be used to rank the importance of auxiliary covariates.
Suppose the variable importance of the $c$-th covariate (i.e., $\bfX_{\cdot c}$) in the $t$-th tree (i.e., $f^{t} \left( \cdot \right)$) is $\calI_{tc}$, then the total importance score is given by
\begin{equation}\label{importance}
    \calI_{c} = \sum_{t=1}^{T} \calI_{tc},
\end{equation}
where $T$ is the total number of trees contained in the model.

\subsubsection{The Multi-Factor MFAI Model}
\label{sssec:kfac}

We now extend to fit the multi-factor MFAI model following \cite{wang2021empirical}.
To do so, we introduce the variational approximations $\left\{ q \left( \bfZ_{\cdot k} \right), q \left( \bfW_{\cdot k} \right) \right\}$ for $k = 1, \dots, K$, and then optimize $\ELBO ( q ( \bfZ_{\cdot 1}, \bfW_{\cdot 1} ), \dots, q ( \bfZ_{\cdot K}, \bfW_{\cdot K} ); \tau, \beta_1, \dots, \beta_K, F_1 ( \cdot ), \dots, F_K ( \cdot ) )$.
Similar to the single-factor case, the optimization can be done by iteratively updating parameters relating to a single factor while keeping others fixed.
The updates of a single pair $\left\{ \bfZ_{\cdot k}, \bfW_{\cdot k} \right\}$ are essentially identical to those for fitting the single-factor model aforementioned, except that $\bfY$ is replaced with the residuals obtained by removing the estimated effects of the other $K - 1$ pairs
\begin{equation}\label{res}
    \begin{aligned}
        \bfR^k = \bfY - \sum_{k^{\prime} \neq k} \widehat{\bfZ}_{\cdot k^{\prime}} \widehat{\bfW}_{\cdot k^{\prime}}^{\rmT}.
    \end{aligned} 
\end{equation}
It is worth mentioning that by doing so, we implicitly assume the full factorization form of $q$ as
\begin{equation}\label{full_fac}
    \begin{aligned}
        q \left( \bfZ, \bfW \right) = \prod_{k=1}^{K} \bfZ_{\cdot k} \prod_{k=1}^{K} \bfW_{\cdot k},
    \end{aligned} 
\end{equation}
which enjoys a fast computation speed at the cost of a slight decrease in accuracy.
We implement two algorithms for fitting the $K$-factor MFAI model: the greedy algorithm and the backfitting algorithm.
The greedy algorithm starts by fitting the single-factor model and then adds factors $k = 2, \dots , K$, one at a time, optimizing over the new factor parameters before moving on to the next factor.
The backfitting algorithm \citep{breiman1985estimating} iteratively refines the estimates for each factor given the estimates for the other factors.
In our MFAI framework, we choose to use the greedy algorithm first to provide rough estimates (for all of the variational approximations, model parameters, and unknown functions) as the initialization for the backfitting algorithm.
The greedy and backfitting algorithms are summarized in Algorithm \ref{algo_greedy} and \ref{algo_backfitting}, respectively.
\begin{algorithm}[!htb]
    \label{algo_greedy}
    \caption{Greedy Algorithm for $K$-Factor MFAI Model}
    \SetAlgoLined
    \KwData{main data matrix $\bfY$ and auxiliary matrix $\bfX$}
    \KwResult{estimate of the latent factors $\bfZ$ and loadings $\bfW$}
    \BlankLine
    \For{$k = 1, \dots, K$}{
        $\bfmu_k, a^2_k; \bfnu_k, b^2_k; \tau_k, \beta_k; F_k \left( \cdot \right) \leftarrow$ MFAI\_SF$(\bfY, \bfX)$ \;
        $\bfY \leftarrow \bfY - \bfmu_k \bfnu_k^{\rmT}$ \;
    }
    \KwRet{$\widehat{\bfZ} = \left( \bfmu_1, \dots, \bfmu_K \right); a_1^2, \dots, a_K^2; \widehat{\bfW} = \left( \bfnu_1, \dots, \bfnu_K \right); b_1^2, \dots, b_K^2$} \;
    \KwRet{$\tau_1, \dots, \tau_K; \beta_1, \dots, \beta_K$} \;
    \KwRet{$F_1 \left( \cdot \right), \dots, F_K \left( \cdot \right)$.}
\end{algorithm}
\begin{algorithm}[!htb]
    \label{algo_backfitting}
    \caption{Backfitting Algorithm for $K$-Factor MFAI Model}
    \SetAlgoLined
    \KwData{main data matrix $\bfY$ and auxiliary matrix $\bfX$}
    \KwResult{estimate of the latent factors $\bfZ$ and loadings $\bfW$}
    \BlankLine
    initialize $\left\{ \bfmu_1, a^2_1; \bfnu_1, b^2_1; \tau_1, \beta_1; F_1 \left( \cdot \right) \right\}, \dots, \left\{ \bfmu_K, a^2_K; \bfnu_K, b^2_K; \tau_1, \beta_K; F_K \left( \cdot \right) \right\}$ using greedy algorithm MFAI\_greedy$(\bfY, \bfX)$ \;
    \Repeat{convergence criterion satisfied}{
        \For{$k = 1, \dots, K$}{
            $\bfR \leftarrow \bfY - \sum_{k^{\prime} \neq k} \bfmu_{k^{\prime}} \bfnu_{k^{\prime}}^{\rmT}$ \;
            $\bfmu_k, a^2_k; \bfnu_k, b^2_k; \tau_k, \beta_k; F_k \left( \cdot \right) \leftarrow$ MFAI\_SF$(\bfR, \bfX)$ with current estimate of $\left\{ \bfmu_k, a^2_k; \bfnu_k, b^2_k; \tau_k, \beta_k; F_k \left( \cdot \right) \right\}$ as initialization \;
        }
    }
    \KwRet{$\widehat{\bfZ} = \left( \bfmu_1, \dots, \bfmu_K \right); a_1^2, \dots, a_K^2; \widehat{\bfW} = \left( \bfnu_1, \dots, \bfnu_K \right); b_1^2, \dots, b_K^2$} \;
    \KwRet{$\tau_1, \dots, \tau_K; \beta_1, \dots, \beta_K$} \;
    \KwRet{$F_1 \left( \cdot \right), \dots, F_K \left( \cdot \right)$.}
\end{algorithm}

A practical issue with matrix factorization is how to select the number of factors $K$.
Regularized methods often rely on technology such as cross-validation, which can introduce high computational costs.
In contrast, taking advantage of the additive model and the stage-wise manner to fit the $K$-factor model sequentially as shown in Algorithm \ref{algo_greedy}, MFAI can automatically determine $K$ with a little modification to the algorithm.
We first set the maximum value of the number of factors $K_{\max}$ and perform the for-loop in the greedy algorithm \ref{algo_greedy} with $K$ replaced by $K_{\max}$.
In this process, if we find the $k$-th factor/loading combination $\bfmu_k \bfnu_k^{\rmT}$ is very close to zero for one specific $k \in \left\{ 1, \dots, K_{\max} \right\}$, then we stop the fitting procedure and only use the first $k - 1$ factors as the final estimates.
This means that the users only need to set $K_{\max}$ sufficiently large, and MFAI can automatically find the suitable $K$ without excessive computational costs.
The modified algorithm is summarized in Appendix Section B.
It is worth emphasizing that our approach to determining $K$ is very similar to the methods using automatic relevancy determination (ARD) prior \citep{mackay1995probable, neal1996bayesian, tipping1999relevance}, such as \cite{babacan2012sparse}, whose key idea is that if the data are consistent with a small absolute value, then the prior precision will be estimated to be large, which results in the shrinkage of the corresponding factor/loading combinations towards zero and hence reduces the rank of the estimate.

An important feature of our MFAI approach is that it can adaptively relate auxiliary information to each factor.
The mean function $F_k \left( \cdot \right)$ and precision $\beta_k$ are specifically fitted for the $k$-th factor without parameter tuning.
In contrast, the methods that rely on cross-validation are often limited to tuning two or three parameters due to the unaffordable computational cost of searching in a large parameter space.


\section{Numerical Experiments}
\label{sec:num}

In this section, we gauge the performance of MFAI in comparison with alternative methods using both simulations and real data analyses.
As there are many methods for matrix factorization, we choose the compared methods based on two considerations.
First, they are scalable to large datasets.
Second, their software are documented and maintained well.
Based on the above criteria, we include EBMF \citep{wang2021empirical}, hardImpute, softImpute \citep{mazumder2010spectral, hastie2015matrix}, and CMF \citep{singh2008relational} in comparison.
We summarize these methods with brief descriptions in \autoref{table:compared_methods}.
\begin{table}[!htb]
    \centering
    \begin{tabular}{p{2.4cm}p{2.4cm}p{10cm}}
        \hline
        Method & R package & Brief description \\
        \hline
        \hline
        MFAI & \textit{mfair} & Variational inference, incorporates auxiliary information via gradient boosting machine \\
        EBMF & \textit{flashr} & Variational inference for a general empirical Bayes matrix factorization model \citep{wang2021empirical} \\
        hardImpute & \textit{softImpute} & Singular value decomposition \citep{mazumder2010spectral} \\
        softImpute & \textit{softImpute} & Fits a regularized low-rank matrix using a nuclear-norm penalty \citep{mazumder2010spectral, hastie2015matrix} \\
        CMF & \textit{cmfrec} & Main matrix and auxiliary matrix share the same latent factors \citep{singh2008relational, cortes2018cold} \\
        \hline
    \end{tabular}
    \caption{Overview of the compared methods.}
    \label{table:compared_methods}
\end{table}
We note that the Bayesian methods (MFAI and EBMF) are self-tuning.
The softImpute has a single tuning parameter $\lambda$ to control the nuclear norm penalty, which is chosen by cross-validation.
We apply CMF with default settings.
In the spirit of reproducibility, the source code and R scripts used to generate the results of our numerical experiments are made publicly available at \url{https://github.com/YangLabHKUST/mfair}.

\subsection{Simulation Studies}
\label{ssec:sim}

\subsubsection{Imputation Accuracy}
\label{sssec:accur}


The simulation datasets were generated as follows.
For all settings, we fixed the number of samples at $N = 1,000$, the number of features at $M = 1,000$, the number of covariates at $C = 3$, and the number of factors at $K = 3$.
The auxiliary matrix $\bfX = \left[ \bfX_{\cdot 1}, \bfX_{\cdot 2}, \bfX_{\cdot 3} \right] \in \bbR^{1,000 \times 3}$ was generated from uniform distribution $\bfX_{nc} \overset{i.i.d.}{\sim} \calU \left( -10, 10 \right)$.
Then we generated the means of three latent factors $\bfZ = \left[ \bfZ_{\cdot 1}, \bfZ_{\cdot 2}, \bfZ_{\cdot 3} \right] \in \bbR^{1,000 \times 3}$ via three functions, $F_1 \left( \bfx \right) = \frac{1}{2} x_1 - x_2$, $F_2 \left( \bfx \right) = \frac{1}{10} x_1^2 - \frac{1}{10} x_2^2 + \frac{1}{5} x_1 x_2$, and $F_3 \left( \bfx \right) = 5 \sin{\left( \frac{1}{100} x_3^3 \right)}$, respectively.
We defined the proportion of variance explained (PVE) by $F_k(\bfX)$ as $\PVE_k = \frac{ \text{Var} \left( F_k \left( \bfX \right) \right) }{ \text{Var} \left( F_k \left( \bfX \right) \right) + \beta^{-1}_k }$, and controlled $\PVE_k = 0.95$ for $k \in \left\{ 1, 2, 3 \right\}$.
Note that $F_1 \left( \cdot \right)$ is linear, while $F_2 \left( \cdot \right)$ and $F_3 \left( \cdot \right)$ are non-linear.
We designed $F_1 \left( \cdot \right)$, $F_2 \left( \cdot \right)$, and $F_3 \left( \cdot \right)$ in these forms to examine whether MFAI could flexibly incorporate auxiliary information.
The latent loading matrix $\bfW = \left[ \bfW_{\cdot 1}, \bfW_{\cdot 2}, \bfW_{\cdot 3} \right] \in \bbR^{1,000 \times 3}$ was generated from normal distribution $\bfW_{mk} \overset{i.i.d.}{\sim} \calN \left( 0, 1 \right)$.
With $\bfZ$ and $\bfW$, we obtained the true value $\bfY^\text{true} = \bfZ \bfW^{\rmT}$.
At last, we added noises to simulate noisy observation $\bfY = \bfY^\text{true} + \bfepsilon$, where $\bfepsilon_{nm} \sim N(0, \tau^{-1})$.
Then the PVE by the factors is defined as $\PVE = \frac{ \text{Var} \left( \bfY^\text{true} \right) }{ \text{Var} \left( \bfY^\text{true} \right) + \tau^{-1} }$.
To mimic the real data with a partially observed main matrix, we randomly masked a subset of entries of $\bfY$ and denoted their index set as $\Omega^\text{miss}$.
The remaining entries were considered as observed entries with index set $\Omega^\text{obs}$.
The missing ratio can then be computed as $\frac{ \left| \Omega^{\text{miss}} \right| }{ NM }$.

To evaluate the imputation accuracy, we used half of the observed entries in $\Omega^\text{obs}$ as the training data and the remaining entries as the test data with index set $\Omega^\text{train}$ and $\Omega^\text{test}$, respectively.
Then, we applied matrix factorization methods to $\bfY^{\text{train}}$ and obtained the output $\widehat{\bfY}$.
The imputation accuracy can be measured by root-mean-square error (RMSE) on the test set $\Omega^\text{test}$
\begin{equation}\label{rmse_imputation}
    \begin{aligned}
        \text{RMSE} \left( \widehat{\bfY}, \bfY\right) = \sqrt{\frac{\sum_{(n,m) \in \Omega^{\text{test}}} \left( \widehat{\bfY}_{nm} - \bfY_{nm} \right)^2}{ \left| \Omega^{\text{test}} \right|}}.
    \end{aligned}
\end{equation}
We designed two sets of experiments to test the methods in a wide range of data quality settings.
In Experiment 1, we fixed the missing ratio, $\frac{ \left| \Omega^{\text{miss}} \right| }{ NM } = 0.5$ and varied $\PVE \in \{0.1, 0.5, 0.9\}$ to examine the performances of the compared methods under different noise levels.
In Experiment 2, we fixed the $\PVE = 0.5$ and varied $\missing \in \{0, 0.5, 0.9\}$ to investigate the influence of data sparsity levels.
We specified the number of factors for all methods to be the true value $K=3$ and repeated the simulations $50$ times for each setting.
The greedy algorithm of our MFAI with default parameter setting took about one minute, and the continuing backfitting algorithm took a few more seconds for each experiment using four CPU cores of Intel(R) Xeon(R) Gold 6230N CPU @ 2.30GHz processor on a Linux computing platform.

\begin{figure}[!htb]
    \begin{center}
        \includegraphics[width=6in]{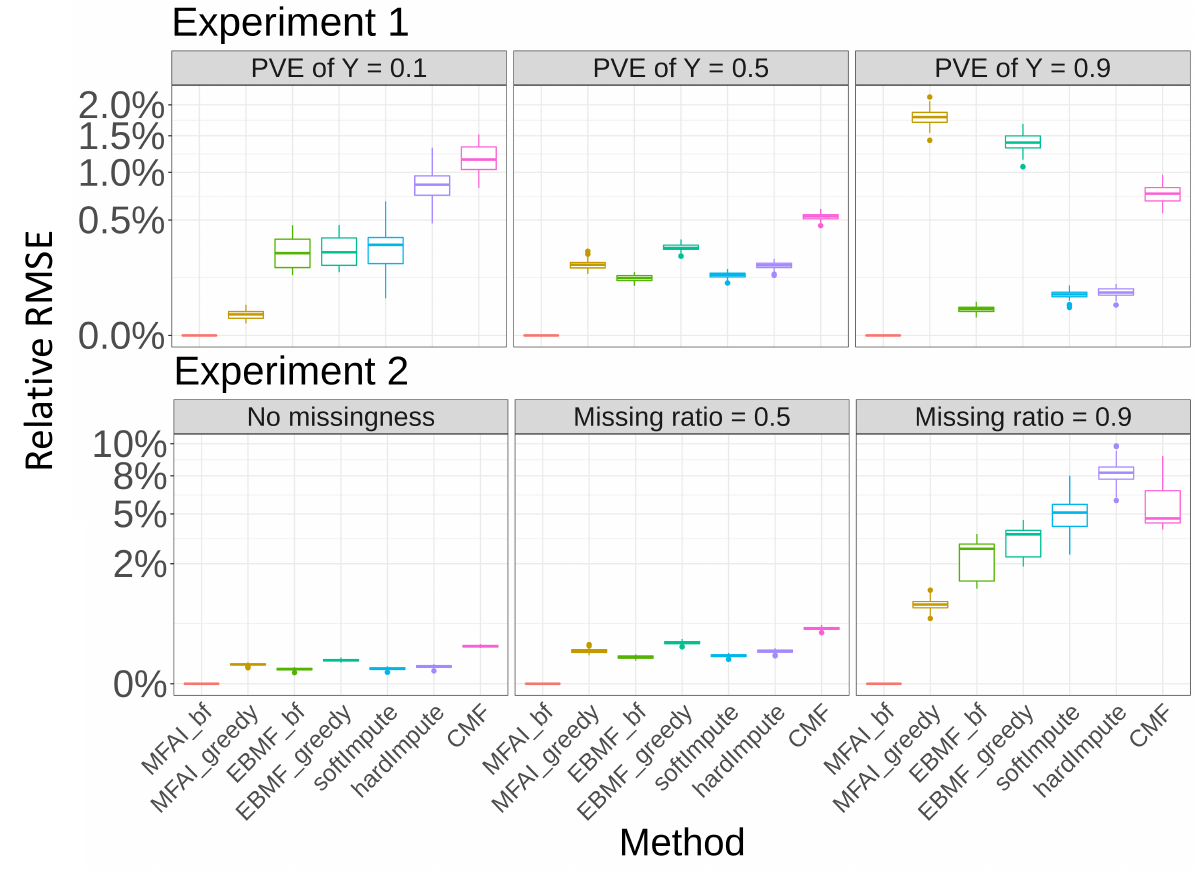}
    \end{center}
    \caption{Boxplots comparing the accuracy of different methods. Experiment 1 involves the main matrix $\bfY$ that varies from the weak signal ($\PVE = 0.1$, left) to the strong signal ($\PVE = 0.9$, right). Experiment 2 involves the main matrix $\bfY$ that varies from low sparsity ($\missing = 0$, left) to high sparsity ($\missing = 0.9$, right). Accuracy is measured by the difference in each method's RMSE from the MFAI's RMSE, then divided by the MFAI's RMSE, with smaller values indicating higher accuracy. The $y$ axis is plotted on the square-root scale to avoid the plots being dominated by methods performed poorly. \label{fig:sim}}
\end{figure}

We summarized the relative RMSE of alternative methods to MFAI in the first row of \autoref{fig:sim} for Experiment 1.
Our MFAI method with backfitting achieved the best accuracy in all parameter settings.
When the signal was strong, the RMSE of EBMF, hardImpute, and softImpute were slightly higher than MFAI, while the advantage of MFAI became more evident as the PVE decreased due to its ability to incorporate auxiliary information.
Although CMF can also do that, it generally performed poorly in this simulation setting because it can use only the linear model.
Furthermore, the parameter tuning of CMF is not adaptive, and its default parameter setting may not be suitable for the simulation study.
Overall, these results suggest that MFAI can effectively leverage the non-linear relationship between auxiliary information and the main matrix to improve imputation accuracy at different levels of signal strength.


In Experiment 2, as shown in the second row of \autoref{fig:sim}, the relative performance of alternative methods to MFAI highly depends on the data sparsity.
When the main matrix $\bfY$ was highly sparse ($\missing = 0.9$), there was little room for improvement if only $\bfY$ was available.
The MFAI approach achieved greater improvement over other methods by effectively incorporating auxiliary information in a stable manner.

To summarize, MFAI can significantly outperform other approaches when the data quality is poor, such as with weak signals and high sparsity.
When the data quality is relatively good, it still retains its competitiveness and achieves slight but steady gains.
The superior performance of MFAI can be attributed to the following facts:
First, MFAI enables a flexible non-linear model to incorporate auxiliary information. 
Second, our algorithm design seamlessly combines the advantage of the EM algorithm and gradient boosting, making MFAI stable and adaptive.

\subsubsection{Robustness}
\label{sssec:rob}

\begin{figure}[!htb]
    \begin{center}
        \includegraphics[width=6in]{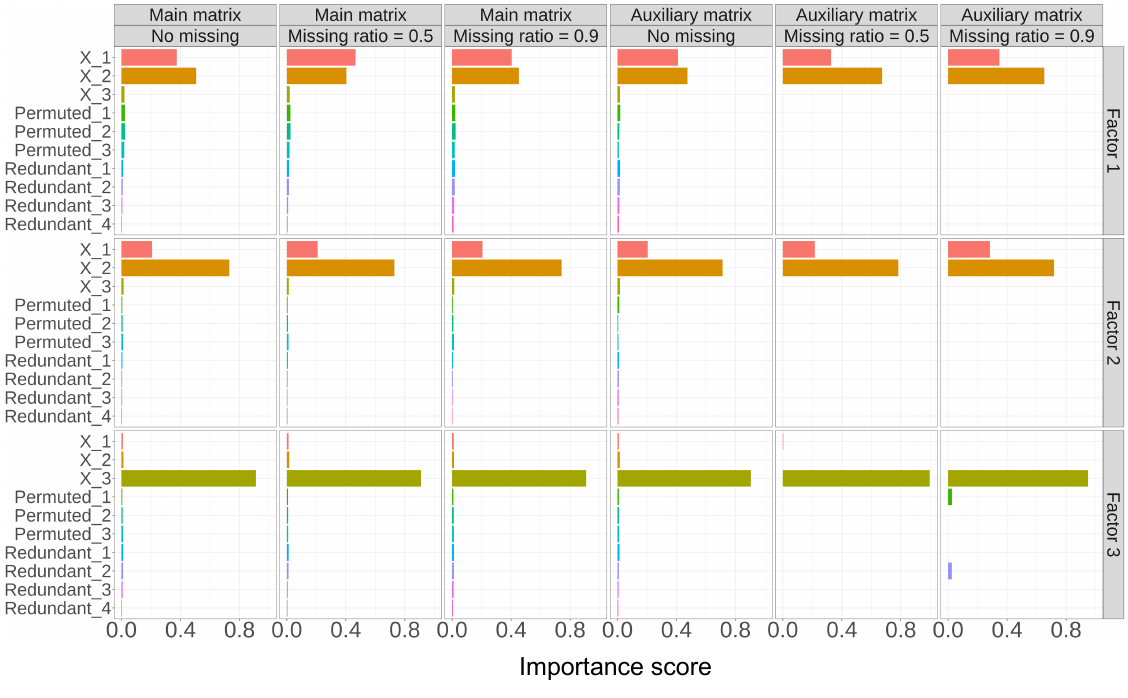}
    \end{center}
    \caption{Barplots for the importance scores of the auxiliary covariates in Factor 1-3. In the left panel (first three columns), we masked the main matrix $\bfY$ randomly and varied from the low sparsity ($\missing = 0$, left) to high sparsity ($\missing = 0.9$, right). In the right panel (next three columns), we first fixed the missing ratio of $\bfY$ as $0.5$, and further masked the auxiliary matrix $\bfX^\text{all}$ randomly and varied from the low sparsity ($\missing = 0$, left) to high sparsity ($\missing = 0.9$, right). The importance scores in each factor have been re-scaled to have a sum of one. The higher the importance score, the more the specific covariate contributes to improving the model. \label{fig:sim_imp}}
\end{figure}

To exhibit MFAI's ability to distinguish useful auxiliary covariates from irrelevant ones, we rank the importance of the covariates based on regression trees \autoref{fig:sim_imp}), which has been defined in Section \ref{sssec:ranking}.
We already have the auxiliary matrix $\bfX \in \bbR^{1,000 \times 3}$ and the main matrix $\bfY \in \bbR^{1,000 \times 1,000}$ with $\PVE = 0.5$, which were generated as described in Section \ref{sssec:accur}.
These three covariates in $\bfX$ were known to be important.
To introduce irrelevant covariates, we included three covariates by permuting the rows of $\bfX$ and four additional redundant variables from the uniform distribution $\calU \left( -10, 10 \right)$, denoted as $\bfX^{\text{pmt}}\in \mathbb{R}^{1,000\times 3}$ and $\bfX^{\text{rdd}} \in \bbR^{1,000 \times 4}$, respectively.  
At last, we combined the three auxiliary matrices column-wise and got $\bfX^\text{all} = \left[ \bfX, \bfX^\text{pmt}, \bfX^\text{rdd} \right] \in \bbR^{1,000 \times 10}$, in which the first three columns were useful auxiliary covariates and the remaining seven columns were useless.
We applied MFAI to $\bfY$ and $\bfX^\text{all}$ in different situations and visualized the importance scores of all the auxiliary covariates in the top three factors.
In the left panel of \autoref{fig:sim_imp}  (first three columns), we masked the main matrix $\bfY$ randomly and varied the missing ratio.
In the right panel of \autoref{fig:sim_imp} (next three columns), we fixed the missing ratio of $\bfY$ at $0.5$, and further masked the auxiliary matrix $\bfX^\text{all}$ randomly at the different missing levels.
\autoref{fig:sim_imp} shows that those unimportant auxiliary covariates get nearly zero importance scores under all data sparsity settings, which indicates that MFAI can effectively distinguish those useful auxiliary covariates, even though the datasets were highly sparse.

\subsubsection{Computational Efficiency}
\label{sssec:comp}

\begin{figure}[!htb]
    \begin{center}
        \includegraphics[width=6in]{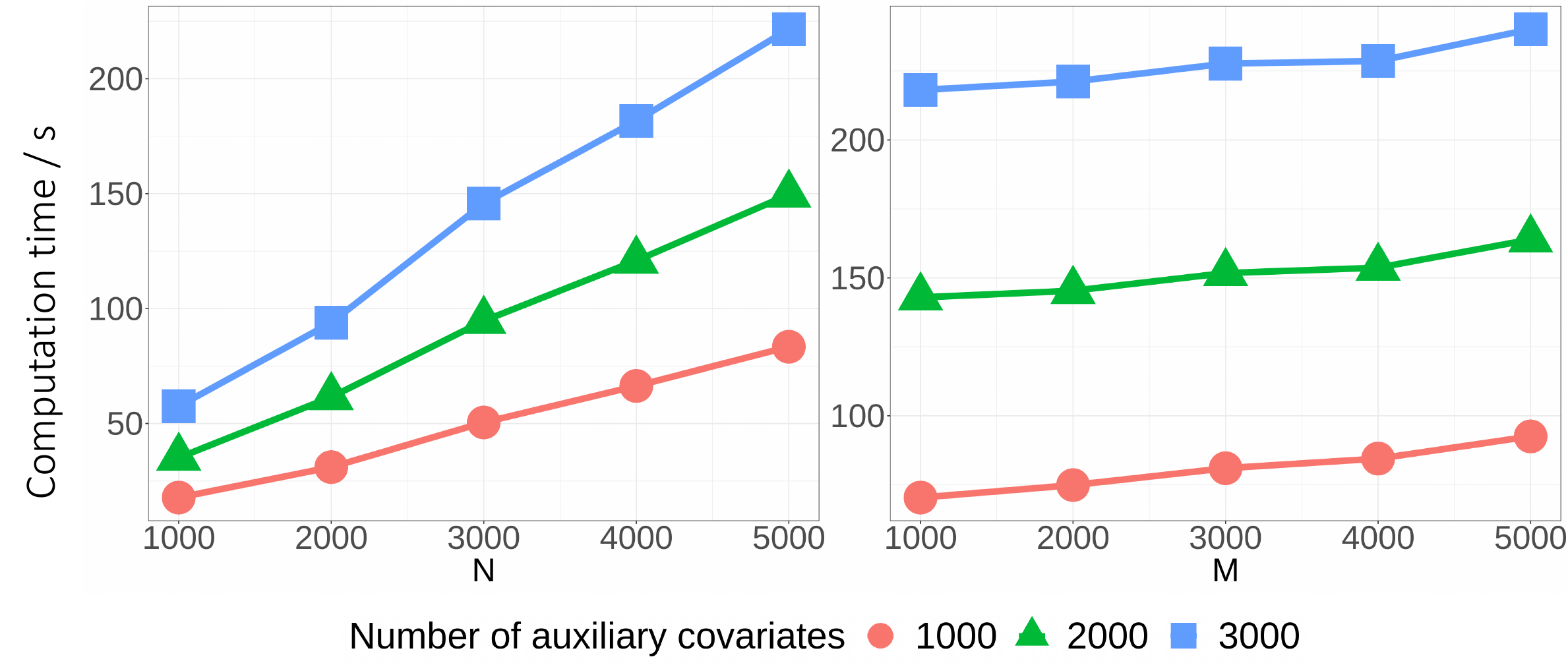}
    \end{center}
    \caption{Lineplots for the computation timings against data size. In the left panel, we fixed feature size $M$ and varied sample size $N$. In the right panel, we fixed sample size $N$ and varied feature size $M$. Different shapes of the points and colors of the lines represent the different numbers of auxiliary covariates (i.e., $C$) used in the model, respectively. \label{fig:sim_time}}
\end{figure}

Finally, we show the computational efficiency of MFAI.
We first fixed the sample size $N = 5,000$ and varied the number of features $M \in \{1,000, \ 2,000, \ 3,000, \ 4,000, \ 5,000\}$ (the left panel of \autoref{fig:sim_time}), and then fixed $M = 5,000$ and varied $N \in \{1,000, \ 2,000, \ 3,000, \ 4,000, \ 5,000\}$ (the right panel of \autoref{fig:sim_time}).
To evaluate the computation time of MFAI, we applied the single-factor MFAI and fixed the number of iteration steps as the same value $20$.
Furthermore, the experiments were repeated with different numbers of auxiliary covariates, $C \in \{ 1,000, \ 2,000, \ 3,000 \}$, indicated by different colors in \autoref{fig:sim_time}.

\subsection{Real Data Analyses}
\label{ssec:real}


\subsubsection{Data Description and Methods Setup}
\label{sssec:des_real}

The two real datasets used in this section are as follows:

\textit{MovieLens 100K data}\footnote{\url{https://movielens.org/}} is an extensively studied dataset to evaluate the performance of the recommender system \citep{harper2015movielens}.
The main matrix $\bfY$ is a $1,682 \times 943$ matrix of ratings (0–5 star rating), where each row represents a movie and each column represents a user.
However, as mentioned in Section \ref{sec:intro}, most users only rate a small number of movies, making the matrix extremely sparse ($\missing = 94\%$), containing only about 100K observed ratings.
The auxiliary matrix $\bfX$ is a $1,682 \times 18$ binary matrix of movie genres, where each row represents a movie and each column represents a genre.
The $18$ genres are ``Action'',  ``Adventure'', ``Animation'', ``Children's'', ``Comedy'', ``Crime'', ``Documentary'', ``Drama'', ``Fantasy'', ``Film-Noir'', ``Horror'', ``Musical'', ``Mystery'', ``Romance'', ``Sci-Fi'', ``Thriller'', ``War'', and ``Western''.
In order to enhance the matrix factorization performance, we utilize this portion of the data based on the presumption that movies of the same genres should be rated similarly in some sense.
We masked a fraction of the observed entries and evaluated the accuracy of different methods in predicting the masked entries.

\textit{Human brain gene expression data}\footnote{\url{https://hbatlas.org/}} is a fully observed (i.e., $\missing = 0$) matrix of bulk gene expression (microarray platform) from the human brain transcriptome (HBT) project, where the expression levels of $17,568$ genes are measured in 16 brain regions across 15 time periods \citep{johnson2009functional, kang2011spatio}.
We applied the following pre-processing procedure to this dataset.
First, periods 1 and 2 correspond to embryonic and early fetal development when most of the 16 brain regions sampled in future periods have not differentiated.
Therefore, data in periods 1 and 2 were excluded from our analysis.
Second, we focused on analyzing the neocortex areas, including 11 brain regions.
Then, the main matrix $\bfY$'s dimension became $886 \times 17,568$, where each row represents a bulk tissue sample, and each column represents a gene.
Recent studies have shown that region and age (that is, spatial-temporal dynamics) contribute more to the global differences in gene expression than do other variables: sex, ethnicity, and inter-individual variation \citep{kang2011spatio}.
Hence, we extracted each sample's brain region and time period information from the raw data as auxiliary information.
The auxiliary information $\bfX$ here is represented as an $886 \times 2$ data frame.
The first column is a vector of factor types indicating which region the sample belongs to.
The second column is a vector containing the time period information.
We provide more detailed information about neocortex areas and time periods in Appendix Section C.1.

Both Bayesian methods, MFAI and EBMF can automatically estimate $K$, and we set $K_{\text{max}} = 20$ for MovieLens 100K data and $K_{\text{max}} = 150$ for human brain gene expression data, which are sufficiently large.
For softImpute, hardImpute, and CMF, we specified $K$ based on the values inferred by MFAI.
We first compared the imputation accuracy in terms of the RMSE and then demonstrated that MFAI could illuminate the logic of how the auxiliary information relates to the main data matrix by investigating the inferred factors and loadings.

\subsubsection{Imputation Accuracy}

\begin{figure}[!htb]
    \begin{center}
        \includegraphics[width=6in]{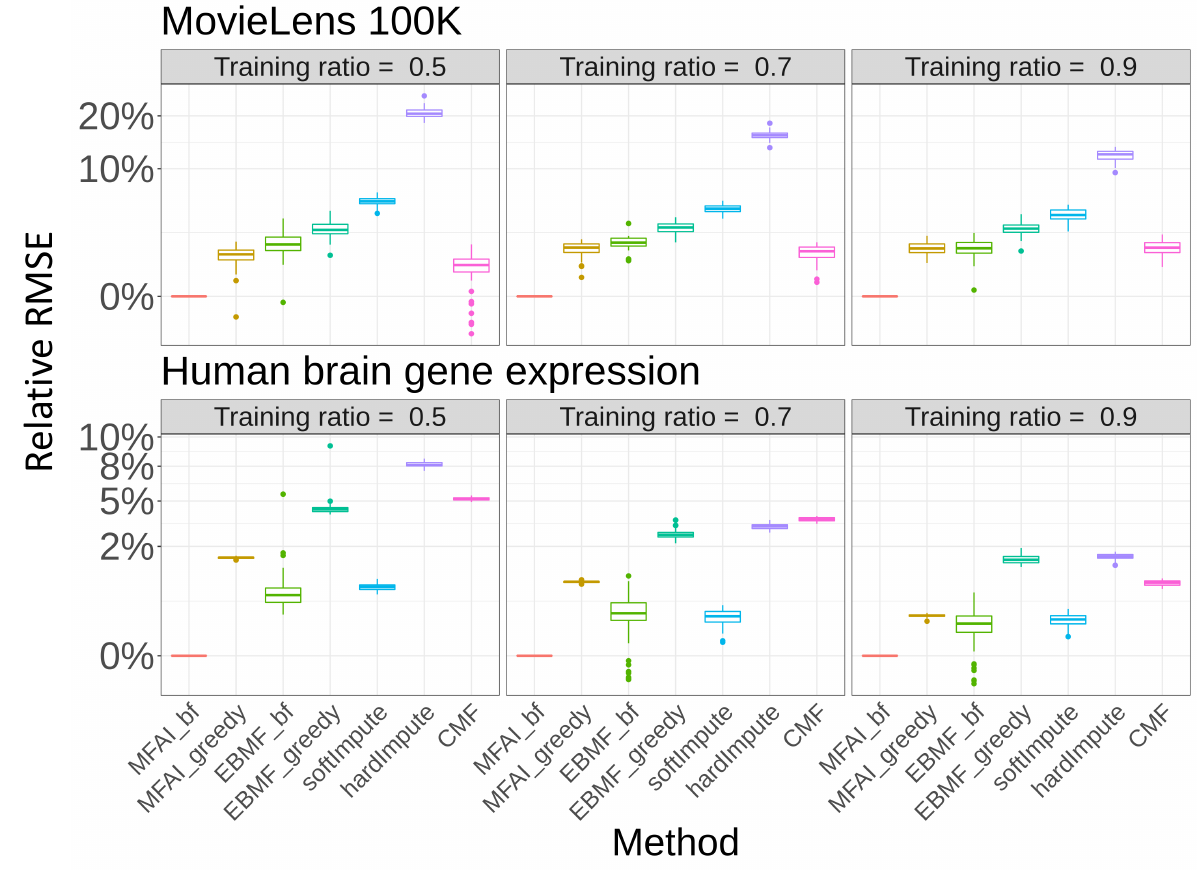}
    \end{center}
    \caption{Boxplots comparing the accuracy of different methods in imputing missing entries. These two sets of experiments involve the main matrix $\bfY$ that varies from rare information ($\text{training ratio} = 0.5$, left) to rich information ($\text{training ratio} = 0.9$, right). Accuracy is measured by the difference in each method's RMSE from the MFAI's RMSE, then divided by the MFAI's RMSE, with smaller values indicating higher accuracy. The $y$ axis is plotted on the square-root scale to avoid the plots being dominated by methods performed poorly. \label{fig:real_rmse}}
\end{figure}

In this section, we examined the imputation performance of compared methods (\autoref{fig:real_rmse}).
First, we randomly split the observed entries $\Omega^{\text{obs}}$ into a training set $\Omega^{\text{train}}$ and a test set $\Omega^{\text{test}}$.
Then, we applied matrix factorization methods to the training data and predicted the entries in the held-out set.
The imputation accuracy of the held-out entries was measured by RMSE \eqref{rmse_imputation}. 
We considered different values of the ``training ratio'' which is defined as $\frac{ |\Omega^{\text{train}}| }{ |\Omega^{\text{obs}}| }$.
For human brain gene expression data, $\Omega^{\text{obs}} = \Omega$ since it is fully observed.
We repeated the experiments 50 times for each setting of the training ratio.
MFAI used only around two minutes to analyze MovieLens 100K data with inferred $K = 9$ and around 150 minutes to analyze human brain gene expression data with inferred $K = 95$, using four CPU cores of Intel(R) Xeon(R) Gold 6230N CPU @ 2.30GHz processor on a Linux computing platform.
By contrast, EBMF, another Bayesian method that cannot incorporate auxiliary information, used around one minute to analyze MovieLens 100K data and around 130 minutes to analyze human brain gene expression data using the same computing resources, suggesting that MFAI can leverage auxiliary information with only minor computational overhead.
We summarized the RMSE across 50 times experiments in \autoref{fig:real_rmse}.
For the MovieLens 100K data, MFAI and CMF outperformed other methods by incorporating the movie genre information, suggesting the movie genre provides useful information to predict user ratings.
MFAI gained greater improvement from the movie genre information than CMF because the gradient boosted tree offers a more flexible structure than the linear model in CMF to characterize the connection between the factor and genre of a movie.
The auxiliary information of the human brain gene expression data comes from two different sources: regions and time periods, where regions are represented as categorical variables and time periods are represented as numerical variables.
MFAI also achieved the best performance among the compared methods because the tree structure in MFAI is very good at handling mixed data types (i.e., categorical and numerical variables).
In contrast, CMF did not perform well in this dataset, which may be attributed to the fact that the linear models are often not good at handling mixed data types and capturing possible spatial-temporal interaction effects in the gene expression data.
These evidence suggests that MFAI can effectively leverage auxiliary information to improve the imputation accuracy in the highly sparse dataset by taking advantage of the gradient boosted tree structure.

\subsubsection{Enrichment of Movie Genres in MovieLens 100K Data Analysis}
\label{sssec:mov}

\begin{figure}[!htb]
    \begin{center}
        \includegraphics[width=6in]{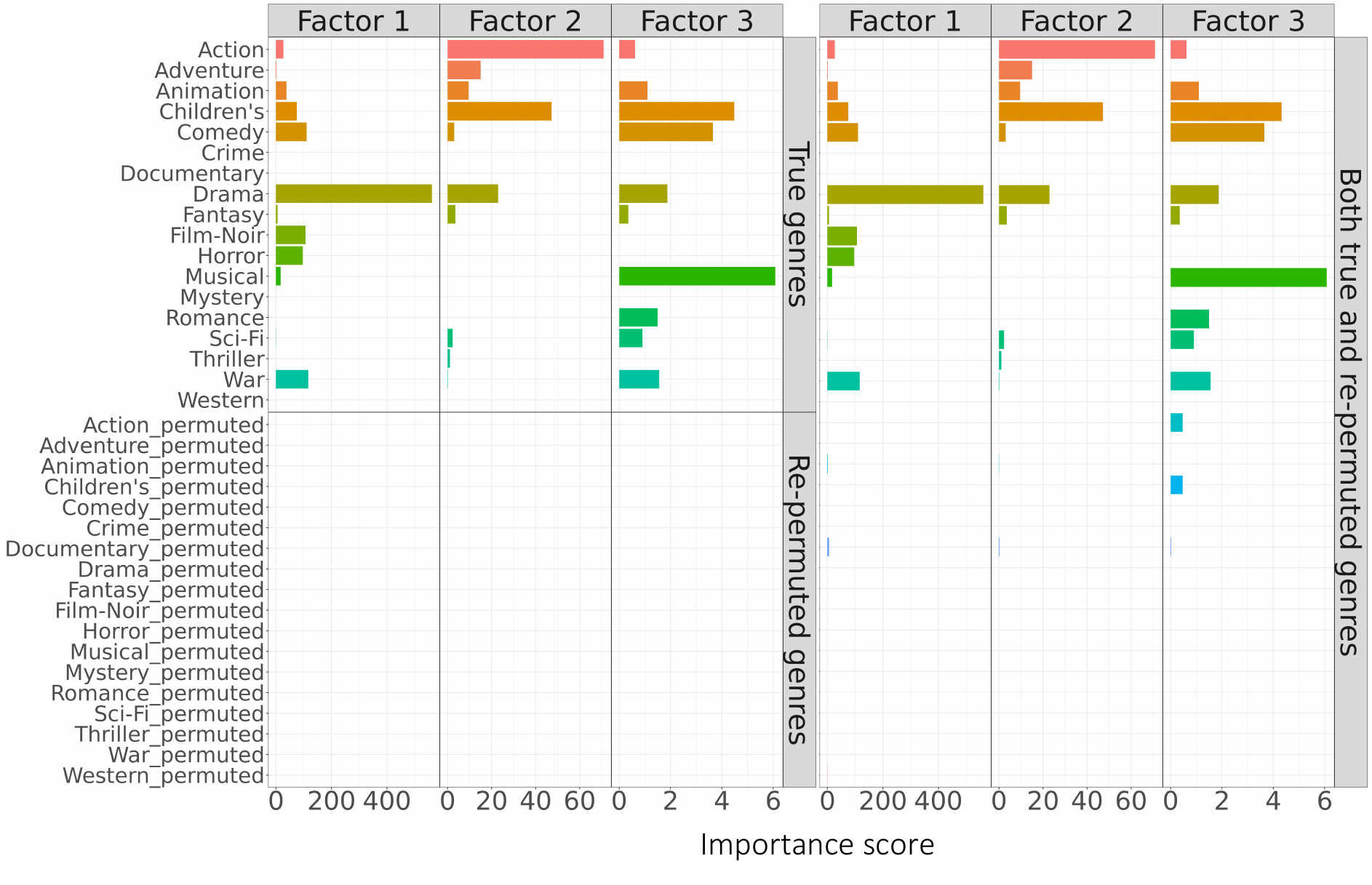}
    \end{center}
    \caption{Barplots for the importance scores of the auxiliary covariates in Factor 1-3 of the MovieLens 100K data. In the top left panel, we only used the true movie genre information $\bfX$ as the input. In the bottom left panel, we only used the re-permuted movie genre information $\bfX^{\text{pmt}}$ as the input. In the right panel, we used both the true and re-permuted movie genre information $\bfX^{\text{pmt}}$ as the input. The higher the importance score, the more a specific movie genre contributes to improving the model. \label{fig:movie_imp}}
\end{figure}

In this section, we use MovieLens 100K data to illustrate the ability of MFAI to identify important variables in auxiliary information through decision trees, which allows us to gain a deeper understanding of the connection between the main matrix and the auxiliary information (\autoref{fig:movie_imp}).
As a negative control, we constructed a permuted movie genre matrix $\bfX^{\text{pmt}} \in \bbR^{1,682 \times 18}$, where the $c$-th column $\bfX^{\text{pmt}}_{\cdot c}$ was obtained by permuting the entries of $\bfX_{\cdot c}$ for $c = 1, \dots, 18$. 
We applied MFAI to the whole MovieLens 100K data with three different auxiliary matrices $\bfX$, $\bfX^{\text{pmt}}$, and $\bfX^{\text{both}}=\left[ \bfX, \bfX^{\text{pmt}} \right] \in \bbR^{1,682 \times 36}$.
In each experiment, we initialized the model parameters with the same values and computed the importance scores of auxiliary covariates.

\autoref{fig:movie_imp} visualizes the importance scores of auxiliary covariates in the top three factors.
The top left panel shows the importance scores obtained with $\bfX$, which indicates the relevance of true movie genres to user ratings.
We can see that in Factor 1, ``Drama'' is the leading genre while ``Children's'', ``Comedy'', ``Film-Noir'', ``Horror'', and ``War'' also contribute to this factor; ``Action'' and ``Children's'' are the two major genres in Factor 2; ``Musical'', ``Children's'', and ``Comedy'' exert influence in Factor 3.
When using the permuted matrix $\bfX^{\text{pmt}}$ as input (bottom left panel of \autoref{fig:movie_imp}), MFAI correctly assigned low importance scores to all permuted genres, suggesting that MFAI avoids incorporating irrelevant auxiliary information.
Finally, in the presence of both true and permuted movie genres (right panel of \autoref{fig:movie_imp}), MFAI successfully distinguished the useful movie genres from irrelevant ones.
By comparing the left panels and the right panel of \autoref{fig:movie_imp}, we can observe that the importance scores obtained using $\bfX^{\text{both}}$ are highly consistent with those obtained using $\bfX$ and $\bfX^{\text{pmt}}$ as separate inputs, indicating the stability and robustness of MFAI.

\subsubsection{Spatial and Temporal Dynamics of Gene Regulation Among Tissues}
\label{sssec:gene}

The spatial and temporal patterns of gene regulation during brain development have attracted a great deal of attention in the neuroscience community.
The availability of gene expression profiles collected from multiple brain regions and time periods provides an unprecedented chance to characterize human brain development.
Despite the availability of rich resources, turning such a wealth of data into knowledge about the brain requires a lot of modeling effort.
The aforementioned human brain gene expression data has been analyzed by several statistical methods, e.g., \citet{lin2015markov} modeled spatial and temporal patterns with Markov random field (MRF), \citet{lin2017joint} proposed a Bayesian neighborhood selection method to estimate the network structure, and \citet{liu2022characterizing} developed a low-rank tensor decomposition method to capture spatial and temporal effects simultaneously.
By modeling the relationship between the spatial-temporal information and the gene expression matrix via non-linear functions $F_k \left( \cdot \right)$, MFAI can offer biological insights into the heterogeneity in temporal dynamics across different brain regions and the evolution of spatial patterns over multiple time periods.
Following \citet{hawrylycz2015canonical}, we selected genes with consistent spatial patterns across individuals using the concept of differential stability (DS), which was defined as the tendency for a gene to exhibit reproducible differential expression relationships across brain structures (see more details in Appendix Section C.2).
As inputs to MFAI, we included 2,000 genes with the highest DS, resulting in the new main matrix $\bfY \in \bbR^{886 \times 2,000}$, and used the same auxiliary $\bfX \in \bbR^{886 \times 2}$ with spatial and temporal information.

\begin{figure}[!htb]
    \begin{center}
        \includegraphics[width=0.85\textwidth]{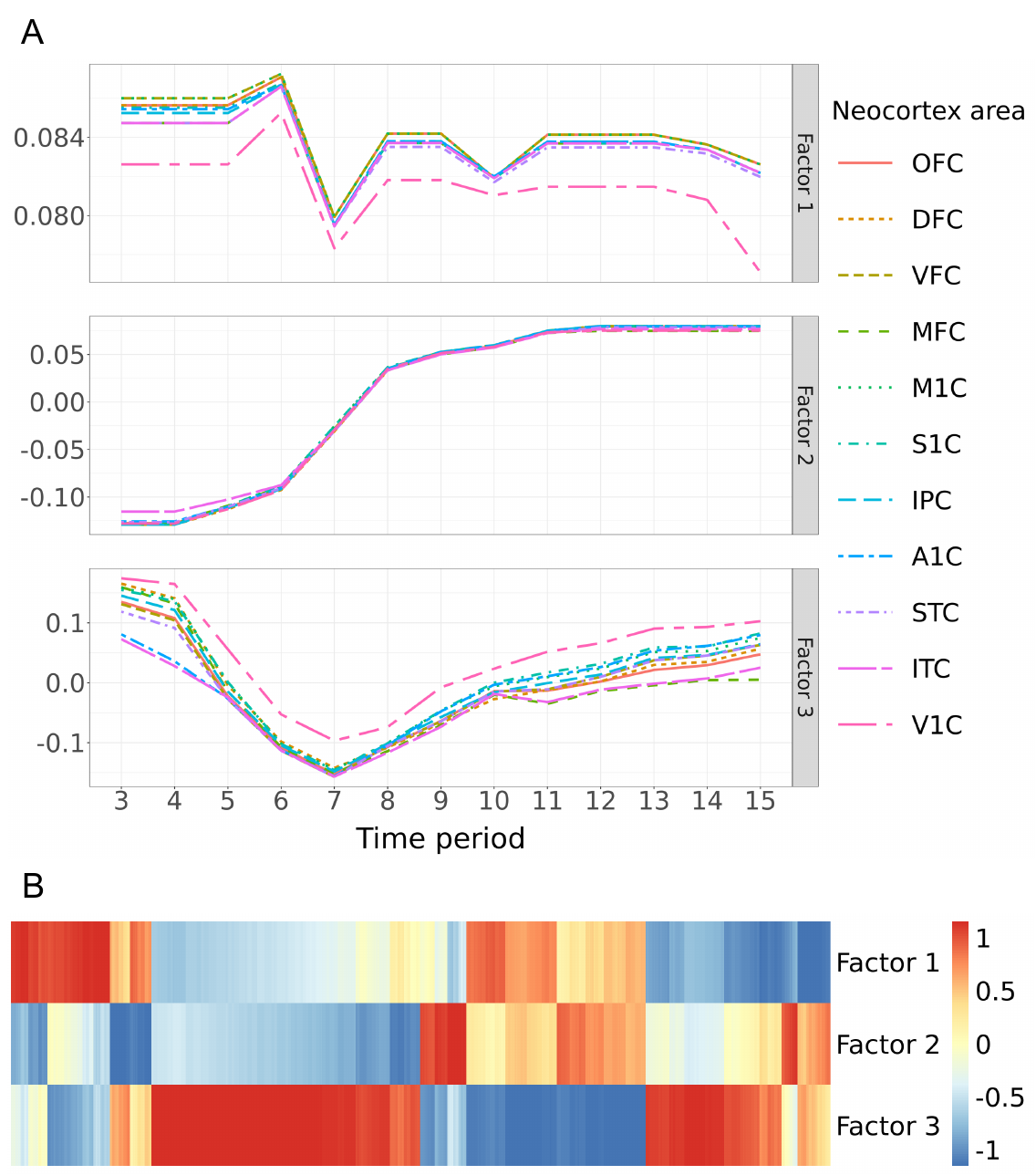}
    \end{center}
    \caption{Spatial-temporal dynamic patterns in Factor 1-3. Figure A shows the normalized factor levels across different neocortex areas and time periods of the top three factors (i.e., $\left\{ F_1 \left( \cdot \right), F_2 \left( \cdot \right), F_3 \left( \cdot \right) \right\}$). Figure B is the heatmap of the corresponding top three loadings (i.e., $\left\{ \bfW_{\cdot 1}, \bfW_{\cdot 2}, \bfW_{\cdot 3} \right\}$), where each column represents a gene. The gene loadings have been normalized before visualization. \label{fig:brain}}
\end{figure}

To gain insights, the dynamic patterns of the top three factors across different neocortex areas and time periods, represented by fitted functions $\left\{ F_1 \left( \cdot \right), F_2 \left( \cdot \right), F_3 \left( \cdot \right) \right\}$, are given in \autoref{fig:brain} A.
Each factor has been normalized to have the $\ell^2$-norm equal one.
It is obvious that fitted functions not only capture the non-linearity across different time periods but also implicate spatial-temporal interactions.
As such, MFAI fully made use of the auxiliary information.
In contrast, methods that only consider simple linear relationships may cause information loss.
Overall, all three factors show stronger temporal differences compared to spatial differences within the neocortex areas.
The temporal trajectories of all three factors show clear signs of prenatal development (from Period 3 to Period 7).
From infancy (Period 8 and afterward), Factor 2 exhibits increasing influence, while Factor 3 exhibits decreasing influence in magnitude.
Then, all three factors maintain steady levels until late adulthood.
All the non-V1C neocortex areas show particularly pronounced correlations and consistency during development.
Factor 1 and Factor 3 in V1C showed distinctive signals throughout development and adulthood, compared to other neocortex areas.

\autoref{fig:brain} B is the heatmap of the top three inferred gene loadings $\left[ \bfW_{\cdot 1}, \bfW_{\cdot 2}, \bfW_{\cdot 3} \right] \in \bbR^{2,000 \times 3}$.
To understand these three loadings better, we conducted the gene set enrichment analysis based on Gene Ontology (\url{http://geneontology.org/}) \citep{ashburner2000gene, gene2023gene, thomas2022panther}.
Specifically, we first calculated the relative weight of the $k$-th loading for the $m$-th gene by $\frac{ \left| \bfW_{mk} \right| }{ \sum_{k^{\prime}=1}^{3} \left| \bfW_{mk^{\prime}} \right| }$, and then selected the top 300 weighted genes in each loading to form the gene sets.
The enriched biological processes with corresponding $p$-values after Bonferroni correction are summarized in \autoref{table:gene_enrichment}.
\begin{table}[!htb]
    \centering
    \begin{tabular}{|c|c|c|}
        \hline
        & Biological process & $P$-value with Bonferroni correction \\
        \hline
        \hline
        \multirow{3}{*}{Loading 1} & Axon development & $1.97 \times 10^{-2}$ \\
        & Neuron development & $1.82 \times 10^{-3}$ \\
        & Neuron differentiation & $8.03 \times 10^{-5}$ \\
        \hline
        \multirow{5}{*}{Loading 2} & Regulation of biological quality & $2.25 \times 10^{-10}$ \\
        & Potassium ion transmembrane transport & $6.25 \times 10^{-5}$ \\
        & Regulation of transport & $2.84 \times 10^{-7}$ \\
        & Signaling & $5.35 \times 10^{-5}$ \\
        & Cell communication & $1.02 \times 10^{-4}$ \\
        \hline
        \multirow{4}{*}{Loading 3} & Regulation of cell junction assembly & $4.04 \times 10^{-3}$ \\
        & Cell adhesion & $8.07 \times 10^{-4}$ \\
        & Cell junction assembly & $3.84 \times 10^{-4}$ \\
        & Nervous system development & $5.58 \times 10^{-5}$ \\
        \hline
    \end{tabular}
    \caption{Gene enrichment analysis on Loading 1-3.}
    \label{table:gene_enrichment}
\end{table}
Loading 1 relates to axon and neuron development, consistent with its status as the leading factor in the neocortex and relatively high signal level across all time periods, as shown in \autoref{fig:brain} A.
Loading 2 is enriched in signaling \citep{luebke2003aging, luebke2004normal} and cell communication \citep{lopez2013hallmarks}, which are aging-related processes.
It is also known that aging induces specific changes in individual ATPases according to their subsynaptic localization \citep{de2014brain}.
For example, $\text{Na}^{+}$, $\text{K}^{+}$-ATPase activity in the hippocampus tends to decrease by age \citep{kinjo2007na}, consistent with the temporal pattern of Factor 2 shown in \autoref{fig:brain} A.
Combining \autoref{fig:brain} A and \autoref{fig:brain} B, the enrichment of Loading 2 in the ion transport provides evidence that the interstitial ion is a key regulator of state-dependent neural activity \citep{rasmussen2020interstitial}.
Loading 3 is mainly enriched in the cell junction, which plays an important role during the development of the mammalian brain \citep{montoro2004gap}.
In the neocortex, gap junctions are already expressed at very early stages of development and are involved in many processes like neurogenesis, migration, and synapse formation \citep{sutor2005involvement}.
In the mammalian central nervous system (CNS), coupling of neurons by gap junctions (i.e., electrical synapses) and the expression of the neuronal gap junction protein, connexin 36 (Cx36), transiently increases during early postnatal development, then subsequently declines and remains low in adulthood, confined to specific subsets of neurons \citep{belousov2013neuronal}.
This trend is highly consistent with the temporal pattern of Factor 3 shown in \autoref{fig:brain} A, reaching a brief high magnitude around birth and quickly falling back.

\section{Discussion}
\label{sec:conc}

The auxiliary information is particularly useful to improve matrix factorization when the observed matrix is noisy and sparse.
Despite the massive availability of auxiliary information in real-world applications, existing methods largely rely on linear models to connect auxiliary covariates with the main matrix, leading to suboptimal performance.
In this article, we propose a scalable Bayesian matrix factorization approach named MFAI to leverage auxiliary information.
By integrating the gradient boosted trees with probabilistic matrix factorization, MFAI enables nonlinear modeling of auxiliary covariates and allows the model parameters to be automatically estimated under the empirical Bayes framework, making MFAI adaptive to the complicated connections between the main matrix and auxiliary information.
Besides, MFAI naturally inherits several salient features of gradient boosted trees, such as robustness to irrelevant features, immunity to missing values in predictors, and the ability to distinguish useful covariates in the auxiliary information.
Under the variational assumption, we developed an efficient algorithm that can simultaneously estimate the number of latent factors and infer the low-rank structure.
With our innovations in the model and algorithm designs, our \textit{mfair} software is effective, stable, and scalable to large datasets.
Through comprehensive simulation studies, we showed that MFAI is statistically more accurate than alternative methods, especially in the scenario of high sparsity and weak signal strength.
We applied MFAI to two real datasets, including one human brain gene expression dataset, and showed that MFAI not only improves the imputation accuracy but also yields biological insights into spatiotemporal gene regulation patterns in the human brain.

Taking advantage of probabilistic modeling, the MFAI can be modified easily to further introduce other properties, such as sparsity through spike and slab prior or automatic relevancy determination (ARD) prior \citep{mackay1995probable, neal1996bayesian, tipping1999relevance} used in Sparse Factor Analysis (SFA) \citep{engelhardt2010analysis}.
Another potential extension is to incorporate auxiliary information not only of the samples to characterize the factors but also of the features to help identify the loadings.


\section*{Acknowledgements}

This work is supported in part by Hong Kong Research Grant Council Grants 16307818, 16301419, 16308120, and 16307221; Hong Kong Innovation and Technology Fund Grant PRP/029/19FX; The Hong Kong University of Science and Technology Startup Grants R9405 and Z0428 from the Big Data Institute; and City University of Hong Kong Startup Grant 7200746.
The computation tasks for this work were performed using the X-GPU cluster supported by the Research Grants Council Collaborative Research Fund Grant C6021-19EF.

\section*{Disclosure Statement}

The authors report there are no competing interests to declare.

\bigskip
\begin{center}
{\large\bf SUPPLEMENTARY MATERIALS}
\end{center}

\begin{description}

\item[Appendices:] The online appendix file contains the detailed derivation, algorithm, and data description (Appendices.pdf).

\item[R Package:] The R-package \textit{mfair} contains the codes used in fitting the MFAI model and analyzing the results (mfair\_1.0.0.tar.gz).

\item[Implementations of Compared Methods:] \

\textit{flashr} \url{https://github.com/stephenslab/flashr}

\textit{cmfrec} \url{https://cran.r-project.org/package=cmfrec}

\textit{softImpute} \url{https://cran.r-project.org/package=softImpute}

\end{description}

\newpage





\appendix

\spacingset{1} 
\begin{center}
    \LARGE
    \bf
    Appendices to ``MFAI: A Scalable Bayesian Matrix Factorization Approach to Leveraging Auxiliary Information''
\end{center}






\spacingset{1.5} 
\section{Fitting the Single-Factor MFAI Model}
\label{append_sec:fit}

Given the main data matrix $\bfY \in \bbR^{N \times M}$ of $N$ samples and $M$ features, and the auxiliary matrix $\bfX \in \bbR^{N \times C}$ of $N$ samples and $C$ covariates, we consider the following single-factor matrix factorization problem
\begin{equation} \label{append_rank1}
    \begin{aligned}
        & \bfY = \bfz \bfw^{\rmT} + \bfepsilon, \\
        & \bfz \sim \calN_N(F \left( \bfX \right), \beta^{-1}\bfI_N), \\
        & \bfw \sim \calN_M(0, \bfI_M), \\
        & \bfepsilon_{nm} \sim \calN(0, \tau^{-1}), \  n = 1, \dots, N \  \text{and} \  m = 1, \dots, M,
    \end{aligned}
\end{equation}
where $\bfz \in \bbR^{N \times 1}$ is the latent factor, $\bfw \in \bbR^{M \times 1}$ is the latent loading, $\bfepsilon \in \bbR^{N \times M}$ is a matrix of residual error terms, $F: \bbR^{C} \rightarrow \bbR$ is the unknown function, $F \left( \bfX \right) \in \bbR^{N \times 1}$ denotes $\left( F \left( \bfX_{1 \cdot} \right), \dots, F \left( \bfX_{N \cdot} \right) \right)^{\rmT}$, $\bfX_{n \cdot} = \left( \bfX_{n1}, \dots, \bfX_{nC} \right)^{\rmT} \in \bbR^{C \times 1}$ is the $n$-th row of $\bfX$ containing auxiliary information for the $n$-th row entity for $n = 1, \dots, N$, and $\beta$ and $\tau$ are two precision parameters.
Then, the joint probabilistic model becomes
\begin{equation}\label{append_joint_rank1}
    \rmPr \left( \bfY, \bfz, \bfw \mid \bfTheta, F \left( \cdot \right) \right) = \rmPr \left( \bfY \mid \bfz, \bfw; \tau \right) \rmPr \left( \bfz \mid \beta, F \left( \cdot \right) \right) \rmPr \left( \bfw \right),
\end{equation}
where $\bfTheta = \{ \tau, \beta \}$ is the collection of model parameters.
The goal is to estimate $\bfTheta$ and $F \left( \cdot \right)$ by optimizing the log marginal likelihood
\begin{equation}\label{append_max_marginal_rank1}
    \begin{aligned}
        \left(\widehat{\bfTheta}, \widehat{F} \left( \cdot \right) \right) & = \argmax{\bfTheta, F \left( \cdot \right)} \log \rmPr \left( \bfY \mid \bfTheta, F \left( \cdot \right) \right) \\
        & = \argmax{\bfTheta, F \left( \cdot \right)} \log \int \rmPr \left( \bfY, \bfz, \bfw \mid \bfTheta, F \left( \cdot \right) \right) \, d\bfz \, d\bfw.
    \end{aligned} 
\end{equation}
The posterior probability of $\bfz$ and $\bfw$ is given as
\begin{equation}\label{append_post_rank1}
    \rmPr \left( \bfz, \bfw \mid \bfY; \widehat{\bfTheta}, \widehat{F} \left( \cdot \right) \right) = \frac{\rmPr \left( \bfY, \bfz, \bfw \mid \widehat{\bfTheta}, \widehat{F} \left( \cdot \right) \right)}{\rmPr \left( \bfY \mid \widehat{\bfTheta}, \widehat{F} \left( \cdot \right) \right)}.
\end{equation}

\subsection{Approximate Bayesian Inference}
\label{append_ssec:approx}

The Bayesian inference using \eqref{append_max_marginal_rank1} and \eqref{append_post_rank1} is intractable since the marginal likelihood $\rmPr \left( \bfY \mid \bfTheta, F \left( \cdot \right) \right)$ cannot be computed by marginalizing all latent variables.
To tackle the Bayesian inference problem, here we propose a variational expectation-maximization (EM) algorithm \citep{bishop2006pattern, blei2017variational} to perform approximate Bayesian inference.
To apply variational approximation, we first define $q \left( \bfz, \bfw \right)$ as an approximated distribution of posterior $\rmPr \left( \bfz, \bfw \mid \bfY; \bfTheta, F \left( \cdot \right) \right)$.
Then the logarithm of the marginal likelihood is
\begin{equation} \label{append_jensen}
    \begin{aligned}
        \log \rmPr \left( \bfY \mid \bfTheta, F \left( \cdot \right) \right)
        & = \log \int \rmPr \left( \bfY, \bfz, \bfw \mid \bfTheta, F \left( \cdot \right) \right) \, d\bfz \, d\bfw \\
        & = \log \int q \left( \bfz, \bfw \right) \frac{\rmPr \left( \bfY, \bfz, \bfw \mid \bfTheta, F \left( \cdot \right) \right)}{q \left( \bfz, \bfw \right)} \, d\bfz \, d\bfw \\
        & \geq \int q \left( \bfz, \bfw \right) \log \frac{\rmPr \left( \bfY, \bfz, \bfw \mid \bfTheta, F \left( \cdot \right) \right)}{q \left( \bfz, \bfw \right)} \, d\bfz \, d\bfw \\
        & = \bbE_{q} \left[ \log \rmPr \left( \bfY, \bfz, \bfw \mid \bfTheta, F \left( \cdot \right) \right) \right] - \bbE_{q} \left[ \log q \left( \bfz, \bfw \right) \right] \\
        & \triangleq \ELBO \left( q; \bfTheta, F \left( \cdot \right) \right),
    \end{aligned}
\end{equation}
where we have adopted Jensen’s inequality to obtain the evidence lower bound (ELBO).
The equality holds if and only if $q \left( \bfz, \bfw \right)$ is the exact posterior $\rmPr \left( \bfz, \bfw \mid \bfY; \bfTheta, F \left( \cdot \right) \right)$.
Instead of maximizing the logarithm of the marginal likelihood, we can iteratively maximize the ELBO with respect to the variational approximate posterior $q$, the model parameters $\bfTheta$, and the function $F \left( \cdot \right)$
\begin{equation}\label{append_max_elbo}
    \begin{aligned}
        \left( \widehat{q}; \widehat{\bfTheta}, \widehat{F}\left( \cdot \right) \right) & = \argmax{q; \bfTheta, F \left( \cdot \right)} \ELBO \left( q; \bfTheta, F \left( \cdot \right) \right).
    \end{aligned} 
\end{equation}
Using the terminology in the EM algorithm, maximization of ELBO with respect to $q$ is known as the E-step, and maximization of ELBO with respect to $\bfTheta$ and $F \left( \cdot \right)$ is known as the M-step.

\subsubsection{E-step}
\label{append_sssec:estep}

In the E-step, we aim to find the optimal solution for the approximate posterior with the current estimate for model parameters $\bfTheta$ and function $F \left( \cdot \right)$, i.e., we treat $\bfTheta$ and $F \left( \cdot \right)$ are fixed.
Then the complete-data log-likelihood is given as
\begin{equation} \label{append_ll}
    \begin{aligned}
        \log\rmPr \left( \bfY, \bfz, \bfw \mid \tau, \beta; F \left( \cdot \right) \right) = & \log \rmPr \left( \bfY \mid \bfz, \bfw; \tau \right) + \log \rmPr \left( \bfz \mid \beta; F \left( \cdot \right) \right) + \log \rmPr \left( \bfw \right) \\
        = & - \frac{NM}{2} \log \left( 2 \pi \right) + \frac{NM}{2} \log \tau - \frac{\tau}{2} \left\| \bfY - \bfz \bfw^{\rmT} \right\|_{\text{F}}^2 \\
        & - \frac{N}{2} \log \left( 2 \pi \right) + \frac{N}{2} \log \beta - \frac{\beta}{2} \left\| \bfz - F \left( \bfX \right) \right\|_2^2 \\
        & - \frac{M}{2} \log \left( 2 \pi \right) - \frac{1}{2} \left\| \bfw \right\|_2^2,
    \end{aligned} 
\end{equation}
where $\|\cdot\|_{\text{F}}$ for a matrix denotes the Frobenius norm and $\|\cdot\|_{2}$ for a vector denotes the Euclidean norm.
The next step is to maximize ELBO instead of working with the marginal likelihood directly.
As in the main text, we use the following factorized distribution to approximate the true posterior
\begin{equation}\label{append_meanfield}
    \begin{aligned}
        q \left( \bfz, \bfw \right) = q \left( \bfz \right) q \left( \bfw \right).
    \end{aligned}
\end{equation}
Amongst all distributions $q \left( \bfz, \bfw \right)$ having the form \eqref{append_meanfield}, we now seek that distribution for which the $\ELBO \left( q \right)$ is largest.
To achieve this, we make a variational optimization of $\ELBO \left( q \right)$ with respect to each of the factors $q \left( \bfz \right)$ and $q \left( \bfw \right)$ in turn.
For simplicity, we ignore the notation of the model parameters $\bfTheta$ and function $F \left( \cdot \right)$ in this part.
We first keep the $q \left( \bfw \right)$ fixed, then the ELBO can be written in the following form:
\begin{equation} \label{append_elbo_qz}
    \begin{aligned}
        \ELBO \left( q \right) & = \bbE_{q \left( \bfz \right)} \left[ \bbE_{q \left( \bfw \right)} \left[ \log \rmPr \left( \bfY, \bfz, \bfw \right) \right] - \log q \left( \bfz \right) \right] + \const \\
        & = \int q \left( \bfz \right) \frac{\log \widetilde{\rmPr} \left( \bfY, \bfz \right)}{\log q \left( \bfz \right)} \, d\bfz + \const,
    \end{aligned}
\end{equation}
where we have defined a new distribution $\log \widetilde{\rmPr} \left( \bfY, \bfz \right)$ by the relation
\begin{equation} \label{append_new_dist_z}
    \begin{aligned}
        \log \widetilde{\rmPr} \left( \bfY, \bfz \right) = \bbE_{q \left( \bfw \right)} \left[ \log \rmPr \left( \bfY, \bfz, \bfw \right) \right] + \const.
    \end{aligned}
\end{equation}
By recognizing that \eqref{append_elbo_qz} is a negative Kullback–Leibler (KL) divergence between $q \left( \bfz \right)$ and $\widetilde{\rmPr} \left( \bfY, \bfz \right)$, it's clear that the ELBO achieves maximum value when $q \left( \bfz \right) = \widetilde{\rmPr} \left( \bfY, \bfz \right)$.
Hence we can obtain the optimal solution for $q \left( \bfz \right)$ by
\begin{equation} \label{append_qz}
    \begin{aligned}
        & \log q \left( \bfz \right) \\
        = & \bbE_{q \left( \bfw \right)} \left[ \log\rmPr \left( \bfY, \bfz, \bfw \right) \right] + \const \\
        = & \bbE_{q \left( \bfw \right)} \left[ - \frac{\tau}{2} \left\| \bfY - \bfz \bfw^{\rmT} \right\|_{\text{F}}^2 \right] - \frac{\beta}{2} \left\| \bfz - F \left( \bfX \right) \right\|_2^2 + \const \\
        = & - \frac{\tau}{2} \sum_{m=1}^M \bbE_{q \left( \bfw \right)} \left[ \left(\bfY_{.m} - \bfw_m \bfz \right)^{\rmT} \left( \bfY_{.m} - \bfw_m \bfz \right) \right] - \frac{\beta}{2} \left( \bfz - F \left( \bfX \right) \right)^{\rmT} \left( \bfz - F \left( \bfX \right) \right) + \const \\
        = & - \frac{1}{2} \bfz^{\rmT} \diag \left( \beta + \tau \bbE_{q \left( \bfw \right)} \left[ \left\| \bfw \right\|_2^2 \right] \right)_N \bfz + \bfz^{\rmT} \left( \beta F \left( \bfX \right) + \tau \bfY \bbE_{q \left( \bfw \right)} \left[ \bfw \right] \right) + \const,
    \end{aligned}
\end{equation}
where $\diag \left( \beta + \tau \bbE_{q \left( \bfw \right)} \left[ \left\| \bfw \right\|_2^2 \right] \right)_N = \left( \beta + \tau \bbE_{q \left( \bfw \right)} \left[ \left\| \bfw \right\|_2^2 \right] \right) \bfI_N \in \bbR^{N \times N}$ is a diagonal matrix.
Then we fix $q \left( \bfz \right)$ and optimize $\ELBO \left( q \right)$ with respect to $q \left( \bfw \right)$.
Similarly, the ELBO can be written as
\begin{equation} \label{append_elbo_qw}
    \begin{aligned}
        \ELBO \left( q \right) & = \bbE_{q \left( \bfw \right)} \left[ \bbE_{q \left( \bfz \right)} \left[ \log \rmPr \left( \bfY, \bfz, \bfw \right) \right] - \log q \left( \bfw \right) \right] + \const \\
        & = \int q \left( \bfw \right) \frac{\log \widetilde{\rmPr} \left( \bfY, \bfw \right)}{\log q \left( \bfw \right)} \, d\bfw + \const,
    \end{aligned}
\end{equation}
where
\begin{equation} \label{append_new_dist_w}
    \begin{aligned}
        \log \widetilde{\rmPr} \left( \bfY, \bfw \right) = \bbE_{q \left( \bfz \right)} \left[ \log \rmPr \left( \bfY, \bfz, \bfw \right) \right] + \const.
    \end{aligned}
\end{equation}
And the optimal solution for $q \left( \bfw \right)$ is given by
\begin{equation} \label{append_qw}
    \begin{aligned}
        & \log q \left( \bfw \right) \\
        = & \bbE_{q \left( \bfz \right)} \left[ \log\rmPr \left( \bfY, \bfz, \bfw \right) \right] + \const \\
        = & \bbE_{q \left( \bfz \right)} \left[ - \frac{\tau}{2} \left\| \bfY - \bfz \bfw^{\rmT} \right\|_{\text{F}}^2 \right] - \frac{1}{2} \left\| \bfw \right\|_2^2 + \const \\
        = & - \frac{\tau}{2} \sum_{n=1}^N \bbE_{q \left( \bfz \right)} \left[ \left(\bfY_{n \cdot} - \bfz_n \bfw \right)^{\rmT} \left( \bfY_{n \cdot} - \bfz_n \bfw \right) \right] -\frac{1}{2} \bfw^{\rmT} \bfw + \const \\
        = & - \frac{1}{2} \bfw^{\rmT} \diag \left( 1 + \tau \bbE_{q \left( \bfz \right)} \left[ \left\| \bfz \right\|_2^2 \right] \right)_M \bfw + \bfw^{\rmT} \left( \tau \bfY^{\rmT} \bbE_{q \left( \bfz \right)} \left[ \bfz \right] \right) + \const,
    \end{aligned}
\end{equation}
where $\diag \left( 1 + \tau \bbE_{q \left( \bfz \right)} \left[ \left\| \bfz \right\|_2^2 \right] \right)_M = \left( 1 + \tau \bbE_{q \left( \bfz \right)} \left[ \left\| \bfz \right\|_2^2 \right] \right) \bfI_M \in \bbR^{M \times M}$ is a diagonal matrix.
The quadratic forms \eqref{append_qz} and \eqref{append_qw} indicate that both $\bfz$ and $\bfw$ follow Gaussian distribution
\begin{equation} \label{append_q_gaussian}
    q \left( \bfz \right) = \calN_N \left( \bfz \mid \bfmu, \bfA \right), \  q \left( \bfw \right) = \calN_M \left( \bfw \mid \bfnu, \bfB \right),
\end{equation}
where $\bfmu \in \bbR^{N \times 1}$ and $\bfnu \in \bbR^{M \times 1}$ are posterior mean vectors and $\bfA \in \bbR^{N \times N}$ and $\bfB \in \bbR^{M \times M}$ are covariance matrices
\begin{equation} \label{append_q_para}
    \begin{aligned}
        & \bfA = \diag \left( a^2 \right)_N, \  \frac{1}{a^2} \bfmu = \beta F \left( \bfX \right) + \tau \bfY \bfnu, \\
        & \bfB = \diag \left( b^2 \right)_M, \  \frac{1}{b^2} \bfnu = \tau \bfY^{\rmT} \bfmu,
    \end{aligned} 
\end{equation}
and
\begin{equation} \label{append_q_var}
    a^2 = \frac{1}{\beta + \tau \left( \left\| \bfnu \right\|_2^2 + M b^2 \right)}, \  b^2 = \frac{1}{1 + \tau \left( \left\| \bfmu \right\|_2^2 + N a^2\right)}.
\end{equation}

\subsubsection{M-step}
\label{append_sssec:mstep}

In the M-step, we turn to fix the variational approximate posterior $q \left( \bfz, \bfw \right)$ and maximize the ELBO with respect to $\bfTheta$ and $F \left( \cdot \right)$.
With \eqref{append_q_gaussian}, the ELBO is given by
\begin{equation}\label{append_elbo}
    \begin{aligned}
        & \ELBO \left( q; \tau, \beta; F \left( \cdot \right) \right) \\
        = & \bbE_{q} \left[ \log \rmPr \left( \bfY, \bfz, \bfw \right) \right] - \bbE_{q} \left[ \log q \left( \bfz, \bfw \right) \right] \\
        = & - \frac{NM}{2} \log\left( 2\pi \right) + \frac{NM}{2} \log \tau \\
        & - \frac{\tau}{2} \left( \left\| \bfY - \bfmu \bfnu^{\rmT} \right\|_{\text{F}}^2 + \left\| \left( \bfmu^2 + \diag \left( \bfA \right) \right) \left( \bfnu^2 + \diag \left( \bfB \right) \right)^{\rmT} - \bfmu^2 \left( \bfnu^2 \right)^{\rmT} \right\|_{1,1} \right) \\
        & - \frac{N}{2} \log \left( 2 \pi \right) + \frac{N}{2} \log \beta - \frac{\beta}{2} \left( \left\| \bfmu - F \left( \bfX \right) \right\|_2^2 + \trace \left( \bfA \right) \right) \\
        & - \frac{M}{2} \log \left( 2 \pi \right) - \frac{1}{2} \left\| \bfnu \right\|_2^2 - \frac{1}{2} \trace \left( \bfB \right) \\
        & + \frac{N}{2} \log \left( 2 \pi \right) + \frac{1}{2} \log \det \left( \bfA \right) + \frac{N}{2} \\
        & + \frac{M}{2} \log \left( 2 \pi \right) + \frac{1}{2} \log \det \left( \bfB \right) + \frac{M}{2} \\
        = & \frac{NM}{2} \log \tau - \frac{\tau}{2} \left( \left\| \bfY - \bfmu \bfnu^{\rmT} \right\|_{\text{F}}^2 + \left\| \left( \bfmu^2 + \diag \left( \bfA \right) \right) \left( \bfnu^2 + \diag \left( \bfB \right) \right)^{\rmT} - \bfmu^2 \left( \bfnu^2 \right)^{\rmT} \right\|_{1,1} \right) \\
        & + \frac{N}{2} \log \beta - \frac{\beta}{2} \left( \left\| \bfmu - F \left( \bfX \right) \right\|_2^2 + N a^2 \right) + \const,
    \end{aligned} 
\end{equation}
where $\diag \left( \cdot \right)$ denotes the vector containing all the entries on the main diagonal of a squared matrix and $\trace \left( \cdot \right)$ denotes the trace.
We consider the model parameters first and fix the current estimate for $F \left( \cdot \right)$.
Setting the derivative of \eqref{append_elbo} with respect to $\tau$ and $\beta$ be 0, we can obtain the estimation of these two parameters
\begin{equation}\label{append_para}
    \begin{aligned}
        & \tau = \frac{NM}{\left\| \bfY - \bfmu \bfnu^{\rmT} \right\|_{\text{F}}^2 + \left\| \left( \bfmu^2 + \diag \left( \bfA \right) \right) \left( \bfnu^2 + \diag \left( \bfB \right) \right)^{\rmT} - \bfmu^2 \left( \bfnu^2 \right)^{\rmT} \right\|_{1,1}}, \\
        & \beta = \frac{N}{\left\| \bfmu - F \left( \bfX \right) \right\|_2^2 + N a^2}. \\
    \end{aligned} 
\end{equation}
Then we fix $\bfTheta$, and the ELBO can be written as
\begin{equation} \label{append_elbo_F}
    \begin{aligned}
        \ELBO = - \frac{\beta}{2} \left\| \bfmu - F \left( \bfX \right) \right\|_2^2 + \const.
    \end{aligned} 
\end{equation}
Maximizing \eqref{append_elbo_F} with respect to $F \left( \cdot \right)$ is equivalent to solving the following optimization problem:
\begin{equation}\label{append_elbo_F_equiv}
    \begin{aligned}
        \min_{F \left( \cdot \right)} \calL \left( \bfmu, F \left( \cdot \right) \right),
    \end{aligned} 
\end{equation}
where
\begin{equation} \label{append_loss_F}
    \begin{aligned}
        \calL \left( \bfmu, F \left( \cdot \right) \right) = \frac{\beta}{2} \left\| \bfmu - F \left( \bfX \right) \right\|_2^2.
    \end{aligned}
\end{equation}
Here, we adopt the idea of the gradient boosting machine in each iteration to construct $F \left( \cdot \right)$.
Suppose in the $(t-1)$-th step, the current estimate for $F \left( \cdot \right)$ is denoted as $F^{(t-1)} \left( \cdot \right)$.
Specifically, boosting finds a minimizer of the loss function $\calL \left( \bfmu, F \left( \cdot \right) \right)$ in a stage-wise manner by sequentially adding an update $f^{(t)} \left( \cdot \right)$ to $F^{(t-1)} \left( \cdot \right)$ in the $t$-th step
\begin{equation} \label{append_boosting}
    F^{(t)} \left( \cdot \right) = F^{(t-1)} \left( \cdot \right) + f^{(t)} \left( \cdot \right),
\end{equation}
where $T$ is the total number of trees, and
\begin{equation} \label{append_ft}
    f^{(t)} \left( \cdot \right) = \argmin{f \left( \cdot \right)} \calL \left( \bfmu, F^{(t-1)} \left( \cdot \right) + f \left( \cdot \right) \right),
\end{equation}
is a single regression tree.
With the adoption of the Newton boosting \citep{sigrist2021gradient}, we need to compute the first and second gradient with respect to $F \left( \cdot \right)$ at the current estimate $F^{(t-1)} \left( \cdot \right)$
\begin{equation} \label{append_m_step_F_deri}
    \begin{aligned}
        & \left. \pdv{ \calL \left( \bfmu, F \left( \cdot \right) \right) }{ F \left( \cdot \right) } \right|_{F \left( \cdot \right) = F^{(t-1)} \left( \cdot \right)} = - \beta \left( \bfmu - F^{(t-1)} \left( \bfX \right) \right) \triangleq \bfG\left( \bfmu, F^{(t-1)} \left( \cdot \right) \right), \\
        & \left. \pdv[2]{ \calL \left( \bfmu, F \left( \cdot \right) \right) }{ F \left( \cdot \right) } \right|_{F \left( \cdot \right) = F^{(t-1)} \left( \cdot \right)} = \beta \bfI_{N} \triangleq \bfH \left( \bfmu, F^{(t-1)} \left( \cdot \right) \right).
    \end{aligned}
\end{equation}
Then, we can obtain the working response
\begin{equation} \label{append_boosting_response}
    - \bfH^{-1} \left( \bfmu, F^{(t-1)} \left( \cdot \right) \right) \bfG\left( \bfmu, F^{(t-1)} \left( \cdot \right) \right) = \bfmu - F^{(t-1)} \left( \bfX \right),
\end{equation}
thus the optimal solution for $f^{(t)} \left( \cdot \right)$ is given by
\begin{equation} \label{append_newton_boosting}
    f^{(t)} \left( \cdot \right) = \argmin{f \left( \cdot \right)} \left\| \bfmu - F^{(t-1)} \left( \bfX \right) - f \left( \bfX \right) \right\|_2^2 .
\end{equation}
Further, it has been empirically observed that higher prediction accuracy can be obtained by damping the update \citep{friedman2001greedy}
\begin{equation} \label{append_shrinkage}
    F^{(t)} \left( \cdot \right) = F^{(t-1)} \left( \cdot \right) + s \cdot f^{(t)} \left( \cdot \right),
\end{equation}
where $0 < s < 1$ is the so-called shrinkage parameter or learning rate.

\subsection{Missing Data in the Main Matrix}
\label{append_ssec:fit_missing}
To handle $\bfY$ with missing entries, we first make the typical assumption that they are missing at random (MAR) \citep{rubin1976inference, little1987statistical}, that is, given the observed data, the missingness does not depend on the unobserved data or latent variables.
Then, we can consider the following probabilistic model only for the observed entries $\bfY^{\text{obs}}$
\begin{equation} \label{append_lh_missing}
    \begin{aligned}
        \rmPr \left( \bfY^{\text{obs}} \mid \bfz, \bfw; \tau \right) = \prod_{\left(n, m \right) \in  \Omega^\text{obs}} \rmPr \left( \bfY_{nm} \mid \bfz, \bfw; \tau \right),
    \end{aligned}
\end{equation}
where $\Omega^\text{obs}$ is the collection of the indices of the observed entries of $\bfY$.
Then, the log-likelihood function can be written as
\begin{equation} \label{append_ll_missing}
    \begin{aligned}
        \log\rmPr \left( \bfY^{\text{obs}}, \bfz, \bfw \mid \tau, \beta; F \left( \cdot \right) \right) = & \log \rmPr \left( \bfY^{\text{obs}} \mid \bfz, \bfw; \tau \right) + \log \rmPr \left( \bfz \mid \beta; F \left( \cdot \right) \right) + \log \rmPr \left( \bfw \right) \\
        = & - \frac{\left| \Omega^{\text{obs}} \right|}{2} \log \left( 2 \pi \right) + \frac{\left| \Omega^{\text{obs}} \right|}{2} \log \tau - \frac{\tau}{2} \left\| \calP_{\Omega^{\text{obs}}} \left( \bfY - \bfz \bfw^{\rmT} \right) \right\|_{\text{F}}^2 \\
        & - \frac{N}{2} \log \left( 2 \pi \right) + \frac{N}{2} \log \beta - \frac{\beta}{2} \left\| \bfz - F \left( \bfX \right) \right\|_2^2 \\
        & - \frac{M}{2} \log \left( 2 \pi \right) - \frac{1}{2} \left\| \bfw \right\|_2^2,
    \end{aligned}
\end{equation}
where $\calP$ is a projection operator and $\calP_{\Omega} \left( \bfY \right)$ outputs a matrix with the same dimension as that of $\bfY$
\begin{equation} \label{append_proj}
    \left( \calP_{\Omega} \left( \bfY \right) \right)_{nm} =
        \begin{cases}
        \bfY_{nm}, & \text{if $\left(n, m \right) \in \Omega$}, \\
        0, & \text{otherwise}.
    \end{cases}
\end{equation}
To simplify the derivation and exploit standard fast matrix multiplication routines, we introduce the data precision matrix $\bftau \in \bbR^{N \times M}$ as follows
\begin{equation} \label{append_tau_missing}
    \begin{aligned}
        \bftau_{nm} =
        \begin{cases}
        \tau, & \text{if $\left(n, m \right) \in \Omega^\text{obs}$}, \\
        0, & \text{otherwise}.
    \end{cases}
    \end{aligned}
\end{equation}
Then similar to Section \ref{append_ssec:approx}, the optimal solutions of the variational approximate posteriors in the E-step are given as
\begin{equation} \label{append_q_missing}
    \begin{aligned}
        & \log q \left( \bfz \right) \\
        = & \bbE_{q \left( \bfw \right)} \left[ \log\rmPr \left( \bfY^{\text{obs}}, \bfz, \bfw \right) \right] + \const \\
        = & \bbE_{q \left( \bfw \right)} \left[ - \frac{\tau}{2} \left\| \calP_{\Omega^{\text{obs}}} \left( \bfY - \bfz \bfw^{\rmT} \right) \right\|_{\text{F}}^2 \right] - \frac{\beta}{2} \left\| \bfz - F \left( \bfX \right) \right\|_2^2 + \const \\
        = & - \frac{1}{2} \sum_{m=1}^M \bbE_{q \left( \bfw \right)} \left[ \left(\bfY_{.m} - \bfw_m \bfz \right)^{\rmT} \left( \bftau_{\cdot m} \circ \left( \bfY_{.m} - \bfw_m \bfz \right) \right) \right] - \frac{\beta}{2} \left( \bfz - F \left( \bfX \right) \right)^{\rmT} \left( \bfz - F \left( \bfX \right) \right) + \const \\
        = & - \frac{1}{2} \bfz^{\rmT} \left( \sum_{m=1}^{M} \bbE_{q \left( \bfw \right)} \left[ \bfw_{m}^{2} \right] \diag \left( \bftau_{\cdot m} \right) \right) \bfz + \bfz^{\rmT} \left( \sum_{m=1}^{M} \bbE_{q \left( \bfw \right)} \left[ \bfw_{m} \right] \left( \bftau_{\cdot m} \circ \bfY_{\cdot m} \right) \right) \\
        & - \frac{\beta}{2} \bfz^{\rmT} \bfz + \bfz^{\rmT} \left( \beta F \left( \bfX \right) \right) + \const \\
        = & - \frac{1}{2} \bfz^{\rmT} \diag \left( \beta + \bftau \bbE_{q \left( \bfw \right)} \left[ \bfw^2 \right] \right) \bfz + \bfz^{\rmT} \left( \beta F \left( \bfX \right) + \left( \bftau \circ \bfY \right) \bbE_{q \left( \bfw \right)} \left[ \bfw \right] \right) + \const, \\
        & \log q \left( \bfw \right) \\
        = & \bbE_{q \left( \bfz \right)} \left[ \log\rmPr \left( \bfY^{\text{obs}}, \bfz, \bfw \right) \right] + \const \\
        = & \bbE_{q \left( \bfz \right)} \left[ - \frac{\tau}{2} \left\| \calP_{\Omega^{\text{obs}}} \left( \bfY - \bfz \bfw^{\rmT} \right) \right\|_{\text{F}}^2 \right] - \frac{1}{2} \left\| \bfw \right\|_2^2 + \const \\
        = & - \frac{1}{2} \sum_{n=1}^{N} \bbE_{q \left( \bfz \right)} \left[ \left( \bfY_{n \cdot} - \bfz_n \bfw \right)^{\rmT} \left( \bftau_{n \cdot} \circ \left( \bfY_{n \cdot} - \bfz_n \bfw \right) \right) \right] -\frac{1}{2} \bfw^{\rmT} \bfw + \const \\
        = & - \frac{1}{2} \bfw^{\rmT} \left( \sum_{n=1}^{N} \bbE_{q \left( \bfz \right)} \left[ \bfz_n^2 \right] \diag \left( \bftau_{n \cdot} \right) \right) \bfw + \bfw^{\rmT} \left( \sum_{n=1}^{N} \bbE_{q \left( \bfz \right)} \left[ \bfz_{n} \right] \left( \bftau_{n \cdot} \circ \bfY_{n \cdot} \right) \right) -\frac{1}{2} \bfw^{\rmT} \bfw + \const \\
        = & - \frac{1}{2} \bfw^{\rmT} \diag \left( 1 + \bftau^{\rmT} \bbE_{q \left( \bfz \right)} \left[ \bfz^2 \right] \right) \bfw + \bfw^{\rmT} \left( \left( \bftau \circ \bfY \right)^{\rmT} \bbE_{q \left( \bfz \right)} \left[ \bfz \right] \right) + \const,
    \end{aligned}
\end{equation}
where $\circ$ denotes the Hadamard product (i.e., element-wise product), $\bfz^2 = \bfz \circ \bfz \in \bbR^{N \times 1}$ and $\bfw^2 = \bfw \circ \bfw \in \bbR^{M \times 1}$, and $\diag \left( \beta + \bftau \bbE_{q \left( \bfw \right)} \left[ \bfw^2 \right] \right) \in \bbR^{N \times N}$ and $\diag \left( 1 + \bftau^{\rmT} \bbE_{q \left( \bfz \right)} \left[ \bfz^2 \right] \right) \in \bbR^{M \times M}$ are two diagonal matrices in which the main diagonal entries are equal to $\beta + \bftau \bbE_{q \left( \bfw \right)} \left[ \bfw^2 \right]$ and $1 + \bftau^{\rmT} \bbE_{q \left( \bfz \right)} \left[ \bfz^2 \right]$ respectively
\begin{equation} \label{append_diag}
    \begin{aligned}
        \diag \left( \beta + \bftau \bbE_{q \left( \bfw \right)} \left[ \bfw^2 \right] \right) 
        & =
        \begin{bmatrix}
            \beta + \bftau_{1 \cdot}^{\rmT} \bbE_{q \left( \bfw \right)} \left[ \bfw^2 \right]  & & \\
            & \ddots & \\
            & & \beta + \bftau_{N \cdot}^{\rmT} \bbE_{q \left( \bfw \right)} \left[ \bfw^2 \right]
        \end{bmatrix} \\
        & = 
        \begin{bmatrix}
            \beta + \sum_{m=1}^{M} \bftau_{1m} \bbE_{q \left( \bfw \right)} \left[ \bfw_m^2 \right]  & & \\
            & \ddots & \\
            & & \beta + \sum_{m=1}^{M} \bftau_{Nm} \bbE_{q \left( \bfw \right)} \left[ \bfw_m^2 \right]
        \end{bmatrix}, \\
        \diag \left( 1 + \bftau^{\rmT} \bbE_{q \left( \bfz \right)} \left[ \bfz^2 \right] \right) & =
        \begin{bmatrix}
            1 + \bftau_{\cdot 1}^{\rmT} \bbE_{q \left( \bfz \right)} \left[ \bfz^2 \right]  & & \\
            & \ddots & \\
            & & 1 + \bftau_{\cdot M}^{\rmT} \bbE_{q \left( \bfz \right)} \left[ \bfz^2 \right]
        \end{bmatrix} \\
        & =
        \begin{bmatrix}
            1 + \sum_{n=1}^{N} \bftau_{n1} \bbE_{q \left( \bfz \right)} \left[ \bfz_n^2 \right]  & & \\
            & \ddots & \\
            & & 1 + \sum_{n=1}^{N} \bftau_{nM} \bbE_{q \left( \bfz \right)} \left[ \bfz_n^2 \right]
        \end{bmatrix}.
    \end{aligned}
\end{equation}
The quadratic forms \eqref{append_q_missing} indicate that both $\bfz$ and $\bfw$ follow Gaussian distribution
\begin{equation} \label{append_q_gaussian_missing}
    q \left( \bfz \right) = \calN_N \left( \bfz \mid \bfmu, \bfA \right), \  q \left( \bfw \right) = \calN_M \left( \bfw \mid \bfnu, \bfB \right),
\end{equation}
where $\bfmu \in \bbR^{N \times 1}$ and $\bfnu \in \bbR^{M \times 1}$ are posterior mean vectors and $\bfA \in \bbR^{N \times N}$ and $\bfB \in \bbR^{M \times M}$ are covariance matrices
\begin{equation} \label{append_q_para_missing}
    \begin{aligned}
        & \bfA = \diag \left( \bfa^2 \right), \  \bfmu = \bfA \left( \beta F \left( \bfX \right) + \left( \bftau \circ \bfY \right) \bfnu 
        \right), \\
        & \bfB = \diag \left( \bfb^2 \right), \  \bfnu = \bfB \left( \bftau \circ \bfY \right)^{\rmT} \bfmu,
    \end{aligned} 
\end{equation}
where $\bfa^2 \in \bbR^{N \times 1}$ and $\bfb^2 \in \bbR^{M \times 1}$ are two vectors
\begin{equation} \label{append_q_var_missing}
    \begin{aligned}
        & {\bfa^2}_{n} = \frac{1}{\beta + \bftau_{n \cdot}^{\rmT} \left( \bfnu^2 + \bfb^2 \right)}, \  \bfmu_{n} = {\bfa^2}_{n} \left( \beta F \left( \bfX_{n \cdot} \right) + \left( \bftau_{n \cdot} \circ \bfY_{n \cdot} \right)^{\rmT} \bfnu \right), \  n = 1, \dots, N \\
        & {\bfb^2}_{m} = \frac{1}{1 + \bftau_{\cdot m}^{\rmT} \left( \bfmu^2 + \bfa^2 \right)}, \  \bfnu_{m} = {\bfb^2}_{m} \left( \bftau_{\cdot m} \circ \bfY_{\cdot m} \right)^{\rmT} \bfmu, \  m = 1, \dots, M.
    \end{aligned}
\end{equation}
In the M-step, we turn to fix the variational approximate posterior $q \left( \bfz, \bfw \right)$ and maximize the ELBO with respect to $\bfTheta$ and $F \left( \cdot \right)$.
With \eqref{append_q_gaussian_missing}, the ELBO is given by
\begin{equation}\label{append_elbo_missing}
    \begin{aligned}
        & \ELBO \left( q; \tau, \beta; F \left( \cdot \right) \right) \\
        = & \bbE_{q} \left[ \log \rmPr \left( \bfY^{\text{obs}}, \bfz, \bfw \right) \right] - \bbE_{q} \left[ \log q \left( \bfz, \bfw \right) \right] \\
        = & - \frac{\left| \Omega^{\text{obs}} \right|}{2} \log\left( 2\pi \right) + \frac{\left| \Omega^{\text{obs}} \right|}{2} \log \tau \\
        & - \frac{\tau}{2} \left( \left\| \calP_{\Omega^{\text{obs}}} \left( \bfY - \bfmu \bfnu^{\rmT} \right) \right\|_{\text{F}}^2 + \left\| \calP_{\Omega^{\text{obs}}} \left( \left( \bfmu^2 + \diag \left( \bfA \right) \right) \left( \bfnu^2 + \diag \left( \bfB \right) \right)^{\rmT} - \bfmu^2 \left( \bfnu^2 \right)^{\rmT} \right) \right\|_{1,1} \right) \\
        & - \frac{N}{2} \log \left( 2 \pi \right) + \frac{N}{2} \log \beta - \frac{\beta}{2} \left( \left\| \bfmu - F \left( \bfX \right) \right\|_2^2 + \trace \left( \bfA \right) \right) \\
        & - \frac{M}{2} \log \left( 2 \pi \right) - \frac{1}{2} \left\| \bfnu \right\|_2^2 - \frac{1}{2} \trace \left( \bfB \right) \\
        & + \frac{N}{2} \log \left( 2 \pi \right) + \frac{1}{2} \log \det \left( \bfA \right) + \frac{N}{2} \\
        & + \frac{M}{2} \log \left( 2 \pi \right) + \frac{1}{2} \log \det \left( \bfB \right) + \frac{M}{2} \\
        = & \frac{\left| \Omega^{\text{obs}} \right|}{2} \log \tau - \frac{\tau}{2} \left( \left\| \calP_{\Omega^{\text{obs}}} \left( \bfY - \bfmu \bfnu^{\rmT} \right) \right\|_{\text{F}}^2 + \left\| \calP_{\Omega^{\text{obs}}} \left( \left( \bfmu^2 + \bfa^2 \right) \left( \bfnu^2 + \bfb^2 \right)^{\rmT} - \bfmu^2 \left( \bfnu^2 \right)^{\rmT} \right) \right\|_{1,1} \right) \\
        & + \frac{N}{2} \log \beta - \frac{\beta}{2} \left( \left\| \bfmu - F \left( \bfX \right) \right\|_2^2 + \left\| \bfa^2 \right\|_1 \right) + \const,
    \end{aligned} 
\end{equation}
where $\diag \left( \cdot \right)$ denotes the vector containing all the entries on the main diagonal of a squared matrix and $\trace \left( \cdot \right)$ denotes the trace.
We consider the model parameters first and fix the current estimate for $F \left( \cdot \right)$.
Setting the derivative of \eqref{append_elbo_missing} with respect to $\tau$ and $\beta$ be 0, we can obtain the estimation of these two parameters
\begin{equation}\label{append_para_missing}
    \begin{aligned}
        & \tau = \frac{\left| \Omega^{\text{obs}} \right|}{\left\| \calP_{\Omega^{\text{obs}}} \left( \bfY - \bfmu \bfnu^{\rmT} \right) \right\|_{\text{F}}^2 + \left\| \calP_{\Omega^{\text{obs}}} \left( \left( \bfmu^2 + \bfa^2 \right) \left( \bfnu^2 + \bfb^2 \right)^{\rmT} - \bfmu^2 \left( \bfnu^2 \right)^{\rmT} \right) \right\|_{1,1}}, \\
        & \beta = \frac{N}{\left\| \bfmu - F \left( \bfX \right) \right\|_2^2 + \left\| \bfa^2 \right\|_1}. \\
    \end{aligned} 
\end{equation}
The remaining part is the same as in Section \ref{append_ssec:approx}.

\subsection{Different Precision Parameters between Features}
\label{append_ssec:diff}

In this section, we relax the MFAI model and allow different precision parameters between features, which is a common assumption in factor analysis.
The original model becomes as
\begin{equation} \label{append_rank1_diff}
    \begin{aligned}
        & \bfY = \bfz \bfw^{\rmT} + \bfepsilon, \\
        & \bfz \sim \calN_N(F \left( \bfX \right), \beta^{-1}\bfI_N), \\
        & \bfw \sim \calN_M(0, \bfI_M), \\
        & \bfepsilon_{nm} \sim \calN(0, \bftau_{m}^{-1}), \  n = 1, \dots, N \  \text{and} \  m = 1, \dots, M,
    \end{aligned}
\end{equation}
and we denote $\bftau = \left( \bftau_1 , \dots, \bftau_M \right)$.
Then, the complete-data log-likelihood is given as
\begin{equation} \label{append_ll_diff}
    \begin{aligned}
        \log\rmPr \left( \bfY, \bfz, \bfw \mid \bftau, \beta; F \left( \cdot \right) \right) = & \log \rmPr \left( \bfY \mid \bfz, \bfw; \bftau \right) + \log \rmPr \left( \bfz \mid \beta; F \left( \cdot \right) \right) + \log \rmPr \left( \bfw \right) \\
        = & - \frac{NM}{2} \log \left( 2 \pi \right) + \frac{N}{2} \sum_{m = 1}^{M} \log \bftau_{m} - \frac{1}{2} \sum_{m = 1}^{M} \bftau_{m} \left\| \bfY_{\cdot m} - \bfw_{m} \bfz \right\|_{2}^2 \\
        & - \frac{N}{2} \log \left( 2 \pi \right) + \frac{N}{2} \log \beta - \frac{\beta}{2} \left\| \bfz - F \left( \bfX \right) \right\|_2^2 \\
        & - \frac{M}{2} \log \left( 2 \pi \right) - \frac{1}{2} \left\| \bfw \right\|_2^2.
    \end{aligned} 
\end{equation}
In the E-step, the optimal solution of $q \left( \bfz \right)$ will become as
\begin{equation} \label{append_qz_diff}
    \begin{aligned}
        & \log q \left( \bfz \right) \\
        = & \bbE_{q \left( \bfw \right)} \left[ \log\rmPr \left( \bfY, \bfz, \bfw \right) \right] + \const \\
        = & \bbE_{q \left( \bfw \right)} \left[ - \frac{1}{2} \sum_{m = 1}^{M} \bftau_{m} \left\| \bfY_{\cdot m} - \bfw_{m} \bfz \right\|_{2}^2 \right] - \frac{\beta}{2} \left\| \bfz - F \left( \bfX \right) \right\|_2^2 + \const \\
        = & - \frac{1}{2} \sum_{m=1}^M \bftau_m \bbE_{q \left( \bfw \right)} \left[ \left(\bfY_{.m} - \bfw_m \bfz \right)^{\rmT} \left( \bfY_{.m} - \bfw_m \bfz \right) \right] - \frac{\beta}{2} \left( \bfz - F \left( \bfX \right) \right)^{\rmT} \left( \bfz - F \left( \bfX \right) \right) + \const \\
        = & - \frac{1}{2} \bfz^{\rmT} \diag \left( \beta + \sum_{m=1}^{M} \bftau_{m} \bbE_{q \left( \bfw \right)} \left[ \bfw_{m}^2 \right] \right)_N \bfz + \bfz^{\rmT} \left( \beta F \left( \bfX \right) +  \bfY \left( \bftau \odot \bbE_{q \left( \bfw \right)} \left[ \bfw \right] \right) \right) + \const,
    \end{aligned}
\end{equation}
and similarly, the optimal solution of $q \left( \bfw \right)$ will become as
\begin{equation} \label{append_qw_diff}
    \begin{aligned}
        & \log q \left( \bfw \right) \\
        = & \bbE_{q \left( \bfz \right)} \left[ \log\rmPr \left( \bfY, \bfz, \bfw \right) \right] + \const \\
        = & \bbE_{q \left( \bfz \right)} \left[ - \frac{1}{2} \sum_{m = 1}^{M} \tau_{m} \left\| \bfY_{\cdot m} - \bfw_{m} \bfz \right\|_{2}^2 \right] - \frac{1}{2} \left\| \bfw \right\|_2^2 + \const \\
        = & - \frac{1}{2} \sum_{m=1}^M \tau_m \bbE_{q \left( \bfz \right)} \left[ \left(\bfY_{.m} - \bfw_m \bfz \right)^{\rmT} \left( \bfY_{.m} - \bfw_m \bfz \right) \right] -\frac{1}{2} \bfw^{\rmT} \bfw + \const \\
        = & - \frac{1}{2} \bfw^{\rmT} \diag \left( 1 + \bftau_m \bbE_{q \left( \bfz \right)} \left[ \left\| \bfz \right\|_2^2 \right] \right)_M \bfw + \bfw^{\rmT} \left( \bftau \odot \left( \bfY^{\rmT} \bbE_{q \left( \bfz \right)} \left[ \bfz \right] \right) \right) + \const.
    \end{aligned}
\end{equation}
Then, the update equations should be
\begin{equation} \label{append_q_para_diff}
    \begin{aligned}
        & \bfA = \diag \left( a^2 \right)_N, \  \frac{1}{a^2} \bfmu = \beta F \left( \bfX \right) +  \bfY \left( \bftau \odot \bfnu \right), \\
        & \bfB = \diag \left( {\bfb^2}_m \right)_M, \  \bfB^{-1} \bfnu = \bftau \odot \left( \bfY^{\rmT} \bfmu \right),
    \end{aligned} 
\end{equation}
and
\begin{equation} \label{append_q_var_diff}
    a^2 = \frac{1}{\beta + \sum_{m=1}^{M} \bftau_m \left( \bfnu_m^2 + {\bfb^2}_m \right)}, \  {\bfb^2}_m = \frac{1}{1 + \bftau_m \left( \left\| \bfmu \right\|_2^2 + N a^2\right)}.
\end{equation}
In the M-step, the ELBO is given by
\begin{equation}\label{append_elbo_diff}
    \begin{aligned}
        & \ELBO \left( q; \tau, \beta; F \left( \cdot \right) \right) \\
        = & \bbE_{q} \left[ \log \rmPr \left( \bfY, \bfz, \bfw \right) \right] - \bbE_{q} \left[ \log q \left( \bfz, \bfw \right) \right] \\
        = & - \frac{NM}{2} \log\left( 2\pi \right) + \frac{N}{2} \sum_{m = 1}^{M} \log \bftau_{m} \\
        & - \sum_{m=1}^{M} \frac{\bftau_m}{2} \left( \left\| \bfY_{\cdot m} - \bfmu \bfnu_{m}^{\rmT} \right\|_{2}^2 + \left\| \left( \bfmu^2 + \diag \left( \bfA \right) \right) \left( \bfnu_{m}^2 + {\bfb^{2}}_m \right) - \bfmu^2 \bfnu_{m}^2 \right\|_{1,1} \right) \\
        & - \frac{N}{2} \log \left( 2 \pi \right) + \frac{N}{2} \log \beta - \frac{\beta}{2} \left( \left\| \bfmu - F \left( \bfX \right) \right\|_2^2 + \trace \left( \bfA \right) \right) \\
        & - \frac{M}{2} \log \left( 2 \pi \right) - \frac{1}{2} \left\| \bfnu \right\|_2^2 - \frac{1}{2} \trace \left( \bfB \right) \\
        & + \frac{N}{2} \log \left( 2 \pi \right) + \frac{1}{2} \log \det \left( \bfA \right) + \frac{N}{2} \\
        & + \frac{M}{2} \log \left( 2 \pi \right) + \frac{1}{2} \log \det \left( \bfB \right) + \frac{M}{2} \\
        = & \frac{N}{2} \sum_{m = 1}^{M} \log \bftau_{m} - \sum_{m=1}^{M} \frac{\bftau_m}{2} \left( \left\| \bfY_{\cdot m} - \bfmu \bfnu_{m}^{\rmT} \right\|_{2}^2 + \left\| \left( \bfmu^2 + \diag \left( \bfA \right) \right) \left( \bfnu_{m}^2 + {\bfb^{2}}_m \right) - \bfmu^2 \bfnu_{m}^2 \right\|_{1,1} \right) \\
        & + \frac{N}{2} \log \beta - \frac{\beta}{2} \left( \left\| \bfmu - F \left( \bfX \right) \right\|_2^2 + N a^2 \right) + \const.
    \end{aligned} 
\end{equation}
The update scheme for $\beta$ and the function $F \left( \cdot \right)$ will keep the same while that of $\bftau$ will become as
\begin{equation}\label{append_para_diff}
    \begin{aligned}
        \bftau_m = \frac{N}{\left\| \bfY_{\cdot m} - \bfmu \bfnu_{m}^{\rmT} \right\|_{2}^2 + \left\| \left( \bfmu^2 + \diag \left( \bfA \right) \right) \left( \bfnu_{m}^2 + {\bfb^{2}}_m \right) - \bfmu^2 \bfnu_{m}^2 \right\|_{1,1}}.
    \end{aligned} 
\end{equation}

\begin{figure}[!htb]
    \setlength{\abovecaptionskip}{-0.5cm}
    \begin{center}
        \includegraphics[width=0.9\textwidth]{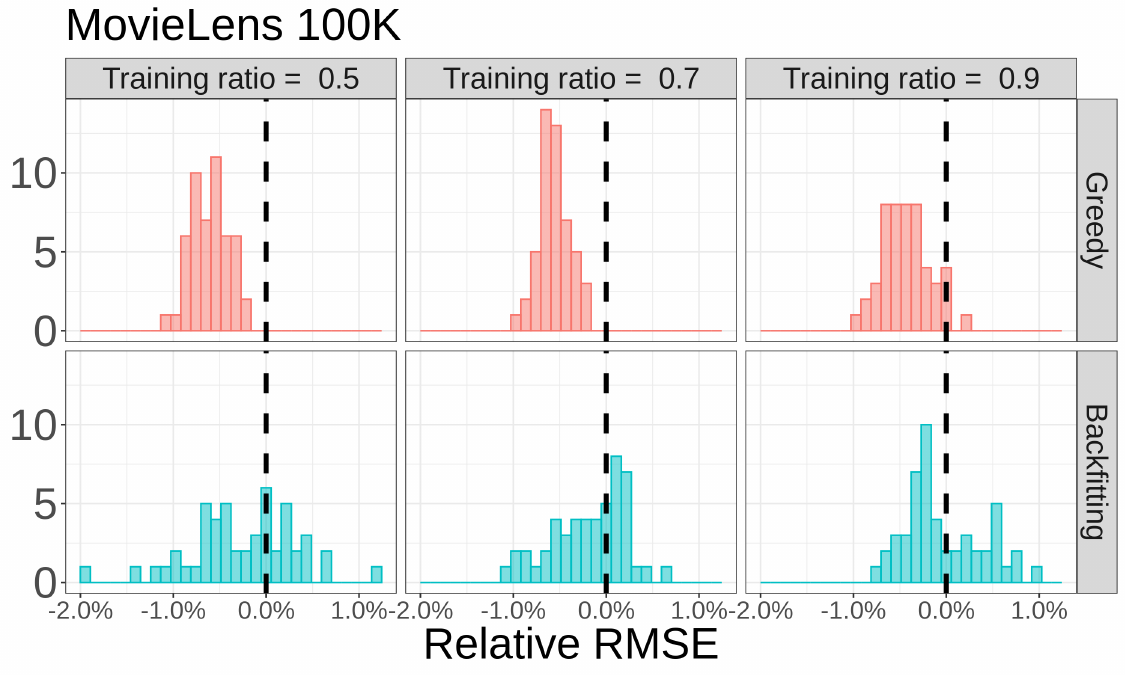}
    \end{center}
    \caption{Histogram plots of the relative RMSE of the method allowing different precision parameters compared to the original version of MFAI in 50 times experiments. The black dashed vertical line at zero represents the baseline MFAI. \label{append_fig:movie_differ_tau}}
\end{figure}

We also implemented the greedy and backfitting algorithms for this modeling way.
For a fair comparison with the original version of MFAI, we applied the two methods to the real dataset \textit{MovieLens 100K}.
To examine the imputation performance of the compared methods, we first randomly split the observed entries $\Omega^{\text{obs}}$ into a training set $\Omega^{\text{train}}$ and a test set $\Omega^{\text{test}}$.
Then, we applied matrix factorization methods to the training set and predicted the entries in the held-out set.
The imputation accuracy of the held-out entries was measured by the root mean squared error (RMSE).
We considered different values of the ``training ratio'' which is defined as $\frac{| \Omega^{\text{train}}| }{ |\Omega^{\text{obs}}| }$.
We repeated the experiments 50 times for each setting of the training ratio.
To make the comparison between the alternative methods more clear, we prepared the histogram plots of the relative RMSE of the method allowing different precision parameters compared to the original version of MFAI in Figure \ref{append_fig:movie_differ_tau}.
We found that in the case of poor data quality (weak signal and high sparsity), the method allowing different precision parameters between features cannot always achieve better imputation accuracy than the original version of MFAI.
It is worth mentioning that the setting of different precision parameters involves the flexibility of the model.
Allowing different $\bftau_m$ makes the model more flexible but also makes it much easier to overfit, especially when the data quality is poor.
 

\section{Fitting the Multi-Factor MFAI Model}
\label{append_sec:algo}

The greedy algorithm automatically selecting rank $K$ is summarised in Algorithm \ref{append_algo_select_K}.
To examine the performance of MFAI in the multi-factor case, we designed two sets of experiments in a wide range of data quality settings.
In Experiment 1, we fixed the missing ratio, $\frac{ \left| \Omega^{\text{miss}} \right| }{ NM } = 0.5$ and varied $\PVE \in \{0.1, 0.5, 0.9\}$ to examine the performances under different noise levels.
In Experiment 2, we fixed the $\PVE = 0.5$ and varied $\missing \in \{0, 0.5, 0.9\}$ to investigate the influence of data sparsity levels.
The true rank was set at $K = 3$ for both experiments.
We repeated the simulations $50$ times for each setting and set the maximum rank allowed $K_{\text{max}} = 10$ for MFAI.
The results of the inferred rank are summarized in \autoref{append_table:inferred_k}.
It is noticed that the number of factors is hard to determine when the data matrix is highly noisy or sparse.
We also note that the inferred rank is influenced by the parameter settings in the implementation, such as the shrinkage/null threshold and convergence criteria.

\begin{algorithm}[!ht]
    \label{append_algo_select_K}
    \caption{Greedy Algorithm Automatically Selecting $K$}
    \SetAlgoLined
    \KwData{main data matrix $\bfY$ and auxiliary matrix $\bfX$}
    \KwResult{estimate of the latent factors $\bfZ$ and loadings $\bfW$}
    \BlankLine
    set the maximum value of the number of factors $K_{\max}$ \;
    set the initial rank $K = 0$ \;
    set the stop criterion $sc$ \;
    \For{$k = 1, \dots, K_{\max}$}{
        $\bfmu_k, a^2_k; \bfnu_k, b^2_k; \tau_k, \beta_k; F_k \left( \cdot \right) \leftarrow$ MFAI\_SF$(\bfY, \bfX)$ \;
        \If{$\Var \left( \bfmu_k \bfnu_k^{\rmT} \right) \cdot \tau_k < sc$}{
            break \;
        }
        $\bfY \leftarrow \bfY - \bfmu_k \bfnu_k^{\rmT}$ \;
        $K \leftarrow K + 1$ \;
    }
    \KwRet{$\widehat{\bfZ} = \left( \bfmu_1, \dots, \bfmu_K \right), \widehat{\bfW} = \left( \bfnu_1, \dots, \bfnu_K \right); a_1^2, \dots, a_K^2, b_1^2, \dots, b_K^2$} \;
    \KwRet{$\tau_1, \dots, \tau_K; \beta_1, \dots, \beta_K$} \;
    \KwRet{$F_1 \left( \cdot \right), \dots, F_K \left( \cdot \right)$.}
\end{algorithm}

\begin{table}[!ht]
    \centering
    \begin{tabular}{|c|c|c|c|c|}
        \hline
        Experiment & Missing ratio & PVE & True rank & Rank inferred by MFAI \\
        \hline
        \multirow{3}{*}{Experiment 1} & \multirow{3}{*}{0.5} & 0.1 & \multirow{6}{*}{3} & 2 for 50 times \\
         & & 0.5 & & 3 for 50 times \\
         & & 0.9 & & 3 for 34 times and 4 for 16 times\\
        \cline{1-3} \cline{5-5}
        \multirow{3}{*}{Experiment 2} & No missingness & \multirow{3}{*}{0.5} & & 3 for 50 times \\
         & 0.5 & & & 3 for 50 times \\
         & 0.9 & & & 3 for 15 times and 2 for 35 times \\
        \hline
    \end{tabular}
    \caption{The inferred rank of MFAI.}
    \label{append_table:inferred_k}
\end{table}

\begin{figure}[!htb]
    \setlength{\abovecaptionskip}{-0.5cm}
    \begin{center}
        \includegraphics[width=0.9\textwidth]{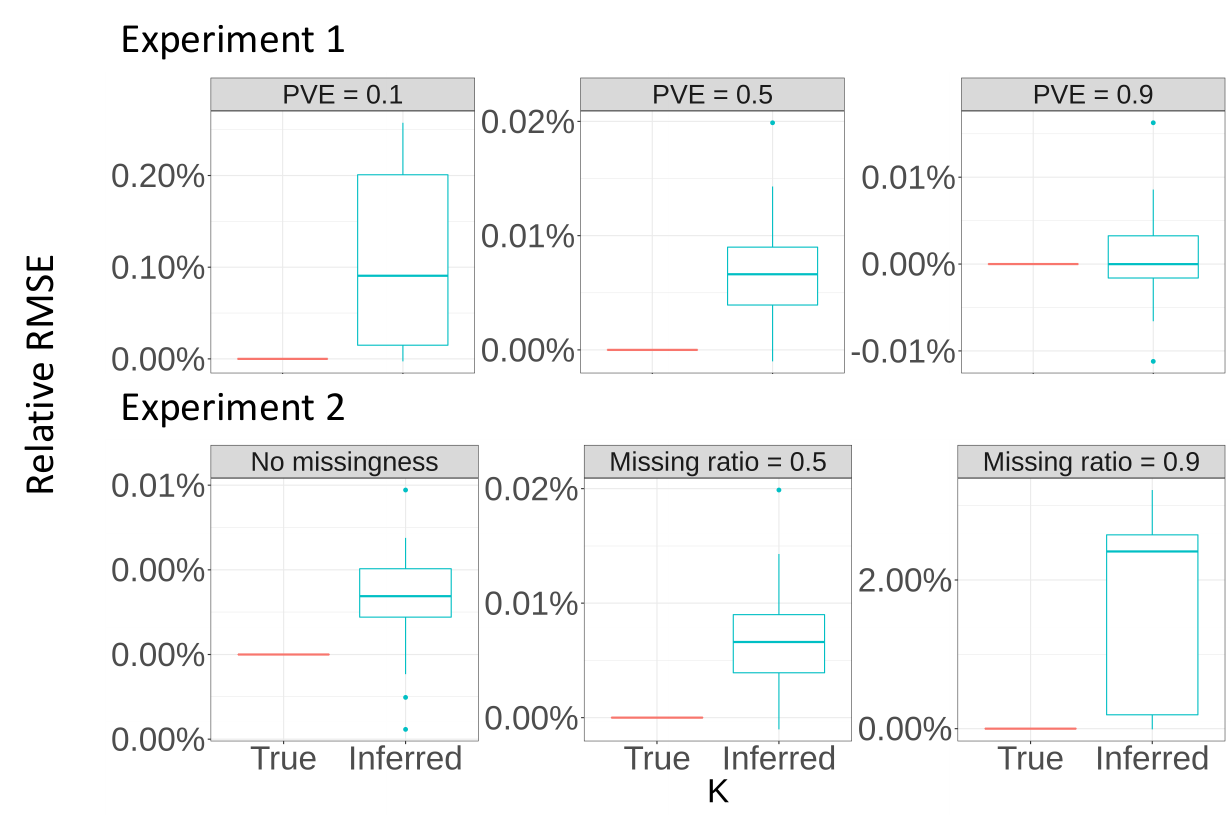}
    \end{center}
    \caption{Boxplots comparing the accuracy of imputing missing entries. Experiment 1 involves the main matrix $\bfY$ that varies from low signal strength (PVE = 0.1, left) to high signal strength (PVE = 0.9, right). Experiment 2 involves the main matrix $\bfY$ that varies from low sparsity (missing ratio = 0, left) to high sparsity (missing ratio = 0.9, right). Accuracy is measured using the relative RMSE with smaller values indicating higher accuracy. \label{append_fig:infer_k}}
\end{figure}

We further compared the imputation accuracy of MFAI, both when using the true rank directly and when inferring it by itself.
The results are summarized in \autoref{append_fig:infer_k}.
In most cases, the performance did not exhibit significant variations.
This was mainly because our null check procedure involved removing the factor and terminating the greedy algorithm when the currently inferred factor was close to zero compared to the estimated noise strength.
Consequently, the factors estimated in the tail had a negligible impact on the imputation accuracy.
However, we observed that when the data was highly sparse, the imputation accuracy differed more due to the underestimated rank.
This finding suggests that in the empirical Bayes framework, overestimation is generally preferable to underestimation in terms of imputation accuracy.

\section{Human Brain Gene Expression Data}
\label{append_sec:brain}

The human brain gene expression dataset was generated from 1,340 tissue samples collected from 57 developing and adult post-mortem brains of clinically unremarkable donors representing males and females of multiple ethnicities \citep{johnson2009functional, kang2011spatio}.

\subsection{Spatial and Temporal Information}

A 15-period system spanning the periods from embryonic development to late adulthood was created to investigate the spatial-temporal dynamics of the human brain transcriptome (\autoref{append_table:brain_period}).
\begin{table}[!ht]
    \centering
    \begin{tabular}{ccc}
        \hline 
        Period & Description & Age \\
        \hline 
        \hline
        1 & Embryonic & $4 \ \mathrm{PCW} \leq \mathrm{Age} < 8 \ \mathrm{PCW}$ \\
        2 & Early fetal & $8 \ \mathrm{PCW} \leq \mathrm{Age} < 10 \ \mathrm{PCW}$ \\
        3 & Early fetal & $10 \ \mathrm{PCW} \leq \mathrm{Age} < 13 \ \mathrm{PCW}$ \\
        4 & Early mid-fetal & $13 \ \mathrm{PCW} \leq \mathrm{Age} < 16 \ \mathrm{PCW}$ \\
        5 & Early mid-fetal & $16 \ \mathrm{PCW} \leq \mathrm{Age} < 19 \ \mathrm{PCW}$ \\
        6 & Late mid-fetal & $19 \ \mathrm{PCW} \leq \mathrm{Age} < 24 \ \mathrm{PCW}$ \\
        7 & Late fetal & $24 \ \mathrm{PCW} \leq \mathrm{Age} < 38 \ \mathrm{PCW}$ \\
        8 & Neonatal and early infancy & $0 \ \mathrm{M} \ \mathrm{(birth)} \leq \mathrm{Age} < 6 \ \mathrm{M}$ \\
        9 & Late infancy & $6 \mathrm{M} \ \leq \mathrm{Age} < 12 \ \mathrm{M}$ \\
        10 & Early childhood & $1 \ \mathrm{Y} \leq \mathrm{Age} < 6 \ \mathrm{Y}$ \\
        11 & Middle and late childhood & $6 \ \mathrm{Y} \leq \mathrm{Age} < 12 \ \mathrm{Y}$ \\
        12 & Adolescence & $12 \ \mathrm{Y} \leq \mathrm{Age} < 20 \ \mathrm{Y}$ \\
        13 & Young adulthood & $20 \ \mathrm{Y} \leq \mathrm{Age} < 40 \ \mathrm{Y}$ \\
        14 & Middle adulthood & $40 \ \mathrm{Y} \leq \mathrm{Age} < 60 \ \mathrm{Y}$ \\
        15 & Late adulthood & $60 \ \mathrm{Y} \leq \mathrm{Age}$ \\
        \hline
    \end{tabular}
    \begin{tablenotes}
        \centering
        \small
        \item M, postnatal months; PCW, post-conceptional weeks; Y, postnatal years.
    \end{tablenotes}
    \caption{Periods of human development and adulthood as defined in \citet{johnson2009functional, kang2011spatio}.}
    \label{append_table:brain_period}
\end{table}
The transient prenatal structures and immature and mature forms of 16 brain regions, including 11 neocortex areas, were sampled from multiple specimens per period (\autoref{append_table:brain_region}).
\begin{table}[!ht]
    \centering
    \begin{tabular}{ccc}
        \hline 
        Periods 1 and 2 & Periods 3-15 \\
        \hline
        \hline
        FC, frontal cerebral wall & OFC, orbital prefrontal cortex \\
        & DFC, dorsolateral prefrontal cortex \\
        & VFC, ventrolateral prefrontal cortex \\
        & MFC, medial prefrontal cortex \\
        & M1C, primary motor (M1) cortex \\
        \hline 
        PC, parietal cerebral wall & S1C, primary somatosensory (S1) cortex \\
        & IPC, posterior inferior parietal cortex \\
        \hline 
        TC, temporal cerebral wall & A1C, primary auditory (A1) cortex \\
        & STC, superior temporal cortex \\
        & ITC, inferior temporal cortex \\
        \hline 
        OC, occipital cerebral wall & V1C, primary visual (V1) cortex \\
        \hline 
        HIP, hippocampal anlage & HIP, hippocampus \\
        \hline 
        --- & AMY, amygdala \\
        \hline
        VF, ventral forebrain & STR, striatum \\
        MGE, medial ganglionic eminence & \\
        LGE, lateral ganglionic eminence & \\
        CGE, caudal ganglionic eminence & \\
        \hline 
        DIE, diencephalon & MD, mediodorsal nucleus of the thalamus \\
        DTH, dorsal thalamus & --- \\
        \hline 
        URL, upper (rostral) rhombic lip & CBC, cerebellar cortex \\
        \hline
    \end{tabular}
    \caption{Ontology and nomenclature of analyzed brain regions and neocortex areas in \citet{johnson2009functional, kang2011spatio}.}
    \label{append_table:brain_region}
\end{table}
It's obvious that periods 1 and 2 correspond to embryonic and early fetal development when most of the 16 brain regions sampled in future periods have not differentiated (i.e., most of the 16 brain regions are missing data in periods 1 and 2).
Therefore, data in periods 1 and 2 are excluded from our analysis.
In this article, we focus on analyzing the neocortex areas, including the orbital prefrontal cortex (OFC), dorsolateral prefrontal cortex (DFC), ventrolateral prefrontal cortex (VFC), medial prefrontal cortex (MFC), primary motor cortex (M1C), primary somatosensory cortex (S1C), posterior inferior parietal cortex (IPC), primary auditory cortex (A1C), posterior superior temporal cortex (STC), inferior temporal cortex (ITC), and primary visual cortex (V1C).

\subsection{Differential Stability}

Differential stability (DS) is defined as the tendency for a gene to exhibit reproducible differential expression relationships across brain structures \citep{shaw2011preservation}.
DS can be measured by a variety of different means, such as the average Kendall-Tau correlation, average Spearman correlation, and average Euclidean difference (see \cite{hawrylycz2015canonical} Supplementary Analysis for comparable metrics).
Here we use the average Pearson correlation as an example.
Suppose the whole brain gene expression dataset contains $B$ individual brains, and we need to calculate the average Pearson correlation over $B(B-1) / 2$ pairs of $B$ brains to measure DS between brains.
We can represent the previous gene expression data matrix to a tensor $\bfX \in \bbR^{B \times S \times G}$, where $B$ denotes the individual brains, $S$ denotes the anatomic structures (refer to the neocortex areas in our analysis), and $G$ denotes the genes as in the main text.
For each gene $g$ expressed in a pair of brains (indexed by $b_{1}$ and $b_{2}$) $\bfX_{b_{1} \cdot g}$ and $\bfX_{b_{2} \cdot g}$, let $\left\{ \left( \bfX_{b_{1} 1 g}, \bfX_{b_{2} 1 g} \right), \dots , \left( \bfX_{b_{1} S g}, \bfX_{b_{2} S g} \right) \right\}$ be $S$ pairs expression level measurements in $S$ common anatomic structures and we computed the Pearson correlation $\rho_{S} \left( \bfX_{b_{1} \cdot g}, \bfX_{b_{2} \cdot g} \right)$.
Then the DS of gene $g$ is defined as the average $\rho_{S}$ over $B(B-1) / 2$ pairs of brains.
For the differentially expressed gene $g$, intuitively, $\bfX_{b_{1} s_{1} g} > \bfX_{b_{1} s_{2} g}$ is expected if $\bfX_{b_{2} s_{1} g} > \bfX_{b_{2} s_{2} g}$ or $\bfX_{b_{1} s_{1} g} < \bfX_{b_{1} s_{2} g}$ is expected if $\bfX_{b_{2} s_{1} g} < \bfX_{b_{2} s_{2} g}$, for any $b_{1}, b_{2} \in \left\{ 1, \dots, B \right\}$ and any $s_{1}, s_{2} \in \left\{ 1, \dots, S \right\}$.
In summary, DS is a correlation-based metric that can be applied to assess the reproducibility of gene expression patterns across various structures in various individual brains and help reveal mesoscale genetic organization.
Recent studies have shown that the genes with the highest DS are highly biologically relevant, with enrichment for brain-related annotations, disease associations, drug targets, and literature citations.

\bibliographystyle{chicago}
\bibliography{main}

\begin{thebibliography}{}

\bibitem[\protect\citeauthoryear{Agarwal and Chen}{Agarwal and
  Chen}{2009}]{agarwal2009regression}
Agarwal, D. and B.-C. Chen (2009).
\newblock Regression-based latent factor models.
\newblock In {\em Proceedings of the 15th ACM SIGKDD International Conference
  on Knowledge Discovery and Data Mining}, pp.\  19–28.

\bibitem[\protect\citeauthoryear{Aleksander, Balhoff, Carbon, Cherry, Drabkin,
  Ebert, Feuermann, Gaudet, Harris, et~al.}{Aleksander
  et~al.}{2023}]{gene2023gene}
Aleksander, S.~A., J.~Balhoff, S.~Carbon, J.~M. Cherry, H.~J. Drabkin,
  D.~Ebert, M.~Feuermann, P.~Gaudet, N.~L. Harris, et~al. (2023).
\newblock The {Gene Ontology} knowledgebase in 2023.
\newblock {\em Genetics\/}~{\em 224\/}(1), iyad031.

\bibitem[\protect\citeauthoryear{Ashburner, Ball, Blake, Botstein, Butler,
  Cherry, Davis, Dolinski, Dwight, Eppig, et~al.}{Ashburner
  et~al.}{2000}]{ashburner2000gene}
Ashburner, M., C.~A. Ball, J.~A. Blake, D.~Botstein, H.~Butler, J.~M. Cherry,
  A.~P. Davis, K.~Dolinski, S.~S. Dwight, J.~T. Eppig, et~al. (2000).
\newblock Gene {ontology}: Tool for the unification of biology.
\newblock {\em Nature Genetics\/}~{\em 25\/}(1), 25--29.

\bibitem[\protect\citeauthoryear{Babacan, Luessi, Molina, and
  Katsaggelos}{Babacan et~al.}{2012}]{babacan2012sparse}
Babacan, S.~D., M.~Luessi, R.~Molina, and A.~K. Katsaggelos (2012).
\newblock Sparse {Bayesian} methods for low-rank matrix estimation.
\newblock {\em IEEE Transactions on Signal Processing\/}~{\em 60\/}(8),
  3964--3977.

\bibitem[\protect\citeauthoryear{Bell and Koren}{Bell and
  Koren}{2007}]{bell2007lessons}
Bell, R.~M. and Y.~Koren (2007).
\newblock Lessons from the {Netflix} prize challenge.
\newblock {\em ACM SIGKDD Explorations Newsletter\/}~{\em 9\/}(2), 75--79.

\bibitem[\protect\citeauthoryear{Belousov and Fontes}{Belousov and
  Fontes}{2013}]{belousov2013neuronal}
Belousov, A.~B. and J.~D. Fontes (2013).
\newblock Neuronal gap junctions: Making and breaking connections during
  development and injury.
\newblock {\em Trends in Neurosciences\/}~{\em 36\/}(4), 227--236.

\bibitem[\protect\citeauthoryear{Bishop}{Bishop}{2006}]{bishop2006pattern}
Bishop, C.~M. (2006).
\newblock {\em Pattern Recognition and Machine Learning}.
\newblock Springer.

\bibitem[\protect\citeauthoryear{Blei, Kucukelbir, and McAuliffe}{Blei
  et~al.}{2017}]{blei2017variational}
Blei, D.~M., A.~Kucukelbir, and J.~D. McAuliffe (2017).
\newblock Variational inference: A review for statisticians.
\newblock {\em Journal of the American Statistical Association\/}~{\em
  112\/}(518), 859--877.

\bibitem[\protect\citeauthoryear{Breiman}{Breiman}{1984}]{breiman1984classification}
Breiman, L. (1984).
\newblock {\em Classification and Regression Trees}.
\newblock Wadsworth International Group.

\bibitem[\protect\citeauthoryear{Breiman}{Breiman}{1998}]{breiman1998arcing}
Breiman, L. (1998).
\newblock Arcing classifier (with discussion and a rejoinder by the author).
\newblock {\em The Annals of Statistics\/}~{\em 26\/}(3), 801--849.

\bibitem[\protect\citeauthoryear{Breiman and Friedman}{Breiman and
  Friedman}{1985}]{breiman1985estimating}
Breiman, L. and J.~H. Friedman (1985).
\newblock Estimating optimal transformations for multiple regression and
  correlation.
\newblock {\em Journal of the American Statistical Association\/}~{\em
  80\/}(391), 580--598.

\bibitem[\protect\citeauthoryear{B{\"u}hlmann and Hothorn}{B{\"u}hlmann and
  Hothorn}{2007}]{buhlmann2007boosting}
B{\"u}hlmann, P. and T.~Hothorn (2007).
\newblock Boosting algorithms: Regularization, prediction and model fitting.
\newblock {\em Statistical Science\/}~{\em 22\/}(4), 477--505.

\bibitem[\protect\citeauthoryear{Cai, Cand{\`e}s, and Shen}{Cai
  et~al.}{2010}]{cai2010singular}
Cai, J.-F., E.~J. Cand{\`e}s, and Z.~Shen (2010).
\newblock A singular value thresholding algorithm for matrix completion.
\newblock {\em SIAM Journal on Optimization\/}~{\em 20\/}(4), 1956--1982.

\bibitem[\protect\citeauthoryear{Cand{\`e}s and Recht}{Cand{\`e}s and
  Recht}{2009}]{candes2009exact}
Cand{\`e}s, E.~J. and B.~Recht (2009).
\newblock Exact matrix completion via convex optimization.
\newblock {\em Foundations of Computational Mathematics\/}~{\em 9\/}(6),
  717--772.

\bibitem[\protect\citeauthoryear{Cand{\`e}s and Tao}{Cand{\`e}s and
  Tao}{2010}]{candes2010power}
Cand{\`e}s, E.~J. and T.~Tao (2010).
\newblock The power of convex relaxation: Near-optimal matrix completion.
\newblock {\em IEEE Transactions on Information Theory\/}~{\em 56\/}(5),
  2053--2080.

\bibitem[\protect\citeauthoryear{Cortes}{Cortes}{2018}]{cortes2018cold}
Cortes, D. (2018).
\newblock Cold-start recommendations in collective matrix factorization.
\newblock {\em arXiv preprint arXiv:1809.00366\/}.

\bibitem[\protect\citeauthoryear{de~Lores~Arnaiz and Ordieres}{de~Lores~Arnaiz
  and Ordieres}{2014}]{de2014brain}
de~Lores~Arnaiz, G.~R. and M.~G.~L. Ordieres (2014).
\newblock Brain {Na+, K+-ATPase} activity in aging and disease.
\newblock {\em International Journal of Biomedical Science\/}~{\em 10\/}(2),
  85.

\bibitem[\protect\citeauthoryear{Elith, Leathwick, and Hastie}{Elith
  et~al.}{2008}]{elith2008working}
Elith, J., J.~R. Leathwick, and T.~Hastie (2008).
\newblock A working guide to boosted regression trees.
\newblock {\em Journal of Animal Ecology\/}~{\em 77\/}(4), 802--813.

\bibitem[\protect\citeauthoryear{Engelhardt and Stephens}{Engelhardt and
  Stephens}{2010}]{engelhardt2010analysis}
Engelhardt, B.~E. and M.~Stephens (2010).
\newblock Analysis of population structure: A unifying framework and novel
  methods based on sparse factor analysis.
\newblock {\em PLoS Genetics\/}~{\em 6\/}(9), e1001117.

\bibitem[\protect\citeauthoryear{Fithian and Mazumder}{Fithian and
  Mazumder}{2018}]{fithian2018flexible}
Fithian, W. and R.~Mazumder (2018).
\newblock Flexible low-rank statistical modeling with missing data and side
  information.
\newblock {\em Statistical Science\/}~{\em 33\/}(2), 238--260.

\bibitem[\protect\citeauthoryear{Freund and Schapire}{Freund and
  Schapire}{1996}]{freund1996experiments}
Freund, Y. and R.~E. Schapire (1996).
\newblock Experiments with a new boosting algorithm.
\newblock In {\em Proceedings of the Thirteenth International Conference on
  Machine Learning}, pp.\  148--156.

\bibitem[\protect\citeauthoryear{Friedman, Hastie, and Tibshirani}{Friedman
  et~al.}{2000}]{friedman2000additive}
Friedman, J., T.~Hastie, and R.~Tibshirani (2000).
\newblock Additive logistic regression: A statistical view of boosting (with
  discussion and a rejoinder by the authors).
\newblock {\em The Annals of Statistics\/}~{\em 28\/}(2), 337--407.

\bibitem[\protect\citeauthoryear{Friedman}{Friedman}{2001}]{friedman2001greedy}
Friedman, J.~H. (2001).
\newblock Greedy function approximation: A gradient boosting machine.
\newblock {\em The Annals of Statistics\/}~{\em 29\/}(5), 1189--1232.

\bibitem[\protect\citeauthoryear{G{\"o}nen and Kaski}{G{\"o}nen and
  Kaski}{2014}]{gonen2014kernelized}
G{\"o}nen, M. and S.~Kaski (2014).
\newblock Kernelized {Bayesian} matrix factorization.
\newblock {\em IEEE Transactions on Pattern Analysis and Machine
  Intelligence\/}~{\em 36\/}(10), 2047--2060.

\bibitem[\protect\citeauthoryear{Grinsztajn, Oyallon, and Varoquaux}{Grinsztajn
  et~al.}{2022}]{grinsztajn2022tree}
Grinsztajn, L., E.~Oyallon, and G.~Varoquaux (2022).
\newblock Why do tree-based models still outperform deep learning on typical
  tabular data?
\newblock In {\em Thirty-sixth Conference on Neural Information Processing
  Systems Datasets and Benchmarks Track}.

\bibitem[\protect\citeauthoryear{Harper and Konstan}{Harper and
  Konstan}{2015}]{harper2015movielens}
Harper, F.~M. and J.~A. Konstan (2015).
\newblock The {MovieLens} datasets: History and context.
\newblock {\em ACM Transactions on Interactive Intelligent Systems\/}~{\em
  5\/}(4), 1--19.

\bibitem[\protect\citeauthoryear{Hastie, Mazumder, Lee, and Zadeh}{Hastie
  et~al.}{2015}]{hastie2015matrix}
Hastie, T., R.~Mazumder, J.~D. Lee, and R.~Zadeh (2015).
\newblock Matrix completion and low-rank {SVD} via fast alternating least
  squares.
\newblock {\em Journal of Machine Learning Research\/}~{\em 16\/}(104),
  3367--3402.

\bibitem[\protect\citeauthoryear{Hawrylycz, Miller, Menon, Feng, Dolbeare,
  Guillozet-Bongaarts, Jegga, Aronow, Lee, Bernard, et~al.}{Hawrylycz
  et~al.}{2015}]{hawrylycz2015canonical}
Hawrylycz, M., J.~A. Miller, V.~Menon, D.~Feng, T.~Dolbeare, A.~L.
  Guillozet-Bongaarts, A.~G. Jegga, B.~J. Aronow, C.-K. Lee, A.~Bernard, et~al.
  (2015).
\newblock Canonical genetic signatures of the adult human brain.
\newblock {\em Nature Neuroscience\/}~{\em 18\/}(12), 1832--1844.

\bibitem[\protect\citeauthoryear{Hubbard and Hegde}{Hubbard and
  Hegde}{2017}]{hubbard2017parallel}
Hubbard, C. and C.~Hegde (2017).
\newblock Parallel computing heuristics for low-rank matrix completion.
\newblock In {\em 2017 IEEE Global Conference on Signal and Information
  Processing}, pp.\  764--768. IEEE.

\bibitem[\protect\citeauthoryear{Ilin and Raiko}{Ilin and
  Raiko}{2010}]{ilin2010practical}
Ilin, A. and T.~Raiko (2010).
\newblock Practical approaches to principal component analysis in the presence
  of missing values.
\newblock {\em Journal of Machine Learning Research\/}~{\em 11}, 1957--2000.

\bibitem[\protect\citeauthoryear{Johnson, Kawasawa, Mason, Krsnik, Coppola,
  Bogdanovi{\'c}, Geschwind, Mane, {\v{S}}estan, et~al.}{Johnson
  et~al.}{2009}]{johnson2009functional}
Johnson, M.~B., Y.~I. Kawasawa, C.~E. Mason, {\v{Z}}.~Krsnik, G.~Coppola,
  D.~Bogdanovi{\'c}, D.~H. Geschwind, S.~M. Mane, N.~{\v{S}}estan, et~al.
  (2009).
\newblock Functional and evolutionary insights into human brain development
  through global transcriptome analysis.
\newblock {\em Neuron\/}~{\em 62\/}(4), 494--509.

\bibitem[\protect\citeauthoryear{Kang, Kawasawa, Cheng, Zhu, Xu, Li, Sousa,
  Pletikos, Meyer, Sedmak, et~al.}{Kang et~al.}{2011}]{kang2011spatio}
Kang, H.~J., Y.~I. Kawasawa, F.~Cheng, Y.~Zhu, X.~Xu, M.~Li, A.~M. Sousa,
  M.~Pletikos, K.~A. Meyer, G.~Sedmak, et~al. (2011).
\newblock Spatio-temporal transcriptome of the human brain.
\newblock {\em Nature\/}~{\em 478\/}(7370), 483--489.

\bibitem[\protect\citeauthoryear{Kinjo, Arida, de~Oliveira, and
  da~Silva~Fernandes}{Kinjo et~al.}{2007}]{kinjo2007na}
Kinjo, {\'E}.~R., R.~M. Arida, D.~M. de~Oliveira, and M.~J. da~Silva~Fernandes
  (2007).
\newblock The {Na+/K+ ATPase} activity is increased in the hippocampus after
  multiple status epilepticus induced by pilocarpine in developing rats.
\newblock {\em Brain Research\/}~{\em 1138}, 203--207.

\bibitem[\protect\citeauthoryear{Koren, Bell, and Volinsky}{Koren
  et~al.}{2009}]{koren2009matrix}
Koren, Y., R.~Bell, and C.~Volinsky (2009).
\newblock Matrix factorization techniques for recommender systems.
\newblock {\em Computer\/}~{\em 42\/}(8), 30--37.

\bibitem[\protect\citeauthoryear{Kullback and Leibler}{Kullback and
  Leibler}{1951}]{kullback1951information}
Kullback, S. and R.~Leibler (1951).
\newblock On information and sufficiency.
\newblock {\em The Annals of Mathematical Statistics\/}~{\em 22\/}(1), 79--86.

\bibitem[\protect\citeauthoryear{Lin, Sanders, Li, Sestan, State, and Zhao}{Lin
  et~al.}{2015}]{lin2015markov}
Lin, Z., S.~J. Sanders, M.~Li, N.~Sestan, M.~W. State, and H.~Zhao (2015).
\newblock A {Markov} random field-based approach to characterizing human brain
  development using spatial--temporal transcriptome data.
\newblock {\em The Annals of Applied Statistics\/}~{\em 9\/}(1), 429--451.

\bibitem[\protect\citeauthoryear{Lin, Wang, Yang, and Zhao}{Lin
  et~al.}{2017}]{lin2017joint}
Lin, Z., T.~Wang, C.~Yang, and H.~Zhao (2017).
\newblock On joint estimation of {Gaussian} graphical models for spatial and
  temporal data.
\newblock {\em Biometrics\/}~{\em 73\/}(3), 769--779.

\bibitem[\protect\citeauthoryear{Lin, Yang, Zhu, Duchi, Fu, Wang, Jiang,
  Zamanighomi, Xu, Li, et~al.}{Lin et~al.}{2016}]{lin2016simultaneous}
Lin, Z., C.~Yang, Y.~Zhu, J.~Duchi, Y.~Fu, Y.~Wang, B.~Jiang, M.~Zamanighomi,
  X.~Xu, M.~Li, et~al. (2016).
\newblock Simultaneous dimension reduction and adjustment for confounding
  variation.
\newblock {\em Proceedings of the National Academy of Sciences\/}~{\em
  113\/}(51), 14662--14667.

\bibitem[\protect\citeauthoryear{Little and Rubin}{Little and
  Rubin}{1987}]{little1987statistical}
Little, R.~J. and D.~B. Rubin (1987).
\newblock {\em Statistical Analysis with Missing Data}.
\newblock John Wiley \& Sons.

\bibitem[\protect\citeauthoryear{Liu, Yuan, and Zhao}{Liu
  et~al.}{2022}]{liu2022characterizing}
Liu, T., M.~Yuan, and H.~Zhao (2022).
\newblock Characterizing spatiotemporal transcriptome of the human brain via
  low-rank tensor decomposition.
\newblock {\em Statistics in Biosciences\/}~{\em 14\/}(3), 485--513.

\bibitem[\protect\citeauthoryear{L{\'o}pez-Ot{\'\i}n, Blasco, Partridge,
  Serrano, and Kroemer}{L{\'o}pez-Ot{\'\i}n et~al.}{2013}]{lopez2013hallmarks}
L{\'o}pez-Ot{\'\i}n, C., M.~A. Blasco, L.~Partridge, M.~Serrano, and G.~Kroemer
  (2013).
\newblock The hallmarks of aging.
\newblock {\em Cell\/}~{\em 153\/}(6), 1194--1217.

\bibitem[\protect\citeauthoryear{Luebke, Chang, Moore, and Rosene}{Luebke
  et~al.}{2004}]{luebke2004normal}
Luebke, J., Y.-M. Chang, T.~Moore, and D.~Rosene (2004).
\newblock Normal aging results in decreased synaptic excitation and increased
  synaptic inhibition of layer 2/3 pyramidal cells in the monkey prefrontal
  cortex.
\newblock {\em Neuroscience\/}~{\em 125\/}(1), 277--288.

\bibitem[\protect\citeauthoryear{Luebke and Rosene}{Luebke and
  Rosene}{2003}]{luebke2003aging}
Luebke, J.~I. and D.~L. Rosene (2003).
\newblock Aging alters dendritic morphology, input resistance, and inhibitory
  signaling in dentate granule cells of the rhesus monkey.
\newblock {\em Journal of Comparative Neurology\/}~{\em 460\/}(4), 573--584.

\bibitem[\protect\citeauthoryear{MacKay}{MacKay}{1995}]{mackay1995probable}
MacKay, D.~J. (1995).
\newblock Probable networks and plausible predictions-a review of practical
  {Bayesian} methods for supervised neural networks.
\newblock {\em Network Computation in Neural Systems\/}~{\em 6\/}(3), 469--505.

\bibitem[\protect\citeauthoryear{Mason, Baxter, Bartlett, and Frean}{Mason
  et~al.}{1999}]{mason1999boosting}
Mason, L., J.~Baxter, P.~Bartlett, and M.~Frean (1999).
\newblock Boosting algorithms as gradient descent.
\newblock In {\em Proceedings of the 12th International Conference on Neural
  Information Processing Systems}, pp.\  512–518.

\bibitem[\protect\citeauthoryear{Mazumder, Hastie, and Tibshirani}{Mazumder
  et~al.}{2010}]{mazumder2010spectral}
Mazumder, R., T.~Hastie, and R.~Tibshirani (2010).
\newblock Spectral regularization algorithms for learning large incomplete
  matrices.
\newblock {\em Journal of Machine Learning Research\/}~{\em 11\/}(80),
  2287--2322.

\bibitem[\protect\citeauthoryear{Montoro and Yuste}{Montoro and
  Yuste}{2004}]{montoro2004gap}
Montoro, R.~J. and R.~Yuste (2004).
\newblock Gap junctions in developing neocortex: A review.
\newblock {\em Brain Research Reviews\/}~{\em 47\/}(1-3), 216--226.

\bibitem[\protect\citeauthoryear{Neal}{Neal}{1993}]{neal1993probabilistic}
Neal, R.~M. (1993).
\newblock Probabilistic inference using {Markov} chain {Monte} {Carlo} methods.
\newblock Technical report, Department of Computer Science, University of
  Toronto.

\bibitem[\protect\citeauthoryear{Neal}{Neal}{1996}]{neal1996bayesian}
Neal, R.~M. (1996).
\newblock {\em Bayesian Learning for Neural Networks}.
\newblock Springer-Verlag.

\bibitem[\protect\citeauthoryear{Park, Kim, and Choi}{Park
  et~al.}{2013}]{park2013hierarchical}
Park, S., Y.-D. Kim, and S.~Choi (2013).
\newblock Hierarchical {Bayesian} matrix factorization with side information.
\newblock In {\em Proceedings of the Twenty-Third International Joint
  Conference on Artificial Intelligence}, pp.\  1593--1599.

\bibitem[\protect\citeauthoryear{Porteous, Asuncion, and Welling}{Porteous
  et~al.}{2010}]{porteous2010bayesian}
Porteous, I., A.~Asuncion, and M.~Welling (2010).
\newblock Bayesian matrix factorization with side information and {Dirichlet}
  process mixtures.
\newblock In {\em Proceedings of the Twenty-Fourth AAAI Conference on
  Artificial Intelligence}, Volume~24, pp.\  563--568.

\bibitem[\protect\citeauthoryear{Rasmussen, O’Donnell, Ding, and
  Nedergaard}{Rasmussen et~al.}{2020}]{rasmussen2020interstitial}
Rasmussen, R., J.~O’Donnell, F.~Ding, and M.~Nedergaard (2020).
\newblock Interstitial ions: A key regulator of state-dependent neural
  activity?
\newblock {\em Progress in Neurobiology\/}~{\em 193}, 101802.

\bibitem[\protect\citeauthoryear{Rennie and Srebro}{Rennie and
  Srebro}{2005}]{rennie2005fast}
Rennie, J.~D. and N.~Srebro (2005).
\newblock Fast maximum margin matrix factorization for collaborative
  prediction.
\newblock In {\em Proceedings of the 22nd International Conference on Machine
  Learning}, pp.\  713--719.

\bibitem[\protect\citeauthoryear{Rubin}{Rubin}{1976}]{rubin1976inference}
Rubin, D.~B. (1976).
\newblock Inference and missing data.
\newblock {\em Biometrika\/}~{\em 63\/}(3), 581--592.

\bibitem[\protect\citeauthoryear{Salakhutdinov and Mnih}{Salakhutdinov and
  Mnih}{2007}]{salakhutdinov2007probabilistic}
Salakhutdinov, R. and A.~Mnih (2007).
\newblock Probabilistic matrix factorization.
\newblock In {\em Proceedings of the 20th International Conference on Neural
  Information Processing Systems}, pp.\  1257–1264.

\bibitem[\protect\citeauthoryear{Salakhutdinov and Mnih}{Salakhutdinov and
  Mnih}{2008}]{salakhutdinov2008bayesian}
Salakhutdinov, R. and A.~Mnih (2008).
\newblock Bayesian probabilistic matrix factorization using {Markov chain Monte
  Carlo}.
\newblock In {\em Proceedings of the 25th International Conference on Machine
  Learning}, pp.\  880–887.

\bibitem[\protect\citeauthoryear{Shang and Zhou}{Shang and
  Zhou}{2022}]{shang2022spatially}
Shang, L. and X.~Zhou (2022).
\newblock Spatially aware dimension reduction for spatial transcriptomics.
\newblock {\em Nature Communications\/}~{\em 13\/}(1), 7203.

\bibitem[\protect\citeauthoryear{Shaw, Shih, Chen, and Hwang}{Shaw
  et~al.}{2011}]{shaw2011preservation}
Shaw, G.~T., E.~S. Shih, C.-H. Chen, and M.-J. Hwang (2011).
\newblock Preservation of ranking order in the expression of human housekeeping
  genes.
\newblock {\em PLoS ONE\/}~{\em 6\/}(12), e29314.

\bibitem[\protect\citeauthoryear{Sigrist}{Sigrist}{2021}]{sigrist2021gradient}
Sigrist, F. (2021).
\newblock Gradient and {Newton} boosting for classification and regression.
\newblock {\em Expert Systems With Applications\/}~{\em 167}, 114080.

\bibitem[\protect\citeauthoryear{Sigrist}{Sigrist}{2022a}]{sigrist2020gaussian}
Sigrist, F. (2022a).
\newblock Gaussian process boosting.
\newblock {\em Journal of Machine Learning Research\/}~{\em 23\/}(232), 1--46.

\bibitem[\protect\citeauthoryear{Sigrist}{Sigrist}{2022b}]{sigrist2022latent}
Sigrist, F. (2022b).
\newblock Latent {Gaussian} model boosting.
\newblock {\em IEEE Transactions on Pattern Analysis and Machine
  Intelligence\/}~{\em 45\/}(2), 1894--1905.

\bibitem[\protect\citeauthoryear{Singh and Gordon}{Singh and
  Gordon}{2008}]{singh2008relational}
Singh, A.~P. and G.~J. Gordon (2008).
\newblock Relational learning via collective matrix factorization.
\newblock In {\em Proceedings of the 14th ACM SIGKDD International Conference
  on Knowledge Discovery and Data Mining}, pp.\  650--658.

\bibitem[\protect\citeauthoryear{Srebro, Rennie, and Jaakkola}{Srebro
  et~al.}{2004}]{srebro2004maximum}
Srebro, N., J.~D. Rennie, and T.~S. Jaakkola (2004).
\newblock Maximum-margin matrix factorization.
\newblock In {\em Proceedings of the 17th International Conference on Neural
  Information Processing Systems}, pp.\  1329--1336.

\bibitem[\protect\citeauthoryear{Sutor and Hagerty}{Sutor and
  Hagerty}{2005}]{sutor2005involvement}
Sutor, B. and T.~Hagerty (2005).
\newblock Involvement of gap junctions in the development of the neocortex.
\newblock {\em Biochimica et Biophysica Acta (BBA)-Biomembranes\/}~{\em
  1719\/}(1-2), 59--68.

\bibitem[\protect\citeauthoryear{Tak{\'a}cs, Pil{\'a}szy, N{\'e}meth, and
  Tikk}{Tak{\'a}cs et~al.}{2009}]{takacs2009scalable}
Tak{\'a}cs, G., I.~Pil{\'a}szy, B.~N{\'e}meth, and D.~Tikk (2009).
\newblock Scalable collaborative filtering approaches for large recommender
  systems.
\newblock {\em Journal of Machine Learning Research\/}~{\em 10\/}(22),
  623--656.

\bibitem[\protect\citeauthoryear{Therneau and Atkinson}{Therneau and
  Atkinson}{2022a}]{rpart}
Therneau, T. and B.~Atkinson (2022a).
\newblock {\em rpart: Recursive Partitioning and Regression Trees}.
\newblock R package version 4.1.19.

\bibitem[\protect\citeauthoryear{Therneau and Atkinson}{Therneau and
  Atkinson}{2022b}]{therneau2022introduction}
Therneau, T.~M. and E.~J. Atkinson (2022b).
\newblock An introduction to recursive partitioning using the rpart routines.
\newblock Technical report, Mayo Foundation.

\bibitem[\protect\citeauthoryear{Thomas, Ebert, Muruganujan, Mushayahama,
  Albou, and Mi}{Thomas et~al.}{2022}]{thomas2022panther}
Thomas, P.~D., D.~Ebert, A.~Muruganujan, T.~Mushayahama, L.-P. Albou, and H.~Mi
  (2022).
\newblock {PANTHER}: Making genome-scale phylogenetics accessible to all.
\newblock {\em Protein Science\/}~{\em 31\/}(1), 8--22.

\bibitem[\protect\citeauthoryear{Tipping}{Tipping}{1999}]{tipping1999relevance}
Tipping, M.~E. (1999).
\newblock The relevance vector machine.
\newblock In {\em Proceedings of the 12th International Conference on Neural
  Information Processing Systems}, pp.\  652--658.

\bibitem[\protect\citeauthoryear{Velten, Braunger, Argelaguet, Arnol, Wirbel,
  Bredikhin, Zeller, and Stegle}{Velten et~al.}{2022}]{velten2022identifying}
Velten, B., J.~M. Braunger, R.~Argelaguet, D.~Arnol, J.~Wirbel, D.~Bredikhin,
  G.~Zeller, and O.~Stegle (2022).
\newblock Identifying temporal and spatial patterns of variation from
  multimodal data using {MEFISTO}.
\newblock {\em Nature Methods\/}~{\em 19\/}(2), 179--186.

\bibitem[\protect\citeauthoryear{Wang and Stephens}{Wang and
  Stephens}{2021}]{wang2021empirical}
Wang, W. and M.~Stephens (2021).
\newblock Empirical {Bayes} matrix factorization.
\newblock {\em Journal of Machine Learning Research\/}~{\em 22\/}(120), 1--40.

\bibitem[\protect\citeauthoryear{Xu, Jin, and Zhou}{Xu
  et~al.}{2013}]{xu2013speedup}
Xu, M., R.~Jin, and Z.-H. Zhou (2013).
\newblock Speedup matrix completion with side information: Application to
  multi-label learning.
\newblock In {\em Proceedings of the 26th International Conference on Neural
  Information Processing Systems}, pp.\  2301--2309.

\bibitem[\protect\citeauthoryear{Yang, Wang, Zhang, and Zhao}{Yang
  et~al.}{2013}]{yang2013accounting}
Yang, C., L.~Wang, S.~Zhang, and H.~Zhao (2013).
\newblock Accounting for non-genetic factors by low-rank representation and
  sparse regression for {eQTL} mapping.
\newblock {\em Bioinformatics\/}~{\em 29\/}(8), 1026--1034.

\bibitem[\protect\citeauthoryear{Zakeri, Simm, Arany, ElShal, and
  Moreau}{Zakeri et~al.}{2018}]{zakeri2018gene}
Zakeri, P., J.~Simm, A.~Arany, S.~ElShal, and Y.~Moreau (2018).
\newblock Gene prioritization using {Bayesian} matrix factorization with
  genomic and phenotypic side information.
\newblock {\em Bioinformatics\/}~{\em 34\/}(13), i447--i456.

\bibitem[\protect\citeauthoryear{Zhou, Yang, and Yu}{Zhou
  et~al.}{2012}]{zhou2012moving}
Zhou, X., C.~Yang, and W.~Yu (2012).
\newblock Moving object detection by detecting contiguous outliers in the
  low-rank representation.
\newblock {\em IEEE Transactions on Pattern Analysis and Machine
  Intelligence\/}~{\em 35\/}(3), 597--610.

\bibitem[\protect\citeauthoryear{Zhou, Yang, Zhao, and Yu}{Zhou
  et~al.}{2014}]{zhou2014low}
Zhou, X., C.~Yang, H.~Zhao, and W.~Yu (2014).
\newblock Low-rank modeling and its applications in image analysis.
\newblock {\em ACM Computing Surveys\/}~{\em 47\/}(2), 1--33.

\bibitem[\protect\citeauthoryear{Zilber and Nadler}{Zilber and
  Nadler}{2022}]{zilber2022inductive}
Zilber, P. and B.~Nadler (2022).
\newblock Inductive matrix completion: No bad local minima and a fast
  algorithm.
\newblock In {\em Proceedings of the 39th International Conference on Machine
  Learning}, Volume 162, pp.\  27671--27692. PMLR.

\end{thebibliography}

\end{document}